\pdfoutput=1
\documentclass{article}

\usepackage[dvipsnames]{xcolor}

\usepackage{iclr2025_conference,times}

\usepackage[utf8]{inputenc} 
\usepackage[T1]{fontenc}    
\usepackage{hyperref}       
\usepackage{url}            
\usepackage{amsfonts}       
\usepackage{nicefrac}       
\usepackage{courier}
\usepackage[english]{babel}
\usepackage{ragged2e}
\usepackage{floatrow}
\usepackage{makecell}
\usepackage{tabularray}
\usepackage{tabularx}
\UseTblrLibrary{booktabs}
\usepackage{wrapfig}
\hypersetup{ %
    pdftitle={},
    pdfauthor={},
    pdfsubject={Proceedings of the International Conference on Machine Learning 2021},
    pdfkeywords={},
    pdfborder=0 0 0,
    pdfpagemode=UseNone,
    colorlinks=true,
    linkcolor=mydarkblue,
    citecolor=mydarkblue,
    filecolor=mydarkblue,
    urlcolor=cyan,
    pdfview=FitH}

\newfloatcommand{capbtabbox}{table}[][\FBwidth]

\usepackage[textwidth=1.7cm]{todonotes}

\iffalse
    
    \newcommand{\henry}[1]{}
    \newcommand{\yc}[1]{}
    \newcommand{\js}[1]{}
\else
    
    \newcommand{\yc}[1]{{\bf \color{blue}{[YC: #1]}}}
    \newcommand{\henry}[1]{{\bf \color{red}{[CH: #1]}}}
    
    \newcommand{\js}[1]{{\bf \color{purple}{[JS: #1]}}}
\fi

\usepackage{diagbox}
\usepackage{makecell}

\usepackage{booktabs}
\usepackage{graphicx}
\usepackage{multirow}
\usepackage{setspace}

\usepackage{subcaption}
\usepackage[font=small,labelfont=bf,tableposition=top]{caption}
\usepackage{amsmath}
\usepackage{amsthm}
\usepackage[english]{babel}
\usepackage[skins]{tcolorbox}
\usepackage{tikz}
\usepackage{color,soul}
\usepackage{enumitem}
\usepackage{amssymb}
\usepackage{pifont}
\newcommand{\cmark}{\ding{51}}%
\newcommand{\xmark}{\ding{55}}%
\usepackage{tabularx}
\usepackage{tablefootnote}
\usepackage{lipsum}

\usepackage{appendix}
\usepackage{etoc}

\usepackage{float}
\floatstyle{plaintop}
\restylefloat{table}

\definecolor{mydarkblue}{rgb}{0,0.08,0.45}

\usepackage{bm}
\usepackage{algorithm}
\usepackage[noend]{algpseudocode}
\usepackage{xspace}

\def\eqref#1{equation~\ref{#1}}

\def\1{\bm{1}}

\makeatletter
\newtheorem*{rep@theorem}{\rep@title}
\newcommand{\newreptheorem}[2]{%
\newenvironment{rep#1}[1]{%
 \def\rep@title{#2 \ref{##1}}%
 \begin{rep@theorem}}%
 {\end{rep@theorem}}}
\makeatother

\newtheorem{theorem}{Theorem}

\newtheorem{proposition}{Proposition}

\newtheorem{definition}{Definition}
\newtheorem{assumption}{Assumption}
\newreptheorem{theorem}{Theorem}
\newreptheorem{corollary}{Corollary}
\newreptheorem{proposition}{Proposition}

\def\rvs{{\mathbf{s}}}

\def\rvw{{\mathbf{w}}}
\def\rvx{{\mathbf{x}}}

\def\vk{{\bm{k}}}

\def\vq{{\bm{q}}}

\def\mA{{\bm{A}}}
\def\mB{{\bm{B}}}
\def\mC{{\bm{C}}}

\def\mI{{\bm{I}}}

\def\mP{{\bm{P}}}
\def\mQ{{\bm{Q}}}

\def\mS{{\bm{S}}}

\def\mU{{\bm{U}}}
\def\mV{{\bm{V}}}
\def\mW{{\bm{W}}}
\def\mX{{\bm{X}}}

\DeclareMathAlphabet{\mathsfit}{\encodingdefault}{\sfdefault}{m}{sl}
\SetMathAlphabet{\mathsfit}{bold}{\encodingdefault}{\sfdefault}{bx}{n}

\DeclareMathOperator{\Tr}{Tr}

\usepackage{colortbl}
\usepackage{multirow}

\newlength\myheight
\newlength\mydepth
\settototalheight\myheight{Xygp}
\definecolor{Gray}{gray}{0.85}
\definecolor{LightCyan}{rgb}{0.88,1,1}

\newcolumntype{a}{>{\columncolor{green!5!white}}c}
\newcolumntype{y}{>{\columncolor{red!5!white}}c}

\newcommand{\locogpt}{\text{MoDeGPT}\xspace} 

\DeclareCaptionLabelFormat{andtable}{#1~#2  \&  \tablename~\thetable}

\usepackage{bookmark}

\title{{\raisebox{-1pt}{\includegraphics[height=0.036\linewidth]{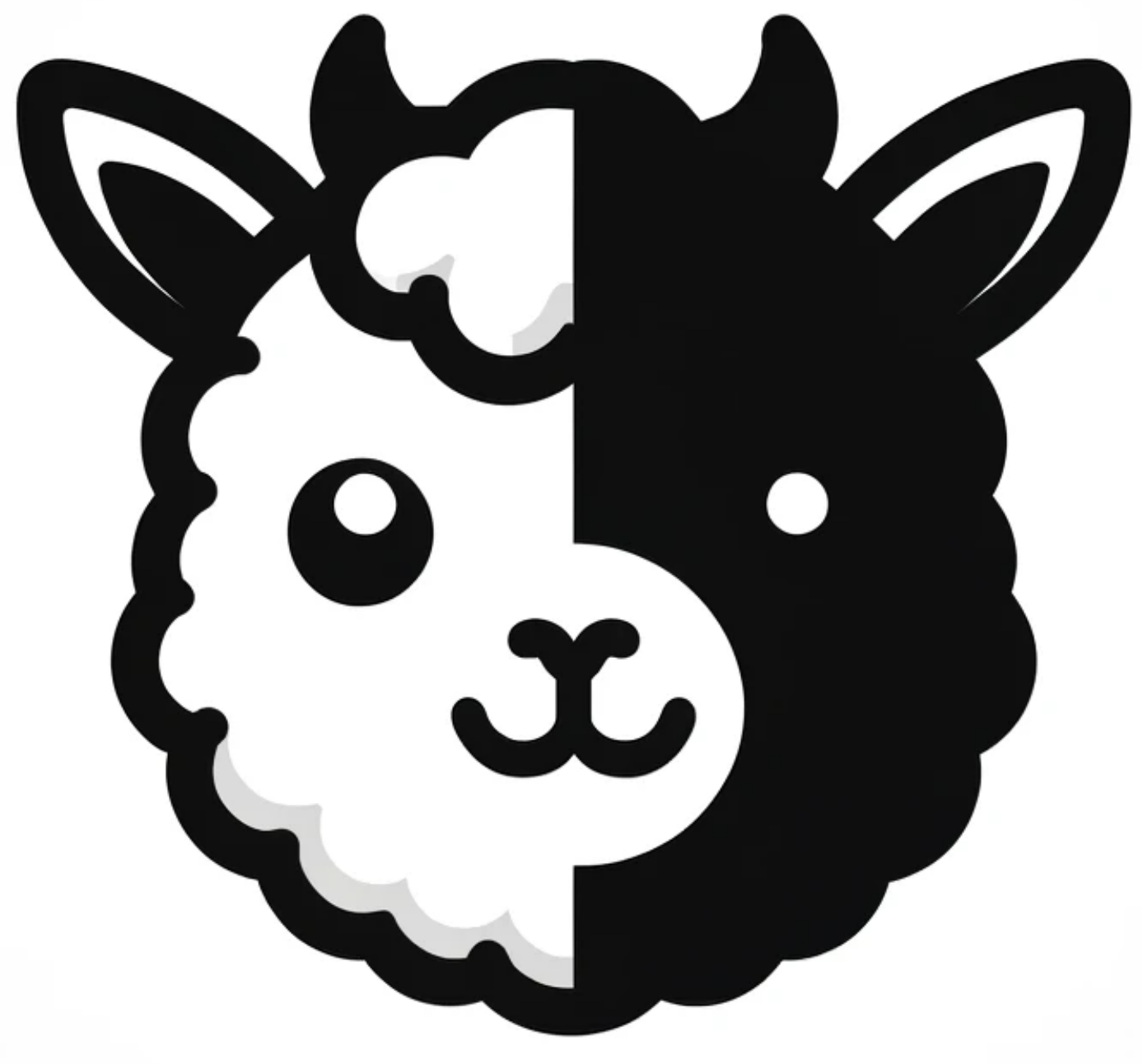}\hspace{-7pt}}} \locogpt: Modular Decomposition for
Large Language Model Compression}

\author{%
  Chi-Heng Lin\thanks{\texttt{chiheng.lin@samsung.com}}
  \\
  Samsung Research America\\
  \And
  Shangqian Gao\\
  Florida State University\\
  \And
  James Seale Smith \\
  Samsung Research America\\
  \AND
  Abhishek Patel\\
    Samsung Research America\\
  \And
  Shikhar Tuli\\
    Samsung Research America\\
    \And
  Yilin Shen\\
    Samsung Research America\\
   \AND
    Hongxia Jin\\
    Samsung Research America\\
   \And
    Yen-Chang Hsu \\
    Samsung Research America\\
}
\iclrfinalcopy
\begin{document}

\maketitle

\begin{abstract}

Large Language Models (LLMs) have significantly advanced AI with their exceptional performance across a wide range of tasks. However, their extensive computational requirements restrict their use on devices with limited resources. While recent compression methods based on low-rank matrices show potential solutions, they often suffer from significant loss of accuracy or introduce substantial overhead in parameters and inference time. In this paper, we introduce Modular Decomposition (MoDeGPT), a new, efficient, and structured compression framework that overcomes these limitations. MoDeGPT jointly decomposes pairs of consecutive subcomponents within Transformer blocks, reduces hidden dimensions through output reconstruction on a larger structural scale than conventional low-rank methods, and repurposes three classical matrix decomposition algorithms—Nyström approximation, CR decomposition, and SVD—to ensure bounded errors in our novel decomposition approach.
Our experiments show that MoDeGPT, without relying on backward propagation, consistently matches or surpasses the performance of prior techniques that depend on gradient information, while achieving a 98\% reduction in compute costs when compressing a 13B-parameter model. On LLaMA-2/3 and OPT models, MoDeGPT retains 90-95\% of zero-shot performance with compression rates of 25-30\%. The compression process can be completed on a single GPU in a few hours, boosting inference throughput by up to 46\%.

\end{abstract}

\etocdepthtag.toc{mtchapter}
\etocsettagdepth{mtchapter}{none}
\etocsettagdepth{mtappendix}{none}

\vspace{-8pt}
\section{Introduction}
Recent advancements in Large Language Models (LLMs) \citep{thoppilan2022lamda,openai2023gpt,touvron2023llama,zhang2022opt,llama3modelcard} have led to remarkable breakthroughs in the understanding and generation of natural language. Despite their significant capabilities, these models are computationally and memory-intensive, posing deployment challenges on resource-limited devices. To mitigate these challenges, model compression \citep{gupta2022compression,zhu2023survey} has emerged as a popular post-training solution, reducing model size and complexity.

\begin{figure}
    \vspace{-20pt}
    \includegraphics[width=1.\linewidth]{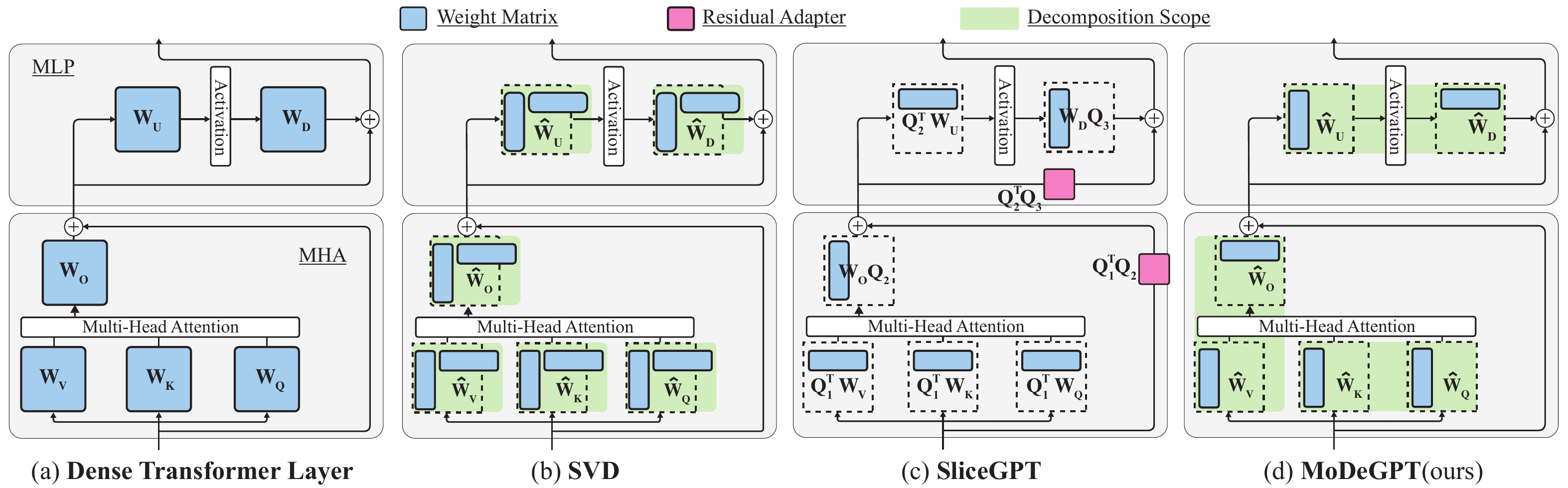}
    \caption{\textbf{Comparison of Matrix Decomposition-Based Methods for Transformer Compression.} \textbf{(a)} Original transformer layer. \textbf{(b)} SVD applied to each weight matrix separately, resulting in dual matrices. \textbf{(c)} SliceGPT multiplies each weight matrix by an orthogonal matrix $\mQ$, reducing dimensions and introducing additional adapters. \textbf{(d)} \locogpt\ organizes matrices into modules (highlighted by green boxes) and jointly decomposes them, producing reduced-size matrices without extra adapters.}
    \label{fig:overview}
\end{figure}

Predominant compression techniques encompass model distillation \citep{sun2019patient,sun2020contrastive,pan2020meta}, pruning \citep{lecun1989optimal,hassibi1993optimal,suzuki2018spectral,wang2019structured,zafrir2021prune,xia2022structured,kurtic2022optimal,ma2023llm,van2023llm}, matrix decomposition \citep{hsu2022language,noach2020compressing,golub1971singular}, and quantization \citep{gholami2022survey, bai2020binarybert, frantar2022gptq, wang2023bitnet}. This study focuses on matrix decomposition techniques that require minimal computing resources and do not involve backward propagation as seen in recovery fine-tuning (RFT) or Fisher matrix calculations from Taylor expansion \citep{ma2023llm,van2023llm}. Conventional matrix decomposition such as SVD typically splits each matrix
$\mW\in \mathbb{R}^{d\times d}$ into two low-rank matrices $\mW=\mA\mB$, requiring the rank less than
$d/2$ to achieve true compression, as shown in Figure \ref{fig:overview}(b).
This stringent requirement often results in a significant drop in accuracy, necessitating the use of RFT \citep{hsu2022language}. A novel decomposition approach, SliceGPT \citep{ashkboos2024slicegpt}, multiplies the original matrix by an orthogonal matrix, effectively projecting inputs into a lower-dimensional subspace and reducing the matrix's overall dimensionality. However, this approach requires additional adapters to manage the reduced dimensions; as illustrated in Figure \ref{fig:overview}(c), adapters $\mQ_i^\top\mQ_j$ are added to the residual paths to facilitate this reduction. For a target sparsity $s$, this introduces additional $2(1-s)^2d^2$ parameters per layer, which can add up to 10\% of additional parameters, significantly offsetting the parameter savings. 
{
In summary, matrix decomposition approaches either (i) \textbf{discard a large portion of ranks}, or (ii) \textbf{introduce substantial parameter overheads}. These challenges significantly hinder the effective reduction of parameters without compromising accuracy.
}


In response to these challenges, we introduce \locogpt, which applies matrix decomposition to multiple matrices jointly, avoiding the dual-matrix structure and extra adapters used in prior methods. As depicted in Figure \ref{fig:overview}(d), \locogpt elevates the matrix decomposition approach to a modular level by grouping weight matrices into modules and then applying matrix decomposition jointly within each module. Unlike SliceGPT, \locogpt reduces the intermediate dimensions within each module rather than between blocks, as illustrated by the matrix shapes in Figure \ref{fig:overview}(c) and (d). This crucial difference eliminates the need for adapters while still enabling dimension reduction in the compressed matrix.
Importantly, \locogpt establishes a comprehensive mathematical framework that maps each module's compression task to one of the three matrix approximation techniques: CR decomposition \citep{drineas2006fast}, singular value decomposition (SVD) \citep{golub1971singular}, and Nyström approximation \citep{gittens2013revisiting,musco2017recursive}. These methods enable \locogpt to efficiently compress matrices.
In summary, we make the following contributions:
\begin{itemize}
    \itemsep 1mm
    \item We introduce \locogpt, a training-free compression method that jointly decomposes multiple matrices within a module using closed-form expressions.  To our knowledge, this is the first method to apply matrix decomposition at the module level for model compression.
    \item We extend the theoretical foundations of language model weight decomposition beyond SVD, introducing a systematic framework for categorizing approximation challenges in Transformer compression, complete with error guarantees.
    \item To our knowledge, this is the first demonstration in large language models where a purely matrix decomposition-based approach achieves state-of-the-art structured compression efficiency, rivaling the compression rates of semi-structured pruning methods—all without the need for recovery fine-tuning
    \item We present a thorough evaluation of MoDeGPT, comparing it against existing methods across key metrics, including perplexity, downstream accuracy, and real-world speed improvements. MoDeGPT preserves up to 90\% of zero-shot performance with compression rates of up to 30\% on LLaMA 2 and 3, significantly outperforming prior approache. Moreover, MoDeGPT delivers a 46\% increase in inference throughput, further enhancing its practical value.
\end{itemize}

\vspace{-10pt}

\section{\textsc{Background and Related Work}}
In this section, we begin by reviewing the existing body of literature related to LLM compression, highlighting key contributions and methodologies in the field. Subsequently, we examine the standard components of a transformer decoder layer. Lastly, we delve into the three matrix approximations employed for our proposed compression across various components in a transformer layer.
\subsection{\textsc{Related Works}}
\paragraph{Pruning}
In early pruning methods like magnitude-based tuning, scalability is achieved but often at the cost of reduced effectiveness in large language models (LLMs) \citep{hagiwara1994simple, han2015learning, li2016pruning, frantar2023sparsegpt, van2023llm}. To improve performance while managing computational demands, frameworks such as Optimal Brain Damage \citep{lecun1989optimal} and Surgeon \citep{hassibi1993optimal, yu2022combinatorial, van2023llm} incorporate second-order loss information, necessitating substantial resources for Hessian calculations.
Recent adaptations like WoodFisher \citep{singh2020woodfisher}, Kronecker factorization \citep{wang2019eigendamage, van2023llm}, and layer-wise compression \citep{dong2017learning, frantar2022optimal} aim to streamline these intensive methods. Concurrently, learnable parameters for pruning in vision and language models have been investigated \citep{Liu2017learning, huang2018data, xia2022structured}, although these techniques generally demand significant computational resources for intensive backward propagation. Other approaches, such as feature-mimic-based methods \citep{an2024fluctuation, ji2024feature}, have not matched the performance of gradient-based methods like LLM Surgeon \citep{van2023llm}.
Alternatives like SparseGPT \citep{frantar2023sparsegpt}, Wanda \citep{sun2024a}, and ZeroPruner \citep{dong2024pruner}, exploring unstructured and semi-structured pruning, offer scalability but often compromise runtime speed. Additional research has utilized layer importance scores for layer pruning and sparsity distribution, as demonstrated by ShortGPT \citep{men2024shortgpt}, OWL \citep{yin2023outlier}, LaCo \citep{yang2024laco}, and others \citep{chen2024compressing}.
Recent advances in LLM compression have introduced innovative methods such as LLM-Pruner \citep{ma2023llm}, LLM Surgeon \citep{van2023llm}, and SliceGPT \citep{ashkboos2024slicegpt}, marking significant progress in the field by providing effective compression techniques for LLMs.

\begin{wraptable}[9]{r}{.4\textwidth}
\caption{{LLM Compression Comparisons.}}\label{tbl:realted}
\resizebox{1.\linewidth}{!}{
    \begingroup
    \setlength{\tabcolsep}{6pt} 
    \renewcommand{\arraystretch}{1.2} 

\Huge
\begin{tabular}{l|c|c|c}
\midrule[4pt]
\multicolumn{1}{c|}{\multirow{2}{*}{\textbf{Method}}} & \multirow{2}{*}{\textbf{\begin{tabular}[c]{@{}c@{}}No Backward \\ Propagation\end{tabular}}} & \multirow{2}{*}{\textbf{\begin{tabular}[c]{@{}c@{}} No Additional \\ Parameters\end{tabular}}} & \multirow{2}{*}{\textbf{\begin{tabular}[c]{@{}c@{}}Fully- \\ Structured\end{tabular}}} \\
                                 &                                                                                   &                                                                                                &                                                                                             \\ \midrule[4pt]
 LLM Pruner                      &    \xmark                                                          &      \cmark                                                                 &   \cmark                                                               \\ 
LLM Surgeon                      &     \xmark                                                                              &   \cmark                                                                                                    &   \cmark                                                                                   \\ 
SliceGPT                          &     \cmark                                                                              &   \xmark                                                                                               &   \cmark                                                                                          \\ 
SparseGPT                       &     \cmark                                                                              &   \cmark                                                                                             &   semi-  \\ 
\textbf{\locogpt  (ours)}                   &     \cmark                                                                              &     \cmark                                                                                                                                                 &   \cmark  \\ \midrule[4pt]
\end{tabular}
\endgroup
}
\end{wraptable}
\vspace{-8pt}
\paragraph{Low-Rank Matrix Approximation}
In related low-rank matrix techniques for compression, the traditional decomposition approach substitutes matrices with two low-rank matrices but retains the original dimensions, which can limit effectiveness \citep{noach2020compressing, hsu2022language, golub1971singular, povey2018semi, xu2023tensorgpt, yuan2023asvd, wang2024svd, yu2023compressing, chen2021drone}. \locogpt improves upon this by applying low-rank approximation to matrix pairs, reducing the size of individual matrices and merging the additional matrices from the decompositions. SliceGPT introduces a technique involving matrix multiplication with orthogonal matrices derived from PCA to compress weights, which reduces matrix sizes but adds additional parameters \citep{ashkboos2024slicegpt}. In contrast, \locogpt compresses without adding parameters by folding the decomposed matrices back to the original weights. A summary of \locogpt's comparison to other leading LLM compression methods is provided in Table \ref{tbl:realted}.

\vspace{-0pt}


\vspace{-5pt}
\subsection{\textsc{Transformer architecture}}
\label{sec:arch}
The transformer architecture \citep{vaswani2017attention} consists of multiple decoder layers.
 A typical layer such as in \textsc{Llama} \citep{touvron2023llama, llama3modelcard} includes two blocks: the Multi-Head Attention (MHA) and Multi-Layer Perceptron (MLP). Let $T $, $d_h$, $d_\text{int}$, and $H$ denote the sequence length, hidden dimension, intermediate dimension, and the number of attention heads, respectively, the formulation of these blocks is as follows:
\begin{align}
    \text{(MLP block)~~}&f_{\text{MLP}}(\mX) = 
    \overset{\text{\hyperref[sec:type1]{Type-I}}}{\colorbox{green!20}{\smash{$\sigma_s(\mX\mW_U)\mW_D$}\strut}}\label{block: mlp},\\
    \text{(MHA block)~~}&f_{\text{MHA}}(\mX) = 
    \sum_{i=1}^H \text{Softmax}\left(\overset{\text{\hyperref[sec:type2]{Type-II}}}{\colorbox{RubineRed!20}{\smash{$\sigma_{r}(\mX\mW_{Q,i})\sigma_{r}^\top(\mX\mW_{K,i})$}\strut}}\right)
    \overset{\text{\hyperref[sec:type3]{Type-III}}}{\colorbox{CornflowerBlue!20}{\smash{$\mX\mW_{V,i}{\mW_{O,i}}$}\strut}}\label{block: mha},
\end{align}
where $\mX\in\mathbb{R}^{T\times d_h}$ is the input matrix, $\mW_{Q,i}, \mW_{K,i}, \mW_{V,i} \in \mathbb{R}^{d_h\times \frac{d_h}{H}}, \mW_{O,i} \in \mathbb{R}^{\frac{d_h}{H}\times d_h}$ are the head-specific query, key, value, and output matrices. The matrices $\mW_U \in \mathbb{R}^{d_h \times d_{\text{int}}}$ and $\mW_D \in \mathbb{R}^{d_{\text{int}} \times d_h}$ denote up and down matrices, respectively, with $\sigma_r$ and $\sigma_s$ denoting positional embedding and nonlinear activation functions. 
Note that our MLP formulation encompasses the gated MLP: the up matrix is defined by the concatenations of the gated and up matrix ${\mW_U}=[\mW_u^\top,\mW_g^\top]^\top$, and the nonlinear function is defined by $\sigma_s(\mX{\mW_U}) := \mX\mW_u\odot\sigma_g(\mX\mW_g)$, where $\sigma_g$ is the gate function.

In the expressions of \eqref{block: mlp} and \eqref{block: mha}, the blocks can be divided into three types of functional modules, each associated with a pair of matrices:
\begin{align*}
    &f_{\text{{\hyperref[sec:type1]{{Type-I}}}}}(\mX;{\mW}_U,\mW_D) = {\colorbox{green!20}{\smash{$\sigma_s(\mX\mW_U)\mW_D$}\strut}},~~ f_{\text{\hyperref[sec:type2]{{Type-II}}}}(\mX;\mW_K^i,\mW_Q^i) = {\colorbox{RubineRed!20}{\smash{$\sigma_{r}(\mX\mW_{Q,i})\sigma_{r}^\top(\mX\mW_{K,i})$}\strut}}, \\
    &\text{\hspace{4cm} }f_{\text{\hyperref[sec:type3]{{Type-III}}}}(\mX;\mW_V^i,\mW_O^i) = {\colorbox{CornflowerBlue!20}{\smash{$\mX\mW_{V,i}{\mW_{O,i}}$}\strut}},
\end{align*}
where $\mX$ denotes the input and the variables after ``;'' denote the associated matrices.
These three types are distinguished by varying levels of nonlinearity. We will employ different matrix decomposition methods for compression based on the optimization tractability of each type.

\vspace{0pt}
\subsection{\textsc{Low-rank matrix approximation}}
\label{sec: mat}
The goal of a low-rank approximation method is to approximate a matrix $\mW\in\mathbb{R}^{d_1\times d_2}$ with two low-rank matrices $\mA\in\mathbb{R}^{d_1\times k}$ and $\mB\in\mathbb{R}^{k\times d_2}$.
For formalism, we make the following definition:
\begin{definition}
    \label{def:low_rank}
For a low-rank approximation method $\mathcal{M}$ that decomposes a matrix $\mW$ into $\mA$ and $\mB$, the approximation matrix is $\mW_\mathcal{M} = \mA\mB$ and the error relative to $\mW$ is $\mathcal{E}_\mathcal{M}(\mW) = \|\mW - \mW_\mathcal{M}\|_F$.
\end{definition}


We review three approximation methods that facilitate our algorithms in the next section. 
\vspace{-5pt}
\paragraph{{\text{{\color{Green}I.} Nystr\"om approximation}} \citep{gittens2013revisiting}} If $\mW$ is a positive semidefinite matrix, let $\mS_k$ be a $k$-column selection matrix where each column has a single non-zero element indicating the selected index, then the corresponding Nystr\"om approximation of $\mW$ is,
\begin{align}\label{nys approx}
    \mW_{\text{Nys}}=\mA\mB,~~\text{where}~~\mA=\mW\mS_k\text{~~and}~~\mB=(\mS^\top_k\mW\mS_k)^\dagger\mS_k^\top\mW.
\end{align}
\vspace{-23pt}
\paragraph{{\text{{\color{RubineRed}II.} CR decomposition \citep{drineas2006fast}}}}
Assuming $\mW$ can be factored as $\mW_1\mW_2$, let $\mS_k$ be a $k$-column selection matrix, the corresponding CR approximation of $\mW$ is
\begin{align}\label{cr approx}
    \mW_{\text{CR}} = \mA\mB,~~\text{where}~~\mA=\mW_1\mS_k\text{~~and}~~\mB=\mS_k^\top\mW_2.
\end{align}
\vspace{-23pt}
\paragraph{{\text{{\color{CornflowerBlue}III.} Singular value decomposition}} \citep{golub1971singular}} SVD is renowned for yielding the minimum approximation error when measured in the Frobenius norm. It decomposes $\mW$ into:
\begin{align}
    \mW_{\text{SVD}} = \mA\mB,~~\text{where}~~\mA=\mU_k\text{~~and}~~\mB= \bm\Sigma_k \mV_k^\top.
\end{align}
Here, $\mU_k$ and $\mV_k$ are matrices containing the top-$k$ left and right singular vectors, respectively, and $\bm{\Sigma}_k$ is the diagonal matrix consisting of the top-$k$ singular values of $\mW$.


\vspace{-0pt}
\section{\textsc{\locogpt}}
\vspace{-0pt}
\locogpt introduces a module-level optimization that jointly compresses two matrices within each of our three defined functional modules, rather than compressing each matrix independently as in traditional low-rank approximation methods. 

\begin{wrapfigure}[19]{r}{0.5\textwidth}
    \centering 
    \vspace{-12pt}
    \includegraphics[width=1.\linewidth]{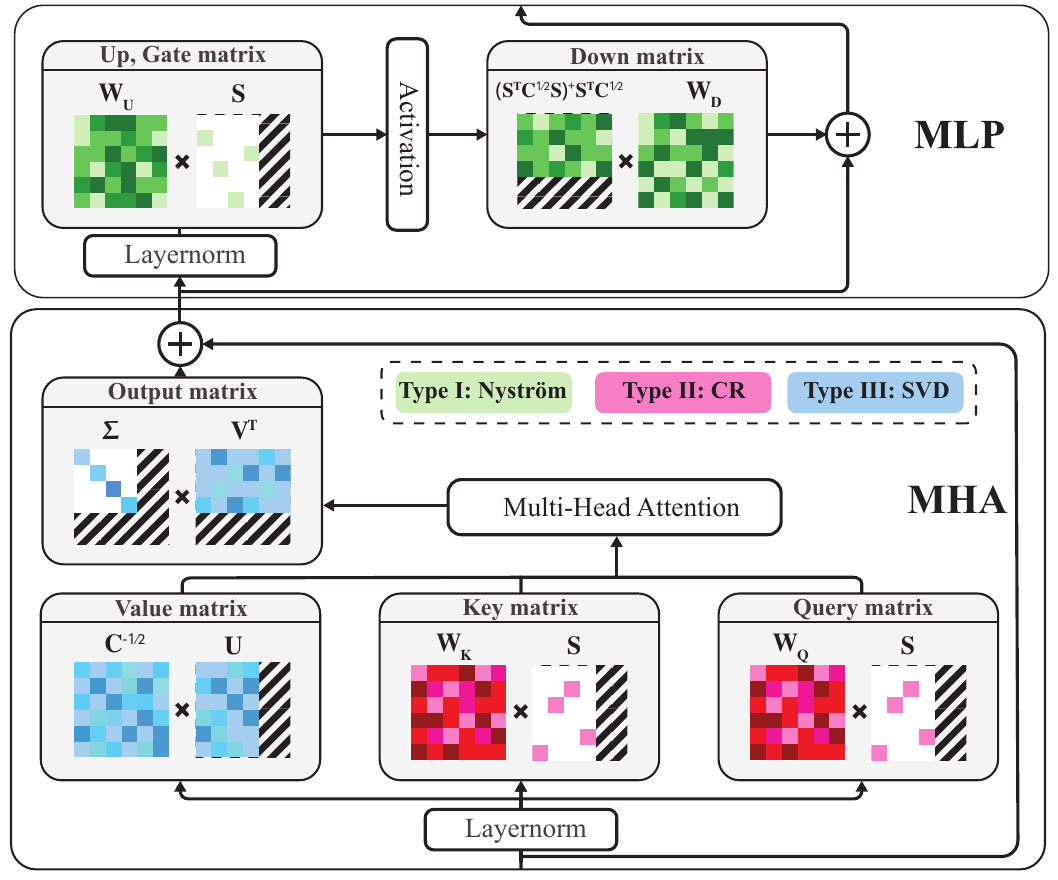}
    \caption{\textbf{The \locogpt Framework.} 
\locogpt divides a transformer layer into three distinct colored modules, each optimizing two matrices using a specific low-rank approximation method. A twill hatch pattern represents the dimension reduction.}

    \label{fig:flowchart}
\end{wrapfigure}

An illustration of \locogpt is presented in Figure \ref{fig:flowchart}, where different colors distinguish the various modules. For each module, we apply a tailored low-rank approximation to compress the matrix pair within it. The twill hatch pattern represents dimension reductions.

In the rest of this section, we first present the mathematical objective for our approach.
Then, we detail our application of low-rank approximations for effective compression within each module. Finally, we introduce a method for assigning sparsity levels across different layers that requires only one forward pass of the model on the calibration data.

\subsection{\textsc{Modular Reconstruction Objective}}\label{sec:obj}
The objective of \locogpt is to jointly optimize two matrices within the module types described in Sec. \ref{sec:arch}, a process we term modular decomposition, to minimize the modular reconstruction error:
\begin{align}\label{obj}
    {V}^*  \triangleq \min_{\hat{\mW}_1,\hat{\mW}_2}\sum_{i=1}^N\|f(\mX_i;\mW_1,\mW_2)-{f}(\mX_i;\hat{\mW}_1,\hat{\mW}_2)\|_F^2~~
    \text{such that } (\hat{\mW}_1, \hat{\mW}_2) \in \mathcal{C},
\end{align}
where $\mX_i\in\mathbb{R}^{T\times d}$ are samples in the calibration set, and $\mathcal{C}$ represents the constrained search space for compressed matrices that mandates specific structures or dimensions.

A key motivation of our objective is that it expands the search space to include dimension-reduced matrices, thereby increasing optimization flexibility and enhancing inference speedup in the compressed model. This contrasts with independent optimization, where each matrix must adhere to the original dimensions.




\subsection{\textsc{Algorithms}}

\paragraph{From LLM compression to matrix decomposition}
{The core technical contribution of this work is the establishment of a one-to-one mapping between a specific type of modular compression problem and a corresponding matrix decomposition problem.
As outlined in Section \ref{sec:arch}, the modules in the transformer architecture can be categorized based on the number of nonlinear functions they contain: Type I, II, and III modules contain 1, 2, and 0 nonlinearties, respectively. For a weight matrix $\mW$ within a nonlinear function, we compress it into a structured form $\hat{\mW} = \mW \mS_k$, where $\mS_k$ is a $k$-column selection matrix to be optimized. This restrictive structural form is a cornerstone of our framework, as it ensures the \textit{tractable} optimization of \eqref{obj}.

\begin{wrapfigure}[10]{r}{0.5\textwidth}
\capbtabbox{%
    \vspace{0pt}
    \resizebox{0.95\linewidth}{!}{
    \begingroup
    \setlength{\tabcolsep}{6pt} 
    \renewcommand{\arraystretch}{1.6} 
        \begin{tabular}{|c|c|c|c|c|}
            \hline
            \multicolumn{1}{|c|}{\textbf{Module Type}}                 & 
            \hyperref[sec:type1]{\cellcolor[RGB]{44,160,44}{\color{white}\textbf{I}}}
             & 
             \hyperref[sec:type2]{\cellcolor[RGB]{214,39,40}{\color{white}\textbf{{II}}}}
             & \hyperref[sec:type3]{\cellcolor[RGB]{31,119,180}{\color{white}\textbf{{III}}}} \\
            \hline
            \textbf{Weight Matrices}             & up,down,gate & key,query&  \multicolumn{1}{c|}{value,output} \\ \hline
            \textbf{Associated Decomp.}            & \multicolumn{1}{c|}{Nystr\"om}  & CR & SVD \\ \hline
            \textbf{\# Nonlinearities}        & \multicolumn{1}{c|}{1} & $2$ & $0$ \\ \hline
            \textbf{Compression Alg.}     &  Alg. \ref{alg:type1} &Alg. \ref{alg:type2} & Alg. \ref{alg:type3}\\
            \hline
        \end{tabular}
    \endgroup
    }
}
{\vspace{5pt}
  \caption{{Module characteristics and their associated matrix decompositions.}\label{roadmap}}
}
\end{wrapfigure}
After characterizing the modules and the structure of the compressed matrices, our framework solves the modular decomposition problem in \eqref{obj} for each module. Since each module contains a different number of nonlinear functions, the corresponding solutions vary. As we demonstrate in the subsequent sections, the solutions correspond to Nystr\"om, CR, and SVD for Type I, II, and III modules, respectively. A summary of this roadmap is provided in Table \ref{roadmap}. The detailed connections are formalized in the following subsections, with detailed proofs included in Appendix \ref{supp: proofs_main}.

}
\paragraph{{\color{Green}{\textsc{Type-I compression}}}}
\label{sec:type1}
First, we focus on the MLP module. As detailed in Section \ref{sec:arch}, the matrices $\mW_1$ and $\mW_2$ that require compression are ${\mW}_U$ and $\mW_D$. Since ${\mW}_U$ resides within a nonlinear function $\sigma_s$, we constrain its approximation to the form ${\mW}_U\mS_k$ for tractable optimization of \eqref{obj}, where $\mS_k$ is the $k$-column selection matrix.
For $\mW_D$, we simply ensure that its dimensions are compatible with ${\mW}_U\mS_k$. 
Our first theorem suggests that when a single column selection matrix is used, the optimization in \eqref{obj} is closely related to the Nyström approximation of the activation correlation matrix.
\begin{theorem}[\textbf{MLP compression by Nystr\"om approximation}]
    \label{thm: nys}
    Let $\hat{{\mW}}_U$ be searched over the matrix multiplication form ${\mW}_U\mS_k$, where $\mS_k$ is a $k$-column selection matrix, and let $\hat{\mW}_D$ be searched over $\mathbb{R}^{k\times d_h}$. The optimal $\hat{\mW}^*_D$ is then given by: $(\mS_k^\top\mC_{\sigma}\mS_k)^\dagger\mS_k^\top\mC_{\sigma}\mW_D$. Using ${\mW}_U\mS_k$ and $\hat{\mW}^*_D$ as the compressed matrices, the Type-I reconstruction error in \eqref{obj} satisfies:
    \begin{align}
        V_{\text{I}} \leq \left\|\mW_D\right\|_{2}^2\|\mC^{-1}_{\sigma}\|_2 \mathcal{E}^2_{\text{Nys}}(\mC_{\sigma}),
    \end{align}
 where $\mathcal{E}_{\text{Nys}}(\mC_{\sigma})$ denotes the Nyström approximation error, defined in Def. \ref{def:low_rank}, relative to the activation correlation $\mC_{\sigma} \triangleq \sum_{i=1}^N \sigma(\mX_i {\mW_U})^\top \sigma(\mX_i {\mW_U})$, 
        using the same $\mS_k$ in the compression of ${{\mW}}_U$.
\end{theorem}
Theorem \ref{thm: nys} shows that effective Type-I compression can be achieved through a well-designed Nyström approximation of $\mC_{\sigma}$. Thus, we propose Algorithm \ref{alg:type1} to control the error as shown below.


\begin{proposition}[\textbf{MLP compression error}]\label{thm: type1}
     Suppose that the rank $k$ and the scores $s_i$ in Algorithm \ref{alg:type1} are chosen such that there exists an error $\varepsilon>0$ satisfying $\varepsilon \geq \sum_{i=k+1}^{d_{int}} s_i$, then the Type-I modular reconstruction error in \eqref{obj} is bounded by $V_I \leq \|\mW_D\|_2^2\|\mC^{-1}_{\sigma}\|_2\frac{\varepsilon^2{d^2_{int}}}{k^2(1-\varepsilon)^2}\sum_{i=k+1}^{d_{int}}\sigma_i^2(\mC_{\sigma})$, where $d_{\text{int}}$ and $\sigma_i$ denote the intermediate dimension (i.e., the input dimension of $W_D$) and singular values, respectively.
\end{proposition}

\begin{algorithm}[t]
\setstretch{1.3}
\caption{{\color{Green}Type-I} compression for MLP by Nystr\"om approximation. 
} 
\label{alg:type1}
\begin{algorithmic}[1]
    \small
    \State \textbf{Input:} concatenated up and gated matrices ${\mW}_U\in\mathbb{R}^{d_h\times d_{\text{int}}}$, down matrix $\mW_D\in\mathbb{R}^{d_{\text{int}}\times d_h}$, activation correlation $\mC_{\sigma}=\sum_{i=1}^N\sigma(\mX_i{\mW}_U)^\top\sigma(\mX_i{\mW}_U)$, rank $k=\left \lceil{(1-\text{sparsity})d_{\text{int}}}\right \rceil$, and ridge intensity $\lambda$
    \State $s_i$ $\leftarrow [\mC_{\sigma}(\mC_{\sigma}+\lambda\mI)^{-1}]_{ii}$, for $i=1,\dots,d_{\text{int}}$ \Comment{{\color{green!50!black}Calculate the ridge leverage score}}
    \State Let $\mS_k \in \mathbb{R}^{d_{\text{int}} \times k}$ be the matrix that selects the top $k$ columns based on $s_i$ scores

    \State\Return $
({\mW}_U, {{\mW}}_D) \leftarrow ({\mW}_U\mS_k,~~(\mS_k^\top\mC_{\sigma}\mS_k)^\dagger\mS_k^\top\mC_{\sigma}\mW_D)$
\end{algorithmic}
\end{algorithm}

\paragraph{{\color{RubineRed}{\textsc{Type-II compression}}}}
\label{sec:type2}
Next, we turn our attention to the Type-II module, which includes the key-query interactions within the multi-head attention mechanisms. We will apply compression to each head independently \footnote{Dependency on the head is omitted in the equations for ease of notation.}. Given that both $\mW_Q$ and $\mW_K$ are embedded with nonlinear functions, for tractability in the optimization of \eqref{obj}, the matrices are compressed using a column selection matrix: $\hat{\mW}_Q = \mW_Q \mS_k$ and $\hat{\mW}_K = \mW_K \mS_k$, where $\mS_k$ is a shared $k$-column selection matrix.
When both two compressed matrices are multiplied by the column selection matrix, the modular reconstruction problem naturally connects to the CR decomposition of the product of key-query correlations, as elaborated in the following theorem.
\begin{theorem}[\textbf{Key-Query compression by CR approximation}]
    \label{thm: cr}
    Let the compressed $\hat{\mW}_Q$, $\hat{\mW}_K$ to be the form of $\mW_Q\mS_k, \mW_K\mS_k$, then Type-II reconstruction error in \eqref{obj} has 
    \begin{align}
        V_{\text{II}} \leq \mathcal{E}_{\text{CR}}^2(\mC^{\frac{1}{2}}_K\mC^{\frac{1}{2}}_Q),
    \end{align}
    where $\mathcal{E}_{\text{CR}}$ denotes the CR approximation error, defined in Def. \ref{def:low_rank}, relative to $\mC_Q^{1/2}\mC_K^{1/2}$, utilizing the same $\mS_k$ in the compression. Here, the matrices $\mC_Q \triangleq \sum_{i=1}^N \sigma(\mX_i \mW_Q)^\top \sigma(\mX_i \mW_Q)$ and $\mC_K \triangleq \sum_{i=1}^N \sigma(\mX_i \mW_K)^\top \sigma(\mX_i \mW_K)$ denote the correlations of query and key states, respectively.
\end{theorem}
The preceding theorem indicates that effective compression for the Type-II module can be achieved using a thoughtfully constructed CR approximation. In response, we present Algorithm \ref{alg:type2}, which offers the following guarantees for reconstruction:

\begin{algorithm}[t]
\setstretch{1.3}
\caption{{\color{RubineRed}Type-II} compression for key-query matrices by CR decomposition.
} 
\label{alg:type2}
\begin{algorithmic}[1]
    \small
    \State \textbf{Input:} head-specific query matrices $\mW_{Q,j}\in\mathbb{R}^{d_h\times d_h/H}$, key matrices $\mW_{K,j}\in\mathbb{R}^{d_h\times d_h/H}$, query state correlations $\mC_{Q, j}=\sum_{i=1}^N\sigma_r(\mX_i\mW_{Q,j})^\top\sigma_r(\mX_i\mW_{Q,j})$, key state correlations $\mC_{K, j}=\sum_{i=1}^N\sigma(\mX_i\mW_{K,j})^\top\sigma(\mX_i\mW_{K,j})$, for head $j=1,\dots,H$, and
    rank $k=\left \lceil{(1-\text{sparsity})d_h/H}\right \rceil$
    \For{$j=1,\dots,H$} \Comment{{\color{green!50!black}Apply compression to each head independently}}
    \State $s_i \leftarrow \|\mC_{Q,j}^{1/2}[:,i]\|\|\mC_{K,j}^{1/2}[:,i]\|$ \Comment{{\color{green!50!black}Calculate the norm score}}
    \State Let $\mS_k \in \mathbb{R}^{d_h \times k}$ be the matrix that selects the top $k$ columns based on $s_i$ scores

    \State $({\mW}_{Q,j}, {\mW}_{K,j}) \leftarrow ({\mW_{Q,j}}\mS_k, {\mW}_{K,j}\mS_k)$
    \EndFor
    \State\Return $({\mW}_{Q}, {\mW}_{K})\leftarrow ([{\mW}_{Q,1},\dots,{\mW}_{Q,H}],~~[{\mW}_{K,1},\dots,{\mW}_{K,H}])$ \Comment{{\color{green!50!black}Concatenate the heads}}
\end{algorithmic}
\end{algorithm}

\begin{proposition}[\textbf{Key-Query compression error}]\label{thm: type2}
    If we adopt Algorithm \ref{alg:type2} then Type-II modular reconstruction error is bounded by $V_\text{II}\leq \left(\frac{d_h-k}{d_h}\right)^2\left(\sum_{i=1}^{d_h}\sigma_i(\mC_{K})\right)\left(\sum_{i=1}^{d_h}\sigma_i(\mC_{Q})\right)$, where $\sigma_i$ denotes the singular values.
\end{proposition}

\paragraph{{\color{CornflowerBlue}{\textsc{Type-III compression}}}}
\label{sec:type3}
Finally, we focus on the Type-III module, which involves the value-output matrices. For clarity and simplicity, we omit the head dependency. The module has no nonlinar function involved $f(\mX) = \mX\hat{\mW}_V\hat{\mW}_O$, so we seek general low-rank matrices for compressions: $\hat{\mW}_V\in\mathbb{R}^{d_h\times k}$, $\hat{\mW}_O \in\mathbb{R}^{k\times d_h}$ such that $\hat{\mW}_V\hat{\mW}_O\approx {\mW}_V{\mW}_O$. 
The subsequent theorem reveals that the reconstruction can be solved optimally by applying the well-known Singular Value Decomposition.
\begin{theorem}[\textbf{Value-Output compression by SVD}]
    \label{thm: svd}
    If we search $\hat{\mW}_V$ and $\hat{\mW}_O$ over $\mathbb{R}^{d_h \times k}$ and $\mathbb{R}^{k \times d_h}$, respectively, the optimum in \eqref{obj} is $\hat{\mW}_V = \mC^{-1/2}\mU_k$ and $\hat{\mW}_O = \bm{\Sigma}\mV^\top$. Here, $\mU\bm{\Sigma}\mV^\top$ and  $\mC\triangleq \sum_{i=1}^N\mX_i^\top\mX_i$ are the SVD of $\mC^{1/2}\mW_V\mW_O$ and input correlation, respectively. The corresponding Type-III reconstruction error in \eqref{obj} is exactly the SVD approximation error, defined in Def. \ref{def:low_rank}, relative to $\mC^{\frac{1}{2}}\mW_V\mW_O$:
    \begin{align}
        V_{\text{III}} = \mathcal{E}^2_{\text{SVD}}(\mC^{\frac{1}{2}}\mW_V\mW_O).
    \end{align}
\end{theorem}
Building on the established equivalence to SVD via Theorem \ref{thm: svd}, we introduce Algorithm 3. This algorithm guarantees the following:
\begin{proposition}[\textbf{Value-Output compression error}]\label{thm: type3}
   Denote $\sigma_i$ as the singular values, Algorithm \ref{alg:type3} yields the optimal Type-III modular reconstruction error $ V_\text{III} = \sum_{i=k+1}^{d}\sigma_i^2(\mC^{\frac{1}{2}} {\mW_V} {\mW_O})$.
\end{proposition}

\begin{algorithm}[t]
\setstretch{1.3}
\caption{{\color{CornflowerBlue}Type-III} compression for value-output matrices by SVD. 
} 
\label{alg:type3}
\begin{algorithmic}[1]
    \small
    \State \textbf{Input:} head-specific value matrices $\mW_{V,j}\in\mathbb{R}^{d_h\times d_h/H}$, output matrices $\mW_{O,j}\in\mathbb{R}^{d_h/H\times d_h}$ for head $j=1,\dots,H$, input correlation $\mC=\sum_{i=1}^N\mX_i^\top\mX_i$, and rank $k=\left \lceil{(1-\text{sparsity})d_h/H}\right \rceil$
    \For{$j=1,\dots,H$} \Comment{{\color{green!50!black}Apply compression to each head independently}}
    \State   $\left(\mU,\bm{\Sigma},\mV^\top\right) \leftarrow SVD(\mC^{1/2} {\mW_{V,j}})$
    \Comment{{\color{green!50!black}Efficient SVD of $\mC^{1/2}\mW_{V,j}\mW_{O,j}$ (1/2)}} 
    \State$\left(\mU',\bm{\Sigma}',\mV'^\top\right) \leftarrow SVD(\bm{\Sigma}\mV^\top\mW_{O,j})$\Comment{{\color{green!50!black}Efficient SVD of $\mC^{1/2}\mW_{V,j}\mW_{O,j}$ (2/2)}}
    \State $(\mW_{V,j}, \mW_{O,j}) \leftarrow (\mC^{-1/2}\mU\mU'[:,:k],~~\bm{\Sigma}'[:k,:k]\mV'[:,:k]^\top)$
    \EndFor
    \State\Return $({\mW}_{V}, {\mW}_{O})\leftarrow ([{\mW}_{V,1},\dots,{\mW}_{V,H}],~~[{\mW}_{O,1},\dots,{\mW}_{O,H}])$\Comment{{\color{green!50!black}Concatenate the heads}}
\end{algorithmic}
\end{algorithm}


\subsection{\textsc{Global Sparsity Allocation}}
\label{sec:global_sparsity}
While \locogpt modules are optimized locally, we propose a global optimization strategy that translates layer importance scores into sparsity allocations across layers. This strategy seeks to maximize the sum of importance scores, weighted by the parameters retained in each layer. To avoid the negative effects of excessive sparsity \citep{yin2023outlier}, we incorporate entropic regularization for smoothing. The formulation of this constrained optimization problem is as follows:
\begin{align}\label{global rank eq}
    \max_{\phi_{1:L}}\sum_{i=1}^L s_i (1-\phi_i) +\varepsilon H(\phi_i)\quad \text{such that } \frac{1}{L}\sum_{i=1}^L \phi_i = \phi_{\text{avg}}, \quad 0\leq \phi_i\leq 1,
\end{align}
where $\phi_i$ and $s_i$ represent the sparsity and importance score of layer $i$, respectively, and $\phi_{\text{avg}}$ denotes the overall target sparsity. 
For sufficiently large 
$\varepsilon$, the following theorem demonstrates that the optimal layer sparsity distribution can be easily computed as:
\begin{align}\label{eq: softmax}
    \bm{\phi} = L\phi_{\text{avg}}\times\text{Softmax}(-\rvs/\varepsilon).
\end{align}
\begin{theorem}\label{thm:global}
    For sufficient large $\varepsilon$, (\ref{eq: softmax}) is the optimal sparsity allocation in the \eqref{global rank eq}.
\end{theorem}
In our implementations, we adopt the Block Influence (BI) score in \citet{men2024shortgpt}, which is the negative correlation between a layer's input and output defined by: $s=1-\mathbb{E}\rvx^{\top}_{\text{in}}\rvx_{\text{out}}/\|\rvx_{\text{in}}\|_2\|\rvx_{\text{out}}\|_2$.

\section{\textsc{Experiments}}
\subsection{\textsc{Setups}}
\paragraph{{Models}}
We evaluated \locogpt on several models that employ a sequential transformer block structure: OPT \citep{zhang2022opt} across multiple scales (125M, 1.3B, 2.7B, 6.7B), \textsc{LLaMA}-1 at 7B, \textsc{LLaMA}-2 \citep{touvron2023llama} at 7B, 13B, 70B, and \textsc{LLaMA}-3 \citep{llama3modelcard} at 8B.

\vspace{-10pt}
\paragraph{{Implementations and environments}}
We implemented our models using Hugging Face Transformers \citep{wolf2019huggingface}, with correlation computations in FP64. Model compression and performance testing were conducted on a single NVIDIA A100 80GB GPU, except for the 70B model, which we used 8 A100 GPUs. Additional details are in Appendix \ref{supp:impl}.
\vspace{-10pt}
\paragraph{Datasets} 
Following calibration setups similar to prior studies \citep{frantar2022gptq, ashkboos2024slicegpt, dettmers2023spqr}, we employed the WikiText-2 \citep{merity2016pointer} and Alpaca datasets \citep{taori2023stanford}, each comprising 128 samples of 2048 characters. Zero-shot performance was evaluated using the LM Evaluation Harness \citep{gao2021framework}, with task details provided in Appendix \ref{supp:impl}.
\vspace{-10pt}
\paragraph{Baseline comparisons}
We benchmarked our approach against several baselines. For non-gradient-based large language model pruning, we compared it with Uniform Pruning, Magnitude Pruning, SVD, SliceGPT \citep{ashkboos2024slicegpt}, ShortGPT \citep{men2024shortgpt}, SLEB \citep{song2024sleb} and Optimal Brain Damage \citep{lecun1989optimal}. For methods involving backward propagation, our comparisons included LLM-Pruner \citep{ma2023llm} and LLM Surgeon \citep{van2023llm}. Additionally, in Appendices \ref{sec:fm_svd} and \ref{sec:unstruct-vs}, we evaluated our methods against feature-mimic compression techniques and SVD-based methods, respectively.
\vspace{-5pt}
\subsection{\textsc{Generation Performance}}
\begin{table}[!htbp]
\centering
    \caption{Perplexity comparisons of structured pruning methods for \textsc{llama}-2 7B and 13B on WikiText-2, calibrated with 128 sequences of 2048 tokens. \label{tbl:gradient}
    }
\setlength\extrarowheight{2pt} 
\resizebox{1\linewidth}{!}{
\begin{tabular}{cccccccccccc}
\midrule[1pt]                                       \multirow{2}{*}{\textbf{Method}} &        \multicolumn{1}{|c|}{\textbf{No}} 
                                & 
                                \multicolumn{10}{c}{~~~~~~~\textbf{7B} (ppl: 5.12 {\color{Red}$\downarrow$}) ~~~~~~~\textbf{\textsc{Llama-2}}
                                ~~~~~~~~
                                    \textbf{13B} (ppl: 4.57 {\color{Red}$\downarrow$})}                                                                                                                                                                \\    &
\multicolumn{1}{|c|}{\small\textbf{Gradient}}               & \cellcolor{white}10\%                     & \cellcolor{white}20\%                     & \cellcolor{white}30\%          & \cellcolor{white}40\%          & \cellcolor{white}50\%           & \cellcolor{white}10\% & \cellcolor{white}20\% & \cellcolor{white}30\%          & \cellcolor{white}40\%          & \cellcolor{white}50\%                              \\ \midrule[1pt]
\multicolumn{1}{l|}{K-OBD \citep{lecun1989optimal}}                    & \multicolumn{1}{c|}\xmark      & \underline{5.48} & {9.14} & 15.43         & 28.03         & \multicolumn{1}{c|}{46.64}          & 4.91                     & \underline{6.29}                     & 10.08         & 13.06         & 16.06                             \\ 


\multicolumn{1}{l|}{LLM-Pruner \citep{ma2023llm} 
}    & \multicolumn{1}{c|}\xmark &  \multicolumn{1}{c} {7.11}     & {9.29}     &     13.56          &    17.90           & \multicolumn{1}{c|}{31.05}              &     5.57                     &    6.67                    &     12.19          &     19.56          &                32.20                   \\ 


\multicolumn{1}{l|}{LLM surgeon \citep{van2023llm}}                    &   \multicolumn{1}{c|}\xmark  &\textbf{5.25}            & \underline{6.18}                     & \underline{7.83}          & \underline{10.39}         & \multicolumn{1}{c|}{\underline{15.38}}          & \textbf{4.69}                     & \textbf{5.29}            & \underline{6.21} & \underline{7.25}          & \underline{9.43}          \\ \midrule[1pt] \midrule[1pt] 
\multicolumn{1}{l|}{Uniform}   &   \multicolumn{1}{c|}\cmark            &  19.09     &  27.13          & 46.25      &    176.24   &  \multicolumn{1}{c|}{327.99}          &   13.78          &  18.18       &29.05 &     45.44        &  82.60         \\
\multicolumn{1}{l|}{Magnitude}   &   \multicolumn{1}{c|}\cmark     &     861.76    &    821.34        &  9623              &  Overflow &\multicolumn{1}{c|}{Overflow}          &        22.41     &   320.39      & 723.60&   2105          &  3004         \\
\multicolumn{1}{l|}{SVD}   &  \multicolumn{1}{c|} \cmark            &  Overflow       &   Overflow           &   52719    &  51229      & \multicolumn{1}{c|}{Overflow}          &       7655         &    9919      & 21248 &           53672  &   39521        \\


\multicolumn{1}{l|}{ShortGPT \citep{men2024shortgpt} }  &  \multicolumn{1}{c|} \cmark                   &   6.98         &     14.31                &  33.21        &  71.04        & \multicolumn{1}{c|}{268.11}          &   5.40                 &  7.69        & {30.48} &     48.83      & 187.23         \\

\multicolumn{1}{l|}{SLEB \citep{song2024sleb} }  &  \multicolumn{1}{c|} \cmark                   &   6.05         &     7.64               &  11.23        &  29.10      & \multicolumn{1}{c|}{103.38}          &  5.23                 &  6.31        & 8.24 &   11.76        &  27.67        \\
 

\multicolumn{1}{l|}{SliceGPT \citep{ashkboos2024slicegpt} }  & \multicolumn{1}{c|}  \cmark                   & {6.46}            & 7.68                     & 10.47         & 15.19         & \multicolumn{1}{c|}{24.82}          & 5.67                     & 6.68            & {8.68} & 12.56          & 20.57          \\

\rowcolor{blue!5!white}
\multicolumn{1}{l|}{\cellcolor{blue!5!white}\locogpt (ours)} & \multicolumn{1}{c|}{\cmark}  
& \underline{5.48}                     & \textbf{6.16}            & \textbf{7.51} & \textbf{8.41} & \multicolumn{1}{c|}{\textbf{11.88}} &          \underline{4.83}                & \textbf{5.29}            & \textbf{6.10}  & \textbf{6.95} & \textbf{8.95} 
\\

 \midrule[1pt]
\end{tabular}
}
\end{table}

\vspace{-5pt}
We evaluated the generation performance of compressed \textsc{Llama}-2 models (7B and 13B) using the WikiText-2 test split in Table \ref{tbl:gradient}, \ref{tbl:vsSemi} and \ref{supp:add_gen}.
Results for OPT and \textsc{Llama}-3 8B are included in Appendices \ref{supp:gqa} and \ref{supp:add_gen}. The table distinguishes between compression methods using gradients (top rows) and those without (bottom rows).
Among non-gradient methods, the traditional matrix decomposition approach using SVD performed the worst. In sharp contrast, \locogpt outperformed all other baselines at various compression rates by jointly applying decomposition to multiple matrices within a module; it only increased the perplexity by 20\% for 20\% compression of the 7B model, which is substantially better than the next best alternative that saw a 50\% increase. In comparison to gradient-based methods, \locogpt surpassed other structured compression 
techniques except for a low compression rate (20\%). This demonstrates that using local reconstruction as a proxy for true loss can achieve state-of-the-art compression.

\vspace{-10pt}
\subsection{\textsc{Zero-shot performance}}
\vspace{-1pt}

\begin{table}[t!]
\centering
\caption{Zero-shot task performance of compressed \textsc{Llama}-2 7B and \textsc{Llama}-3 8B.\label{tbl:zero_shot_main}}
\setlength\extrarowheight{2pt}
\hspace{-115pt}
\resizebox{.72\linewidth}{!}{
\begin{tabularx}{\textwidth}{c|c|c|c|c|c|c|c|c}
\cmidrule[1pt]{1-9}
\textbf{Model}  & \textbf{Compress.} & \textbf{Method} & \textbf{ARC-e} & \textbf{ARC-c} & \textbf{PIQA} & \textbf{WinoG.} & \textbf{HellaS.} & \textbf{Average} \\
\cmidrule[1pt]{1-9}

\multirow{13}{*}{\begin{tabular}[c]{@{}c@{}}\textbf{\textsc{llama}-2} \\ \textbf{7B}\end{tabular}} & 0\% & \multicolumn{1}{l|}{Dense} & 74.58 & 46.25 & 79.11 & 69.06 & 75.99 & 69.00 \\

\cmidrule{2-9}

& \multirow{5}{*}{30\%} & \multicolumn{1}{l|}{ShortGPT \citep{men2024shortgpt}} & 48.65 & 32.85 & 64.31 & 64.33 & 56.13 & 53.25 \\
& & \multicolumn{1}{l|}{SliceGPT \citep{ashkboos2024slicegpt}} & 58.88 & 33.36 & 68.55 & 58.01 & 49.86 & 53.73 \\
& & \multicolumn{1}{l|}{LLM surgeon \citep{van2023llm}} & \underline{63.09} & 36.69 & \underline{\textbf{73.56}} & 61.09 & 60.72 & 59.03 \\
& \cellcolor{white!50} & \multicolumn{1}{l|}{\cellcolor{blue!5!white}\locogpt (ours)} & \cellcolor{blue!5!white}63.26 & \cellcolor{blue!5!white}\underline{38.73} & \cellcolor{blue!5!white}70.40 & \cellcolor{blue!5!white}\underline{\textbf{67.32}} & \cellcolor{blue!5!white}\underline{63.26} & \cellcolor{blue!5!white}\underline{60.78} \\
& & \multicolumn{1}{l|}{\cellcolor{blue!12!white}\locogpt-Alpaca (ours)} & \cellcolor{blue!12!white}\textbf{65.49} & \cellcolor{blue!12!white}\textbf{39.16} & \cellcolor{blue!12!white}73.34 & \cellcolor{blue!12!white}66.22 & \cellcolor{blue!12!white}\textbf{65.90} & \cellcolor{blue!12!white}\textbf{62.02} \\

\cmidrule{2-9}

& \multirow{5}{*}{40\%} & \multicolumn{1}{l|}{ShortGPT \citep{men2024shortgpt}} & 41.16 & 29.94 & 60.12 & 60.46 & 43.67 & 47.07 \\
& & \multicolumn{1}{l|}{SliceGPT \citep{ashkboos2024slicegpt}} & 36.49 & 24.57 & 54.90 & 53.43 & 34.80 & 40.84 \\
& & \multicolumn{1}{l|}{LLM surgeon \citep{van2023llm}} & \underline{52.31} & \underline{30.29} & \underline{69.26} & 54.38 & 48.04 & 50.86 \\
& \cellcolor{white!50} & \multicolumn{1}{l|}{\cellcolor{blue!5!white}\locogpt (ours)} & \cellcolor{blue!5!white}49.45 & \cellcolor{blue!5!white}30.03 & \cellcolor{blue!5!white}64.96 & \cellcolor{blue!5!white}\underline{61.96} & \cellcolor{blue!5!white}\underline{53.01} & \cellcolor{blue!5!white}\underline{51.88} \\
& & \multicolumn{1}{l|}{\cellcolor{blue!12!white}\locogpt-Alpaca (ours)} & \cellcolor{blue!12!white}\textbf{59.76} & \cellcolor{blue!12!white}\textbf{34.73} & \cellcolor{blue!12!white}\textbf{70.35} & \cellcolor{blue!12!white}\textbf{64.40} & \cellcolor{blue!12!white}\textbf{58.63} & \cellcolor{blue!12!white}\textbf{57.58} \\

\cmidrule[1pt]{1-9}
\morecmidrules
\cmidrule[1pt]{1-9}

\multirow{4}{*}{\begin{tabular}[c]{@{}c@{}}\textbf{\textsc{llama}-3} \\ \textbf{8B}\end{tabular}} & 0\% & \multicolumn{1}{l|}{Dense} & 77.69 & 53.58 & 80.63 & 72.69 & 79.16 & 72.75 \\

\cmidrule{2-9}

& \multirow{3}{*}{25\%} & \multicolumn{1}{l|}{ShortGPT-Alpaca \citep{men2024shortgpt}} & 38.13 & 31.40 & 60.94 & 54.22 & 31.52 & 43.24 \\
& & \multicolumn{1}{l|}{SliceGPT-Alpaca \citep{ashkboos2024slicegpt}} & 44.44 & 29.27 & 57.56 & 58.48 & 41.08 & 46.17 \\
& & \multicolumn{1}{l|}{\cellcolor{blue!12!white}\locogpt-Alpaca (ours)} & \cellcolor{blue!12!white}\textbf{67.05} & \cellcolor{blue!12!white}\textbf{41.13} & \cellcolor{blue!12!white}\textbf{75.52} & \cellcolor{blue!12!white}\textbf{69.61} & \cellcolor{blue!12!white}\textbf{66.49} & \cellcolor{blue!12!white}\textbf{63.96} \\
\cmidrule[1pt]{1-9}

\end{tabularx}
}
\end{table}
\vspace{-5pt}
\begin{table}[htb!]
\centering
\caption{{Comparisons of 30\% compression on \textsc{Llama}-2 70B using 128 wikitext-2 samples for calibration.}\label{tbl:large_model}}
\setlength\extrarowheight{7pt}
\renewcommand{\arraystretch}{1.35} 
\resizebox{1\linewidth}{!}{
\begin{tabular}{l|c|ccccccccccc|c}
\midrule[1pt]
\multicolumn{1}{c|}{\Large \textbf{Method}} &{\color{Red} \Large $\downarrow$} \Large \textbf{WikitText-2} &{\color{Green} \Large $\uparrow$} \Large \textbf{ARC-e} & \Large \textbf{ARC-c} & \Large \textbf{PIQA} & \Large \textbf{WinoG.} & \Large \textbf{HellaS.} & \Large \textbf{BoolQ} & \Large \textbf{OBQA} & \Large \textbf{MathQA} & \Large \textbf{MMLU-ml} & \Large \textbf{COPA} & \Large \textbf{Lamb.} & {\color{Green} \Large $\uparrow$} \Large \textbf{Average.} \\ 
\midrule[1pt]
\Large Dense \textsc{llama}-2 70B & \Large 3.12 & \Large 80.98 & \Large 57.25 & \Large 82.75 & \Large 77.90 & \Large 83.83 & \Large 83.79 & \Large 48.80 & \Large 38.42 & \Large 42.86 & \Large 94.00 & \Large 79.60 & \Large 70.02 \\ \cmidrule[1pt]{1-14} 
\Large SliceGPT \citep{ashkboos2024slicegpt} & \Large 5.76 & \Large 67.05 & \Large 42.06 & \Large 67.52 & \Large 71.11 & \Large 55.57 & \Large 41.56 & \Large 40.20 & \Large 27.87 & \Large 32.14 & \Large 82.00 & \Large 52.03 & \Large 52.65 \\ 
\Large ShortGPT \citep{men2024shortgpt} & \Large 66.33 & \Large 60.65 & \Large {34.47} & \Large {72.74} & \Large 64.01 & \Large {63.80} & \Large 66.88 & \Large 34.40 & \Large 23.05 & \Large 31.25 & \Large 75.00 & \Large 27.01 & \Large 48.06\\ 
\Large SLEB \citep{song2024sleb} & \Large 5.54 & \Large 71.97 & \Large 44.20 & \Large \underline{77.74} & \Large 69.38 & \Large {73.54} & \Large {67.25} & \Large 41.80 & \Large 27.47 & \Large {32.15} & \Large {88.00} & \Large 64.22 & \Large 59.79 \\ \cmidrule[1pt]{1-14} 
\rowcolor{blue!5!white}
\Large \locogpt + OWL sparsity & \Large \bf 4.67 & \Large {76.01} & \Large {50.34} & \Large 74.70 & \Large {72.85} & \Large 72.43 & \Large {69.88} & \Large {44.20} & \Large {32.26} & \Large \bf 44.64 & \Large 87.00 & \Large {69.61} & \Large {63.08} \\
\rowcolor{blue!5!white}
\Large \locogpt + our sparsity & \Large \underline{4.89} & \Large \underline{77.69} & \Large \underline{50.94} & \Large {77.53} & \Large \underline{76.87} & \Large \underline{78.16} & \Large \underline{74.71} & \Large \underline{45.60} & \Large \bf 35.04 & \Large \underline{42.86} & \Large \underline{89.00} & \Large \bf 72.17 & \Large \underline{65.51} \\
\midrule[1pt]
\rowcolor{blue!12!white}
\Large \locogpt + our sparsity + Alpaca & \Large {5.73} & \Large \bf{78.57} & \Large \bf 51.54 & \Large \bf{80.85} & \Large \bf 77.19 & \Large \bf 79.60 & \Large \bf{82.81} & \Large \bf 46.40 & \Large \underline{32.83} & \Large {40.18} & \Large \bf 94.00 & \Large \underline{70.72} & \Large \bf 66.79 \\
\midrule[1pt]
\end{tabular}
}
\end{table}

We evaluated our method on zero-shot tasks, comparing it to leading baselines in Table \ref{tbl:zero_shot_main}. Our method showed superior performance at higher compression rates. The bottom rows indicate that calibrating with the Alpaca dataset (instead of WikiText-2) significantly improved performance, with a 30\% compression resulting in only a 10\% accuracy drop. This effect was more pronounced for \textsc{Llama}-13B, as shown in Table \ref{tbl:alpaca13b} in Appendix \ref{supp:add_gen}. We also tested the newer \textsc{llama}-3 8B model, adapting our algorithm for grouped query attention head dependency as detailed in Appendix \ref{supp:gqa}. The performance gap between our method and baselines was notably larger with this model, aligning with quantization challenges observed in \citep{huang2024empirical}.

\begin{wrapfigure}[15]{r}{0.35\textwidth}
    \centering
    \vspace{-18pt}
    \ttabbox{
    \Large 
        \setlength\extrarowheight{2pt}
        \resizebox{.9\linewidth}{!}{
        \renewcommand{\arraystretch}{1.35} 
        \begin{tabular}{|c|ll|ll|}
        \hline
        \multirow{2}{*}{ \textbf{Model}} & \multicolumn{2}{c|}{\textbf{\locogpt (ours)} } & \multicolumn{2}{c|}{ \textbf{LLM surgeon }} \\
        & \multicolumn{1}{c|}{Time} & \multicolumn{1}{c|}{GPUs} & \multicolumn{1}{c|}{Time} & \multicolumn{1}{c|}{GPUs} \\ \hline
         \textsc{Llama-2} 7B & \multicolumn{1}{l|}{4h09m} & 1xA100 & \multicolumn{1}{l|}{17h08m} & 4xH100 \\ \hline
         \textsc{Llama-2} 13B & \multicolumn{1}{l|}{8h26m} & 1xA100 & \multicolumn{1}{l|}{1d9h26m} & 8xH100 \\ \hline
        \end{tabular}
        }
    }{
        \caption{Compute time.}
        \label{tbl:compute}
    }
    \vspace{15pt}
    
    \ffigbox[\FBwidth]{
    \vspace{-15pt}
    \hspace{-10pt}
        \includegraphics[clip, trim=0cm 0cm 0cm 0cm,width=1\linewidth]{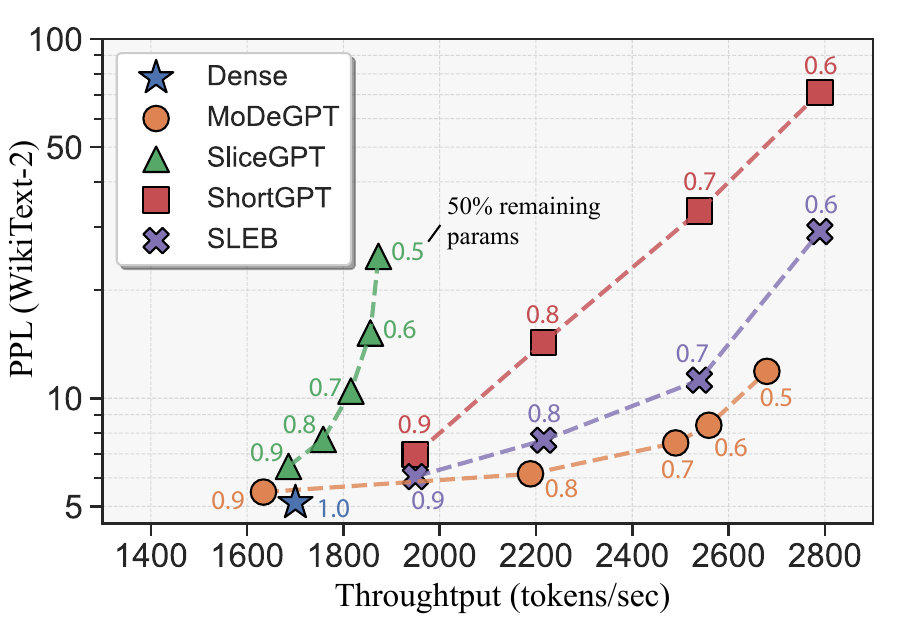} 
    }
    {\vspace{-2pt}
        \caption{{PPL vs. throughput.}}
        \label{fig:speed_ppl}
    }
\end{wrapfigure}
Finally, we compared our method against decomposition and layer-pruning baselines in a large-scale experiment on the \textsc{LLAMA}-2 70B, as shown in Table \ref{tbl:large_model}. Our method demonstrates improved performance in larger models, with minimal drops of $4.51\%$ and $3.23\%$ in zero-shot task performance at $30\%$ compression. This is achieved using only $128$ samples from WikiText-2 and Alpaca, respectively, for calibration, without requiring recovery fine-tuning. This result highlights the scalability and effectiveness of our approach in large models.
On the middle two rows, we compared our sparsity allocation strategy with the recent state-of-the-art OWL method \citep{yin2023outlier}. While our method shows a slightly higher perplexity, it consistently achieves superior zero-shot performances. 
Appendix \ref{supp:global_sa} provides additional analysis on the comparisons with OWL.


\vspace{-5pt}\subsection{\textsc{Computation and Throughput}}
\vspace{-5pt}

Table \ref{tbl:compute} compares the compression costs of \locogpt with LLM surgeon, the best-performing prior work. 
Given that the hourly rate for H100 is about twice that of A100 \citep{lambda}, 
\locogpt offers significant cost savings—97\% for the 7B model and 98.5\% for the 13B model. 
{
Next, we analyze the trade-off between model performance, measured in perplexity (on WikiText-2), and throughput (tokens/sec), as illustrated in Figure \ref{fig:speed_ppl}. For this experiment, we set the sequence length to 256 and measured the average generation time of \textsc{Llama}-2 7B on a single A100 GPU with a batch size of 256.
As illustrated, \locogpt consistently outperforms other models in both speedup and accuracy across various compression sizes, with size ratios annotated alongside each point. Remarkably, at 50\% compression, \locogpt increases throughput by 58\% while maintaining perplexity on par with the top competitor at 30\% compression. Further details on speedup, including different batch sizes and hardware setups, are in Appendix \ref{supp:speedup}."

\begin{figure}

\hspace{-60pt}
\begin{floatrow}
\ffigbox{
  \includegraphics[width=.7\linewidth]{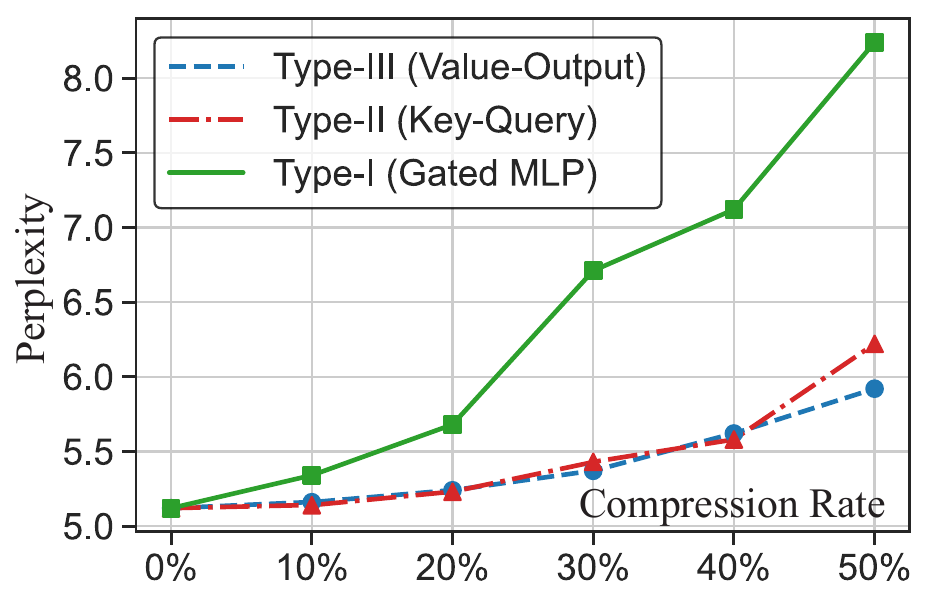}
}
{%
  \caption{{Module-wise compression.}
  }%
  \label{fig:component}
}
\hspace{-48pt}
\capbtabbox{%
    \vspace{-10pt}
    \resizebox{.85\linewidth}{!}{
    \begingroup
    \setlength{\tabcolsep}{6pt} 
    \renewcommand{\arraystretch}{1.47} 

        \begin{tabular}{|c|c|c|c|}
            \hline
            \multicolumn{1}{|c|}{\textbf{Type}}                 & 
            \hyperref[sec:type1]{\cellcolor[RGB]{44,160,44}{\color{white}\textbf{I}}}
             & 
             \hyperref[sec:type2]{\cellcolor[RGB]{214,39,40}{\color{white}\textbf{{II}}}}
         & \hyperref[sec:type3]{\cellcolor[RGB]{31,119,180}{\color{white}\textbf{{III}}}} \\
            \hline
            \textbf{Parameters}             & MLP &K-Q &  \multicolumn{1}{c|}{V-O} \\ \hline
            \textbf{Method}            & \multicolumn{1}{c|}{Nystr\"om}  & CR & SVD \\ \hline
            \textbf{Size Ratio}        & \multicolumn{1}{c|}{66.84\%} & $16.58\%$ & $16.58\%$ \\ \hline
            \textbf{Complexity}        & \multicolumn{1}{c|}{$O(d^3_{\text{int}})$} & $O(d^3_h/ H^2)$ & $O(Hd^3_h)$ \\ \hline
            \textbf{Effective $r$}     & 0.094 & 0.121 & 0.095\\
            \hline
            \multirow{1}{*}{\textbf{Time}} & \multicolumn{1}{c|}{1h13m} & 0h36m & 2h26m \\
                                           \hline
        \end{tabular}
    \endgroup
    }
}
{\vspace{10pt}
  \caption{{Module breakdown statistics.}}%
  \label{tbl:component}
}
\hspace{-30pt}
\capbtabbox{%
    \vspace{-19pt}
    
    \resizebox{.6906\linewidth}{!}{
    \begingroup
    \setlength{\tabcolsep}{6pt} 
    \renewcommand{\arraystretch}{1.4} 
        \begin{tabular}{|c|cc|}
            \hline
            \multirow{4}{*}{\bf Block} 
            & \multicolumn{2}{c|}{\textbf{\textsc{LLAMA}-7B}} \\ 
            & \multicolumn{2}{c|}{(model size: 13.81 GiB)} \\
            \cline{2-3}
            & \multicolumn{1}{c|}{\bf Peak Memory} & \bf GPU\\ 
            & \multicolumn{1}{c|}{(GiB)} & \bf hours \\ 
            \hline
            \multirow{2}{*}{\bf MHA}  & \multicolumn{1}{c|}{15.54} & \multirow{2}{*}{2h52m} \\
            & \multicolumn{1}{c|}{({+11.5\%})} &   \\
            \cline{1-3}
            \multirow{2}{*}{ \bf MLP}  & \multicolumn{1}{c|}{23.33} & \multirow{2}{*}{1h13m} \\
            & \multicolumn{1}{c|}{({+68.9\%})} &   \\
            \hline
        \end{tabular}    
    \endgroup
    }
}
{\vspace{10pt}
  \caption{{Memory utilizations.}}%
  \label{tbl:gpu_cons}
}

\end{floatrow}
\vspace{-15pt}
\end{figure}

\vspace{-10pt}
\subsection{\textsc{Ablation study}}
\begin{wrapfigure}[22]{r}{0.35\textwidth}
    \vspace{-35pt}
    \centering
    \ffigbox[\FBwidth]{
    \vspace{-10pt}
        \includegraphics[width=.98\linewidth]{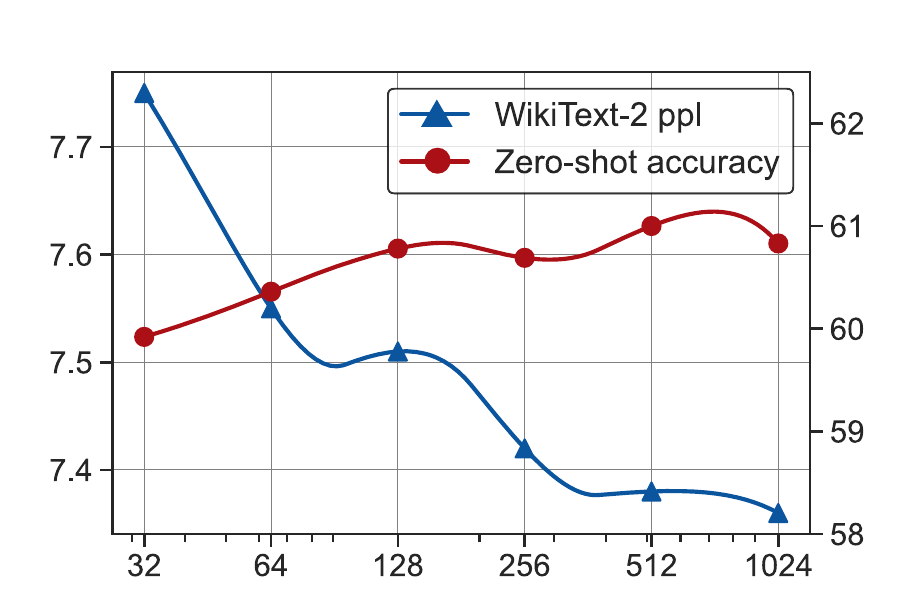} 
    }
    {\vspace{-0pt}
        \caption{Impact of calibration size.}
        \label{fig:cali_size}
    }
    \vspace{18pt}
        \ffigbox[\FBwidth]{
    \vspace{-10pt}
        \includegraphics[clip, trim=0cm 0cm 15.25cm 0cm,width=.95\linewidth]{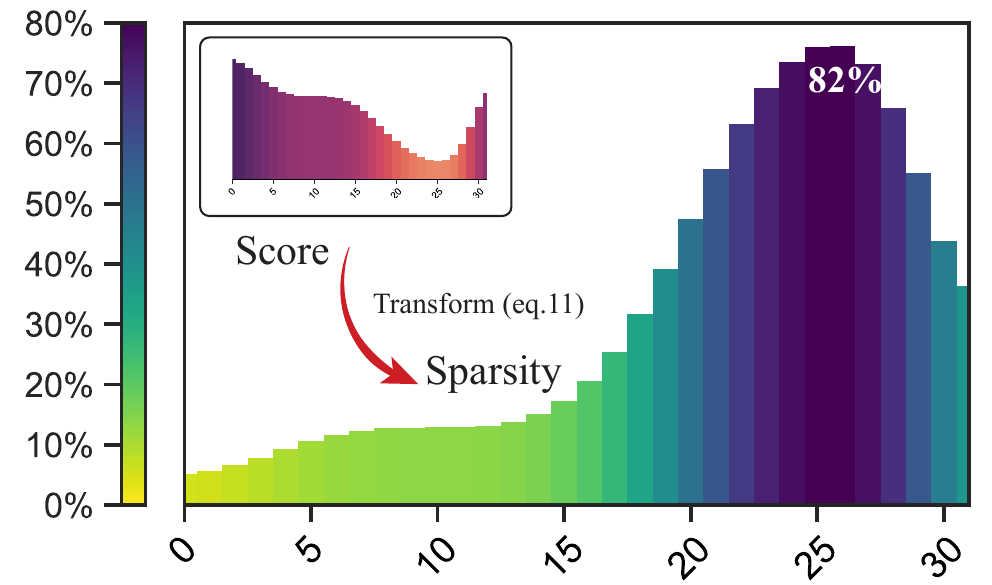} 
    }
    {\vspace{-5pt}
        \caption{Score-to-sparsity conversion.}
        \label{fig:score_sparse}
    }
    \vspace{8pt}
    \ttabbox{
        \resizebox{.94\linewidth}{!}{
        \renewcommand{\arraystretch}{1.2} 
        \begin{tabular}{|l|c|c|}
        \Xhline{1\arrayrulewidth}
        & \multicolumn{1}{l|}{\textbf{Perplexity}\color{Red} $\downarrow$} & \multicolumn{1}{l|}{\textbf{Zero-Shot Acc.\color{Green} $\uparrow$}} \\ \Xhline{1\arrayrulewidth}
        Dense & 5.12 & 69.00\% \\ \hline
        Uniform & 9.06 & 53.47\% \\ \hline
        Sparsity Alloc. & \underline{7.51} & \underline{60.78\%} \\ \hline
        \end{tabular}
        }
    }{
        \caption{Sparsity allocation.}
        \label{tbl:spar_alloc}
    }

\end{wrapfigure}
\vspace{-5pt}
We first analyzed the impact of compression on each module type within the \textsc{LLama}-2 7B model using a single A100 GPU. As shown in Figure \ref{fig:component}
, the majority of perplexity degradation occurs when the MLP module is compressed. However, after normalizing by parameter size, i.e., the effective ratio $r$ in Table \ref{tbl:component}, it becomes evident that the Type-II module is the most sensitive to compression. This observation aligns with our theoretical analysis, which demonstrates that Type-II has the weakest error bounds, as it is constrained by the complete spectrum rather than just the residuals (see Propositions \ref{thm: type1}, \ref{thm: type2}, \ref{thm: type3}).
In the middle, Table \ref{tbl:component} provides a detailed breakdown of the module-wise compression statistics. Notably, the SVD method dominates the compression time for the value-output components, suggesting that techniques such as SVD approximation \citep{yuan2023asvd} have the potential to reduce overall compression time. Meanwhile, Table \ref{tbl:gpu_cons} reports memory usage, showing that MLP compression requires the most memory, as it has the largest correlation dimension among the modules. Despite this, all compression tasks only consumed up to 23 GiB of memory, which is approximately double the model size. A similar memory consumption pattern for the 13B model is discussed in Appendix \ref{supp:memory}.
Second, we explored the effects of calibration size 
on a 30\% compressed \textsc{llama}-2 7B model. As shown in Figure \ref{fig:cali_size}, increasing the calibration size initially boosts performance; however, the gains in zero-shot performance diminish for sizes larger than 128.

Lastly, we evaluated sparsity allocation effects on the same model. Figure \ref{fig:score_sparse} shows the score-to-sparsity mapping from Section \ref{sec:global_sparsity} with $\varepsilon=0.1$, highlighting areas of higher sparsity in darker shades. Our findings indicate that certain layers, such as layer 26, can forgo up to 82\% of parameters with minimal accuracy loss. Table \ref{tbl:spar_alloc} demonstrates that our global sparsity allocation significantly surpasses a uniform approach, affirming the efficacy of our decomposition method with a simple scoring function for sparsity distribution.

\vspace{-10pt}
\section{\textsc{Conclusion}}
\vspace{-10pt}
In this work, we introduced \textbf{\locogpt}, a novel structured compression method that generalizes matrix decomposition to the modular level, achieving state-of-the-art results for structured model compression via low-rank decomposition.
Our approach has a strong theoretical grounding, offering bounds on the reconstruction errors for the components in the Transformers.
Furthermore, \locogpt stands out by relying solely on forward propagation to achieve comparable or better compression performance to methods that use the gradients from backward propagation. 
We believe our novel methods and findings will inspire more theoretical and algorithmic innovations for training-free model compression.

\bibliography{bibio}
\bibliographystyle{iclr2025_conference}

\newpage
\appendix
\section*{\centering\ul{Supplementary Material}}
\vspace{5mm}
\etocdepthtag.toc{mtappendix}
\etocsettagdepth{mtchapter}{none}
\etocsettagdepth{mtappendix}{subsection}
\tableofcontents
\newpage
\section{\textsc{Proofs}}
\label{supp: proofs_main}
This section provides proofs for the theorems and propositions in the main text, along with definitions and assumptions for formalism.

First, we define the following notation.
\begin{definition}[Column Selection Matrix]
A $k$-column selection matrix $\mS_k$ is a matrix with $k$ columns where each column has a single non-zero element indicating the selected index. For example, $\mS_3 = [[0, 0, 1, 0]^\top, [0, 1, 0, 0]^\top, [0, 0, 0, 1]^\top]$ is a 3-column selection matrix selecting the third, second, and fourth columns. An important property is that any matrix right-multiplied by a column selection matrix will result in a matrix consisting of the selected columns.
    
\end{definition}
Next, we make an assumption regarding the nonlinear functions used in all modules, which is crucial for validating our algorithms.
\begin{assumption}\label{assump: nonlinear}
    Any column selection matrix \(\mS\) is commutative with the nonlinear functions under consideration. Specifically, the function \(\sigma\) satisfies the property that \(\sigma(\mX)\mS = \sigma(\mX\mS)\) for any \(\mX\) and any column selection matrix $\mS$.
\end{assumption}

Importantly, Assumption \ref{assump: nonlinear} is met by any activation function that operates element-wise on the inputs, as well as by widely used embedding functions, such as the rotary positional embedding \citep{su2024roformer}. 
\subsection{Proof of Theorem \ref{thm: nys} and Proposition \ref{thm: type1}: MLP Compression with Nystr\"om Approximation}

\begin{reptheorem}{thm: nys}[\textbf{MLP compression can be solved by Nystr\"om approximation}]
    Let $\hat{{\mW}}_U$ be searched over the matrix multiplication form ${\mW}_U\mS_k$, where $\mS_k$ is a $k$-column selection matrix, and let $\hat{\mW}_D$ be searched over $\mathbb{R}^{k\times d_h}$. The optimal $\hat{\mW}^*_D$ can then be expressed as: $(\mS_k^\top\mC_{\sigma}\mS_k)^\dagger\mS_k^\top\mC_{\sigma}\mW_D$. Using ${\mW}_U\mS_k$ and $\hat{\mW}^*_D$ as the compressed matrices, the Type-I reconstruction error in \eqref{obj} satisfies:
    \begin{align}
        V_{\text{I}} \leq \left\|\mW_D\right\|_{2}^2 \mathcal{E}^2_{\text{Nys}}(\mC_{\sigma}^{\frac{1}{2}}),
    \end{align}
 where $\mathcal{E}_{\text{Nys}}(\mC_{\sigma}^{\frac{1}{2}})$ denotes the Nyström approximation error, defined in Def. \ref{def:low_rank}, relative to the activation correlation matrix $\mC_{\sigma} \triangleq \sum_{i=1}^N \sigma(\mX_i {\mW_U})^\top \sigma(\mX_i {\mW_U})$, 
        using the same $\mS_k$ in the compression of ${{\mW}}_U$.
\end{reptheorem}

\begin{proof}
    Ideally, we want to seek low-rank matrices $\mW_U$, $\mW_D$ without any constraint; however, the nonlinearity between the two matrices makes the optimal solution intractable. To overcome the nonlinearity between the two matrices, we instead restrict the compressed up matrix $\hat{{\mW}}_U$ to be of the form $\mW_U\mS_k$,  $\mS_k\in\mathbb{R}^{d\times k}$ is a $k$-column selection matrix and $\hat{\mW}_D$ is a general $\mathbb{R}^{k\times d}$ matrix. Plug in this form, we can simplify \eqref{obj} as
    \begin{align}\label{obj2}
        &\min_{\mS_k, \hat{\mW}_D} \sum_{i=1}^N\|f(\mX_i)-\sigma(\mX_i\mW_U\mS_k)\hat{\mW}_D\|_F^2 \nonumber\\
        \overset{(a)}{=}~&\min_{\mS_k, \hat{\mW}_D} \sum_{i=1}^N\|\sigma(\mX_i\mW_U)\mW_D-\sigma(\mX_i\mW_U)\mS_k\hat{\mW}_D\|_F^2\nonumber\\
        =~&\min_{\mS_k, \hat{\mW}_D} \Tr\left(\sum_{i=1}^N\sigma(\mX_i\mW_U)^\top\sigma(\mX_i\mW_U)\left(\mW_D-\mS_k\hat{\mW}_D\right)\left(\mW_D-\mS_k\hat{\mW}_D\right)^\top\right)\nonumber\\
        =~&\min_{\mS_k, \hat{\mW}_D}\|\mC^{\frac{1}{2}}_{\sigma}\left(\mW_D-\mS_k\hat{\mW}_D\right)\|_F^2,
    \end{align}
    where $\mC_{\sigma}$ is the empirical correlation matrix of latent features
    $\mC_{\sigma}= \sum_{i=1}^N\sigma(\mX_i\mW_U)^\top\sigma(\mX_i\mW_U)$, and (a) follows from Assumption \ref{assump: nonlinear}.
    Setting the gradient of the last expression with respect to $\hat{\mW}_D$ to zero, we obtain the optimal down matrix $\hat{\mW}^*_D=\left(\mS_k^\top\mC_{\sigma}\mS_k\right)^\dagger\mS_k^\top\mC_{\sigma}\mW_D$.
    After Plugging this back to the objective, 
    we can further simplify the objective into,
    \begin{align}
        &\min_{\mS_k}\left\|\left(\mC^{\frac{1}{2}}_{\sigma}-\mC^{\frac{1}{2}}_{\sigma}\mS_k\left(\mS_k^\top\mC_{\sigma}\mS_k\right)^\dagger\mS_k^\top\mC_{\sigma}\right)\mW_D\right\|_F^2\nonumber\\
        \leq  ~&\left\|\mW_D\right\|_{2}^2\|\mC^{-\frac{1}{2}}_{\sigma}\|^2_2\min_{\mS_k} \left\| \mC_{\sigma}-\mC_{\sigma}\mS_k\left(\mS_k^\top\mC_{\sigma}\mS_k\right)^\dagger\mS_k^\top\mC_{\sigma}\right\|_F^2=\left\|\mW_D\right\|_{2}^2 \|\mC^{-1}_{\sigma}\|_2\mathcal{E}^2_{\text{Nys}}(\mC_{\sigma}). \label{type1_err}
    \end{align}
    Now, observe that the error on the right side of \eqref{type1_err} is proportional to the Nystr\"om matrix approximation to the matrix $\mC_{\sigma}$  in Definition \ref{def:low_rank}. Hence, the variable $\mS_k$ can be optimized with any Nystr\"om approximation algorithm \citep{gittens2013revisiting}. 
    In this work, we adapt a deterministic Nystr\"om algorithm, Algorithm \ref{alg:type1}, that has the theoretical guarantee proved in the next proposition.
\end{proof}

\begin{repproposition}{thm: type1}
     Suppose that the rank $k$ and the scores $s_i$ in Algorithm \ref{alg:type1} are chosen such that there exists an error $\varepsilon>0$ satisfying $\varepsilon \geq \sum_{i=k+1}^{d_{int}} s_i$, then the Type-I modular reconstruction error in \eqref{obj} is bounded by $V_I \leq \|\mW_D\|_2^2\|\mC^{-1}_{\sigma}\|_2\frac{\varepsilon^2{d^2_{int}}}{k^2(1-\varepsilon)^2}\sum_{i=k+1}^{d_{int}}\sigma_i^2(\mC_{\sigma})$, where $d_{\text{int}}$ and $\sigma_i$ denote the intermediate dimension (i.e., the input dimension of $W_D$) and singular values, respectively.
\end{repproposition}
\begin{proof}
    Since our column selection is equivalent to applying the deterministic ridge leverage score sampling (DRLS) to $\mC_{\sigma}^{\frac{1}{2}}$ \citep{mccurdy2018ridge} , Theorem 1 in \cite{mccurdy2018ridge} implies that
    \begin{align}
        &(1-\varepsilon) \mC_{\sigma}-\frac{\epsilon}{k}\left\|(\mC_{\sigma}^{\frac{1}{2}})_{\backslash k}\right\|_F^2 \mathbf{I} \preceq \mC_{\sigma}^{\frac{1}{2}}\mS_k \mS_k^\top \mC_{\sigma}^{\frac{1}{2}} \preceq \mC_{\sigma} \\ \Rightarrow~& 
        \mC_{\sigma}\preceq \frac{\epsilon}{k(1-\varepsilon)}\left\|(\mC_{\sigma}^{\frac{1}{2}})_{\backslash k}\right\|_F^2 \mathbf{I} + \frac{1}{1-\varepsilon} \mC_{\sigma}^{\frac{1}{2}}\mS_k \mS_k^\top \mC_{\sigma}^{\frac{1}{2}}.
    \end{align}
    Next, we define $\mP=\mC^{\frac{1}{2}}_{\sigma}\mS_k(\mS_k^\top\mC_{\sigma}\mS_k)^\dagger\mS_k^\top\mC^{\frac{1}{2}}_{\sigma}$. We note that $\mP$ is the projection matrix of the column space of $\mC^{\frac{1}{2}}_{\sigma}\mS$. Now, we multiply $\mI-\mP$ to both sides in the previous inequality to get
    \begin{align}
        &(\mI-\mP)\mC_{\sigma}(\mI-\mP) \\\preceq~& 
        \frac{\epsilon}{k(1-\varepsilon)}\left\|(\mC_{\sigma}^{\frac{1}{2}})_{\backslash k}\right\|_F^2 (\mI-\mP) + \frac{1}{1-\varepsilon}(\mI-\mP)  \mC_{\sigma}^{\frac{1}{2}}\mS_k \mS_k^\top \mC_{\sigma}^{\frac{1}{2}}(\mI-\mP)\\
        \preceq~&  \frac{\epsilon}{k(1-\varepsilon)}\left\|(\mC_{\sigma}^{\frac{1}{2}})_{\backslash k}\right\|_F^2 \mI, 
    \end{align}
    where in the last inequality we use the fact that $\mI-\mP \preceq \mI$ and that $\mI-\mP$ is the orthogonal projection to the orthogonal complement of the column space $\mC_{\sigma}^{\frac{1}{2}}\mS$ so that $\mS^\top \mC_{\sigma}^{\frac{1}{2}}(\mI-\mP)=\mathbf{0}$.
    Now, we have
    \begin{align}
       &\|(\mI-\mP)\mC^{\frac{1}{2}}_{\sigma}\mC^{\frac{1}{2}}_{\sigma}(\mI-\mP)\|_2 \leq  \frac{\varepsilon}{k(1-\varepsilon)}\left\|(\mC_{\sigma}^{\frac{1}{2}})_{\backslash k}\right\|_F^2 = \frac{\varepsilon}{k(1-\varepsilon)}\sum_{i=k+1}^{d_{\text{int}}}\sigma_i(\mC_{\sigma})\\
       \Rightarrow~& \|\mC^{\frac{1}{2}}_{\sigma}(\mI-\mP)^2\mC^{\frac{1}{2}}_{\sigma}\|_2 = \|\mC^{\frac{1}{2}}_{\sigma}(\mI-\mP)\mC^{\frac{1}{2}}_{\sigma}\|_2 \leq  \frac{\varepsilon}{k(1-\varepsilon)}\sum_{i=k+1}^{{d_{\text{int}}}}\sigma_i(\mC_{\sigma}).
    \end{align}
    Since $\mC^{\frac{1}{2}}_{\sigma}\mP\mC^{\frac{1}{2}}_{\sigma}=\mC_{\sigma}\mS_k\left(\mS_k^\top\mC_{\sigma}\mS_k\right)^\dagger\mS_k^\top\mC_{\sigma}$, the inequality is equivalent to 
    \begin{align}\label{eq:inter}
        \|\mC_{\sigma}-\mC_{\sigma}\mS\left(\mS^\top\mC_{\sigma}\mS\right)^\dagger\mS^\top\mC_{\sigma}\|_2\leq \frac{\varepsilon}{k(1-\varepsilon)}\sum_{i=k+1}^{d_{\text{int}}}\sigma_i(\mC_{\sigma}). 
    \end{align}
    Finally, we complete the proof by,
    \begin{align}
        \mathcal{E}_{\text{Nys}}^2(\mC_{\sigma})~&\overset{(a)}{\leq} {d_{\text{int}}}\|\mC_{\sigma}-\mC_{\sigma}\mS\left(\mS^\top\mC_{\sigma}\mS\right)^\dagger\mS^\top\mC_{\sigma}\|^2_2\nonumber\\
        ~&\overset{(b)}{\leq} \frac{\varepsilon^2{d_{int}}}{k^2(1-\varepsilon)^2}\left(\sum_{i=k+1}^{d_{int}}\sigma_i(\mC_{\sigma})\right)^2\nonumber\\
        ~&\overset{(c)}{\leq} \frac{\varepsilon^2{d^2_{int}}}{k^2(1-\varepsilon)^2}\sum_{i=k+1}^{d_{int}}\sigma_i^2(\mC_{\sigma}),
    \end{align}
    where  (a) follows from that$\|\mA\|_F\leq \sqrt{d}\|\mA\|_2$ for any matrix $\mA\in\mathbb{R}^{d\times d}$, (b) from \eqref{eq:inter}, and (c) from Cauchy inequality that $\left(\sum_{i=1}^n x_i\right)^2\leq n \sum_{i=1}^n x_i^2$ for any sequence $\{x_i\}_i$.
\end{proof}

\subsection{Proof of Theorem \ref{thm: cr} and Proposition \ref{thm: type2}: Key-Query Compression with CR Approximation}
\begin{reptheorem}{thm: cr}[\textbf{Key-Query compression can be solved by CR approximation}]
      Let the compressed $\hat{\mW}_Q$, $\hat{\mW}_K$ to be the form of $\mW_Q\mS_k, \mW_K\mS_k$, then Type-II reconstruction error in \eqref{obj} has 
    \begin{align}
        V_{\text{II}} \leq \mathcal{E}_{\text{CR}}^2(\mC^{\frac{1}{2}}_K\mC^{\frac{1}{2}}_Q),
    \end{align}
    where $\mathcal{E}_{\text{CR}}$ denotes the CR approximation error, defined in Def. \ref{def:low_rank}, relative to $\mC_Q^{1/2}\mC_K^{1/2}$, utilizing the same $\mS_k$ in the compression. Here, the matrices $\mC_Q \triangleq \sum_{i=1}^N \sigma(\mX_i \mW_Q)^\top \sigma(\mX_i \mW_Q)$ and $\mC_K \triangleq \sum_{i=1}^N \sigma(\mX_i \mW_K)^\top \sigma(\mX_i \mW_K)$ denote the correlation matrices of query and key states, respectively.
\end{reptheorem}
\begin{proof}
    Regarding two nonlinear functions satisfying Assumption \ref{assump: nonlinear}, we propose to optimize the reconstruction error with compressed key query matrices of the form $\mW_K\mS_k, \mW_Q\mS_k$, where $\mS_k$ is some column selection matrix. 
    Now the reconstruction error of this module is 
    \begin{align}
        &\sum_{i=1}^N\|f(\mX_i)-\sigma_r(\mX_i\mW_Q\mS_k)\sigma_r^\top(\mX_i\mW_K\mS_k)\|_F^2\nonumber\\
        \overset{(a)}{=}~& \sum_{i=1}^N\|\sigma_r(\mX_i\mW_Q)\left(\mI-\mS_k\mS_k^\top\right)\sigma^\top _r(\mX_i\mW_K)\|_F^2\nonumber\\
        =~&\sum_{i=1}^N\Tr\left(\left(\mI-\mS_k\mS_k^\top\right)\sigma_r(\mX_i\mW_Q)^\top\sigma_r(\mX_i\mW_Q)\left(\mI-\mS_k\mS_k^\top\right)\sigma_r(\mX_i\mW_K)^\top\sigma_r(\mX_i\mW_K)\right)\nonumber \\
        \overset{(b)}{\leq}~& \Tr\left(\sum_{i=1}^N\left(\mI-\mS_k\mS_k^\top\right)\sigma_r(\mX_i\mW_Q)^\top\sigma_r(\mX_i\mW_Q)\left(\mI-\mS_k\mS_k^\top\right)\sum_{j=1}^N\sigma_r(\mX_j\mW_K)^\top\sigma_r(\mX_j\mW_K)\right)\nonumber\\
        \overset{(c)}{\leq}~&\Tr\left(\mC_{K}\left(\mI-\mS_k\mS_k^\top\right)\mC_{Q}\right) = \|\mC_{K}^{\frac{1}{2}}\mC_{Q}^{\frac{1}{2}}-\mC_{K}^{\frac{1}{2}}\mS_k\mS_k^\top\mC_{Q}^{\frac{1}{2}}\|_F^2=\mathcal{E}_{\text{CR}}^2(\mC^{\frac{1}{2}}_K\mC^{\frac{1}{2}}_Q),
    \end{align}
    where $\mC_{K}=\sum_{i=1}^N\sigma(\mX_i\mW_Q\mS_k)^\top\sigma(\mX_i\mW_Q\mS_k)$, $\mC_{Q}=\sum_{i=1}^N\sigma(\mX_i\mW_K\mS_k)^\top\sigma(\mX_i\mW_K\mS_k)$ are the correlation matrices associated with the outputs of $\mW_Q$ and $\mW_K$, respectively. 
    Here, (a) follows from Assumption \ref{assump: nonlinear}, (b) follows from that $\left(\mI-\mS_k\mS_k^\top\right)\sigma_r(\mX_i\mW_Q)^\top\sigma_r(\mX_i\mW_Q)\left(\mI-\mS_k\mS_k^\top\right)$ and $\sigma_r(\mX_j\mW_K)^\top\sigma_r(\mX_j\mW_K)$ are positive semidefinite, and (c) follows from that $\mathbf{I}-\mS_k\mS_k^\top\preceq \mathbf{I}$.
    From the last expression, we observe that the reconstruction is bounded by the CR approximation \citep{drineas2006fast} to the matrix-product $\mC_{K}^{\frac{1}{2}}\mC_{Q}^{\frac{1}{2}}$.
\end{proof}

\begin{repproposition}{thm: type2}
    If we adopt Algorithm \ref{alg:type2} then Type-II modular reconstruction error is bounded by $V_\text{II}\leq \left(\frac{d_h-k}{d_h}\right)^2\left(\sum_{i=1}^{d_h}\sigma_i(\mC_{K})\right)\left(\sum_{i=1}^{d_h}\sigma_i(\mC_{Q})\right)$, where $\sigma_i$ denotes the singular values.
\end{repproposition}

\begin{proof} 
 Our Algorithm \ref{alg:type2} is a deterministic variant of \cite{drineas2006fast}.
Recall that
    \begin{align}
        \mathcal{E}_{\text{CR}}(\mC_{K}^{\frac{1}{2}}\mC_{Q}^{\frac{1}{2}}) = \|\mC_{k}^{\frac{1}{2}}\mC_{Q}^{\frac{1}{2}}-\mC_{K}^{\frac{1}{2}}\mS_k\mS_k^\top\mC_{Q}^{\frac{1}{2}}\|_F=\|\sum_{i=k+1}^d \vk_i \vq_i^\top\|_F,
    \end{align}
    where $\vk_i$ and $\vq_i$ are the $i$-th column and $i$-th row of $\mC_{k}^{\frac{1}{2}}$ and $\mC_{q}^{\frac{1}{2}}$, respectively. Then,
    \begin{align}
        \|\sum_{i=k+1}^d \vk_i \vq_i^\top\|_F\leq~ &\sum_{i=k+1}^d\|\vk_i\|_2\|\vq_i\|_2\overset{(a)}{\leq}~ \sqrt{\left(\sum_{i=k+1}^d\|\vk_i\|_2^2\right)\left(\sum_{i=k+1}^d\|\vq_i\|_2^2\right)} \\
        \overset{(b)}{\leq}~ & \frac{d-k}{d}\sqrt{\left(\sum_{i=1}^d\|\vk_i\|_2^2\right)\left(\sum_{i=1}^d\|\vq_i\|_2^2\right)} =~\frac{d-k}{d}\|\mC_{K}^{\frac{1}{2}}\|_F\|\mC_{Q}^{\frac{1}{2}}\|_F\\
        =&~  \frac{d-k}{d}\sqrt{\left(\sum_{i=1}^d\sigma_i(\mC_{K})\right)\left(\sum_{i=1}^d\sigma_i(\mC_{Q})\right)},
    \end{align}
    where in (a) we use Cauchy-Schwartz inequality and in (b) we use the fact that the column selection is based on the norm product in Algorithm \ref{alg:type2}.
\end{proof}

\subsection{Proof of Theorem \ref{thm: svd} and Proposition \ref{thm: type3}: Value-Output Compression with SVD}

\begin{reptheorem}{thm: svd}[\textbf{Value-Output compression can be solved by SVD}]
       If we search $\hat{\mW}_V$ and $\hat{\mW}_O$ over $\mathbb{R}^{d_h \times k}$ and $\mathbb{R}^{k \times d_h}$, respectively, the optimum in \eqref{obj} is $\hat{\mW}_V = \mC^{-\frac{1}{2}}\mU_k$ and $\hat{\mW}_O = \bm{\Sigma}\mV^\top$. Here, $\mU\bm{\Sigma}\mV^\top$ and  $\mC\triangleq \sum_{i=1}^N\mX_i^\top\mX_i$ are the SVD of $\mC^{\frac{1}{2}}\mW_V\mW_O$ and input correlation matrix, respectively. The corresponding Type-III reconstruction error in \eqref{obj} is the SVD approximation error, defined in Def. \ref{def:low_rank}, relative to $\mC^{\frac{1}{2}}\mW_V\mW_O$:
    \begin{align}
        V_{\text{III}} = \mathcal{E}^2_{\text{SVD}}(\mC^{\frac{1}{2}}\mW_V\mW_O).
    \end{align}
\end{reptheorem}
\begin{proof}
    $\hat{\mW}_V\in\mathbb{R}^{d\times k}$ and $\hat{\mW}_O\in\mathbb{R}^{k\times d}$. Plug in $\hat{f}(\mX)=\mX \hat{\mW}_V \hat{\mW}_O$ into \eqref{obj} and simplify yields the objective
    \begin{align}\label{obj1}
        &\min_{\hat{\mW}_V,\hat{\mW}_O}\sum_{i=1}^N\Tr \left(\mX_i^\top\mX_i({\mW_V} {\mW_O}-\hat{\mW}_V \hat{\mW}_O)({\mW_V} {\mW_O}-\hat{\mW}_V \hat{\mW}_O)^\top\right)\nonumber\\
        =~&\min_{\hat{\mW}_V,\hat{\mW}_O}\|\mC^{\frac{1}{2}} {\mW_V} {\mW_O}-\mC^{\frac{1}{2}}\hat{\mW}_V \hat{\mW}_O\|_F^2=\mathcal{E}^2_{\text{SVD}}(\mC^{\frac{1}{2}}\mW_V\mW_O),
    \end{align}
    where $\mC = \sum_{i=1}^N \mX_i^\top\mX_i$ is the input correlation matrix. 
\end{proof}

\begin{repproposition}{thm: type3}
   Denote $\sigma_i$ as the singular values, Algorithm \ref{alg:type3} yields the optimal Type-III modular reconstruction error $ V_\text{III} = \sum_{i=k+1}^{d}\sigma_i^2(\mC^{\frac{1}{2}} {\mW_V} {\mW_O})$.
\end{repproposition}

\begin{proof}
    As $\mC^{\frac{1}{2}}\hat{\mW}_V\hat{\mW}_O$ has low rank $k$, this reconstruction error is upper bounded by the residue of the spectrum of the matrix $\mC^{\frac{1}{2}}\mW_V\mW_O$, i.e.,
    $\mathcal{E}_{\text{SVD}}\leq\sqrt{\sum_{i=k+1}^{d}\sigma_i^2(\mC^{\frac{1}{2}} {\mW_V} {\mW_O})}$.
    In fact, the upper bound is achievable by Algorithm \ref{alg:type2} since $\mC^{\frac{1}{2}}\hat{\mW}_V\hat{\mW}_O=\mU_k\bm{\Sigma}_k\mV_k^\top$, which is the optimal rank $k$ approximation to the matrix $\mC^{\frac{1}{2}}\mW_V\mW_O$.
\end{proof}

\subsection{Proof of Theorem \ref{thm:global}: Global Sparsity Allocation}
\begin{reptheorem}{thm:global}
    For sufficient large $\varepsilon$, (\ref{eq: softmax}) is the optimal sparsity allocation in the \eqref{global rank eq}.
\end{reptheorem}
\begin{proof}
{
Consider the relaxed optimization problem 
    \begin{align}
        \max_{\phi_{1:L}}\sum_{i=1}^L s_i (1-\phi_i) +\varepsilon H(\phi_i)\quad \text{s.t. } \frac{1}{L}\sum_{i=1}^L \phi_i = \phi_\text{avg}.
    \end{align}
    Its associated Lagrangian  is 
    \begin{align}
        \mathcal{L}(\phi_{1:L},\lambda) &= \sum_{i=1}^L s_i (1-\phi_i) +\varepsilon H(\phi_i) + \lambda \left(\frac{1}{L}\sum_{i=1}^L \phi_i - \phi_\text{avg}\right).
    \end{align}
    To find the optimum, we set the gradient of the Lagrangian to zero, which yields
    \begin{align}
        0=\nabla_\phi \mathcal{L}(\phi_{1:L},\lambda)&=\nabla_\phi\left(\sum_{i=1}^L s_i (1-\phi_i) -\varepsilon H(\phi_i)i\right) + \lambda \nabla_\phi\left(\frac{1}{L}\sum_{i=1}^L \phi_i - \phi_\text{avg}\right)\\
        &=\nabla_\phi\left(\sum_{i=1}^L s_i (1-\phi_i) -\varepsilon \sum_{i=1}^L\phi_i\log\phi_i\right) + \lambda \nabla_\phi\left(\frac{1}{L}\sum_{i=1}^L \phi_i - \phi_\text{avg}\right).
    \end{align}
This is equivalent to that, for any $j=1,\dots,L$,
\begin{align}
    0&=\partial_{\phi_j}\left(\sum_{i=1}^L s_i (1-\phi_i) -\varepsilon \sum_{i=1}^L\phi_i\log\phi_i\right) + \lambda \partial_{\phi_j}\left(\frac{1}{L}\sum_{i=1}^L \phi_i - \phi_\text{avg}\right)\\
    &= -s_j -\varepsilon \log \phi_j-\varepsilon +\lambda \frac{1}{L}.
    \end{align}
    After rearrangement, we have $\phi_j=C\exp(-s_j/\varepsilon)$ for some constant $C$. On the other hand, $\phi_j$ satisfies the constraint of $\frac{1}{L}\sum_{i=1}^L \phi_i = \phi_\text{avg}$, which implies
    \begin{align}
       &\sum_{j=1}^L C\exp(-s_j/\varepsilon) = L\phi_\text{avg}\\
       \Rightarrow \quad & C = L\phi_\text{avg} /\sum_{j=1}^L\exp(-s_j/\varepsilon)\\
       \Rightarrow \quad &\phi_i = L\phi_\text{avg}\exp(-s_i/\varepsilon)/\sum_{j=1}^L\exp(-s_j/\varepsilon).
    \end{align}
    Finally, we must verify that for any $i=1,\dots,L$, the above expression of $\phi_i$ is a valid sparsity allocation, satisfying $\phi_i\leq 1$ for sufficiently large $\varepsilon$, to ensure it is also the optimum solution to the original optimization problem in \eqref{global rank eq}.
    Since 
    $ \phi_i = L\phi_\text{avg}\exp(-s_i/\varepsilon)/\sum_{j=1}^L\exp(-s_j/\varepsilon)$ is a continuous function of $\varepsilon$ and 
    $\lim\limits_{\varepsilon \rightarrow \infty} \phi_i = \phi_\text{avg} < 1$, there must exist some constant $N_i$ such that when $\varepsilon\geq N_i$, $\phi_i$ is less than 1.
    Hence, the sparsity allocation is a valid optimal solution to \eqref{global rank eq} if $\varepsilon > \max(N_1, \dots, N_L)$, completing the proof.
    }    
\end{proof}
\newpage

\section{\textsc{Additional Experiments}}
\label{supp:exp}

\subsection{{Modified Algorithms for Grouped Query Attention}}
\label{supp:gqa}
Some modern LLMs such as Gemma \citep{team2024gemma}, Llama 3 \citep{llama3modelcard} utilize a shared key-query strategy to improve inference efficiency. 
This design adopts the grouped-query attention (GQA) mechanism \citep{ainslie2023gqa} that couples multiple queries and output matrices with shared keys and values, so compressed key and value matrices also have the constraints of sharing across different heads. Its mechanism is illustrated as follows.
\begin{align}
    \text{(GQA)~~}\sum_{i=1}^{H/G}\sum_{j \in G_i}\text{Softmax}(\underbrace{\sigma_{r}(\mX\mW_Q^j)\sigma_{r}^\top(\mX\mW_K^i)}_{\text{{\color{RubineRed}Type-II}}})\underbrace{\mX\mW_V^i{\mW_O^j}}_{\text{{\color{CornflowerBlue}Type-III}}},
\end{align}
\begin{wrapfigure}[20]{r}{0.5\textwidth}
    \centering 
    \includegraphics[width=1\textwidth]{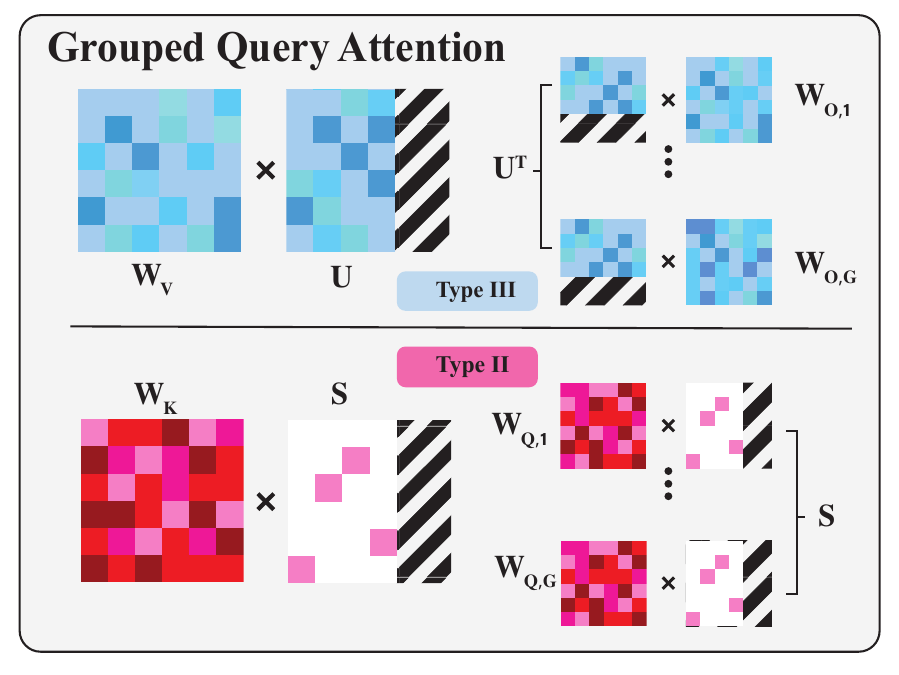}
    \caption{\footnotesize{\textbf{Illustration of Type-II and Type-III modifications in GQA.} In Type-I, the index selection matrix $\mS$ is shared among different key projection matrices in the same group. Similarly, in Type-II, the eigenmatrix $\mU$ is shared among different output matrices within the same group.}\label{conv_rm}}
\end{wrapfigure}
where $G_i$ denotes the set of each group and $G=|G_i|$ is each of its size. We see that our compressed $\mW_V^i$, $\mW_K^j$ must be jointly optimized within each group. To address it, we modify Algorithm \ref{alg:type2}, \ref{alg:type3} by using projection strategies. 
In line 3 of Algorithm \ref{alg:type2} for Type-II module, we calculate the group score equal to the square root of sum of the scores of each head within a group, i.e., 
$s_i=\sqrt{\sum_{h\in G}\|\mC_{h,Q}^{\frac{1}{2}}[:,i]\|^2\|\mC_{h,K}^{\frac{1}{2}}[:,i]\|^2}$, where $h$ indicates the head. By doing in this way, we ensure that the column selection matrix for compressions remains equal within the group. For grouped Type-III modification, in line 3 of Algorithm \ref{alg:type3}, we calculate the SVD of $\mC\mW_V=\mU\bm{\Sigma}\mV^\top$ and skip the calculation of $\mW_O'$ and the second SVD and then outputs $\hat{\mW}_V = \mW_V\mU_k$, $\hat{\mW}_{O,j}=\mU_k^\top \mW_{O,j}, \forall j\in G_i$. Since $\mW_V$ is shared within a group, this ensures that the compressed $\hat{\mW}_V$ is also shared. In Table \ref{table: llama3}, we apply this modification to a Llama-3 8B compression. 

\subsection{Implementation Details}
\label{supp:impl}
\paragraph{Setup}
We utilize the HuggingFace generation library \citep{wolf2019huggingface} to implement our LLM models and adapt the SliceGPT \citep{ashkboos2024slicegpt} GitHub repository for correlation matrix estimations. All compression experiments were conducted on a single NVIDIA A100 80GB GPU, except for the 70B model compressions, which utilized 8 A100 GPUs. The models use the FP16 data format. Unless otherwise specified, the calibration set consists of a random sample of 128 sequences, each of length 2048, from WikiText-2, following the common practice in the literature \cite{ashkboos2024slicegpt,van2023llm}.

\paragraph{Datasets}
We consider multiple tasks in
LM Evaluation Harness \citep{gao2021framework}, including ARC-e, ARC-c \citep{clark2018think}, PIQA \citep{bisk2020piqa}, WinoGrande \citep{sakaguchi2021winogrande}, and HellaSwag \citep{zellers2019hellaswag}, OpenBookQA \citep{mihaylov2018can}, MathQA \citep{amini2019mathqa}, BoolQ \citep{clark2019boolq}, COPA \citep{roemmele2011choice}, MMLU \citep{hendrycks2020measuring}, and LAMBADA \citep{paperno2016lambada}.

\paragraph{Conversion of Layernorm to RMSNorm}
We adapt SliceGPT \citep{ashkboos2024slicegpt} 
official code to implement our compression. As shown in the work, this conversion is an invariant transformation that preserves the model output. SliceGPT uses this transformation and 
the orthogonal invariance property of RMSNorm
to slice the weight matrices. On the other hand, \locogpt does not reply on the invariance property. We use the transformation simply for easy adaptation from SliceGPT by avoiding building all from the scratch. A side benefit is that our work is compatible with SliceGPT where a slicing and our compression can be applied independently. Although our experiments on OPT and \textsc{llama} do not find clear improvement when incorporating the two (see Table \ref{tbl:slice+loco}), it might be beneficial for some other LLMs.

\paragraph{Correlation Matrix Estimations}
Our algorithms utilize various input correlation matrices as detailed in Algorithms \ref{alg:type1}, \ref{alg:type2}, and \ref{alg:type3}. Following the approach used in SliceGPT \citep{ashkboos2024slicegpt}, we employ the Catcher function to gather empirical data from the calibration set. For matrix decomposition, we upcast correlation matrices from FP16 to FP64 and then downcast the decomposed weights back to FP16. Our process sequentially compresses weights across all layers, mirroring SliceGPT’s method. Additionally, our approach is adaptable to parallel structural models like Phi-2, showcasing flexibility similar to that demonstrated by SliceGPT.

\paragraph{Matrix Operations}
We utilize {\fontfamily{qcr}\selectfont torch.svd} and {\fontfamily{qcr}\selectfont torch.pinv} in PyTorch for performing Singular Value Decomposition (SVD) and computing the Moore-Penrose inverse on tensors of dtype FP64.

\paragraph{MLP Module}
Algorithm \ref{alg:type1} requires a ridge leverage score parameter $\lambda$. We find that the results are largely insensitive to this parameter; therefore, we simply use $\lambda=1$ across all experiments.

\paragraph{Key-Query Module}
\locogpt reduces the feature dimension in the key-query module. 
In our current setup, we store head-dependent index selections, which specify the rows of cosine and sine matrices in the rotary embedding, using only $O(d_h)$ INT8 numbers, together with the reduced dimension matrices. We’ve observed that this method may slow down generation speed at compression rates below 10\%. A feasible modification is zeroing out columns associated with pruned indices; however, this increases the memory footprint because the matrices stored do not undergo dimension reduction. We think there is potential for improvements through enhanced engineering efforts that could better optimize the balance between memory savings and generation speed.

\paragraph{Value-Output Module}
Lines 3-5 in Algorithm \ref{alg:type3} provide a more computationally efficient implementation of the SVD of $\mC^{\frac{1}{2}}\mW_{V,j}\mW_{O,j}$. Since $\mW_{V,j}$ and $\mW_{O,j}$ are thin matrices, applying SVD directly on their product incurs $O(d_h^3)$ complexity, while applying SVD to them sequentially incurs only $O(d_h \times (d_h/H)^2)$ computations.

\paragraph{Global Sparsity Allocation}
To allocate global sparsity, we first calculate the BI scores with a single forward pass on the calibration set. We then set the sparsity according to a chosen temperature $\varepsilon$, as detailed in Section \ref{sec:global_sparsity}. 
A high $\varepsilon$ leads to a very uniform allocation, while a low value introduces excessive sparsity in some layers.
Empirically, we find that a simple rule of thumb is to choose a temperature $\varepsilon$ that results in maximal layer sparsity around 80\%. 

\paragraph{Throughput Benchmark}
We use the official SliceGPT \citep{ashkboos2024slicegpt} codebase to benchmark the throughput of all methods, with both sequence length and batch size set to 256 and utilizing KVCache.

\subsection{Additional Generation and Zero-Shot Experiments}
\label{supp:add_gen}
\begin{table}[htp!]
\centering
    \caption{{Perplexities of none gradient-based structured compression methods on WikiText-2.}\label{tbl:opt}}
\setlength\extrarowheight{2pt}
\resizebox{1\linewidth}{!}{
\begin{tabular}{c|c|cccc|c|c}
\midrule[1pt]
\multirow{2}{*}{\begin{tabular}[c]{@{}c@{}}\textbf{Method} \end{tabular}}& \multirow{2}{*}{\begin{tabular}[c]{@{}c@{}}\textbf{Compression} \end{tabular}}  &\multicolumn{4}{c|}{\textbf{OPT}}  & \textbf{\textsc{Llama-2}} &  \multirow{2}{*}{\begin{tabular}[c]{@{}c@{}}\textbf{Complexity} \end{tabular}}  \\
\multicolumn{1}{c|}{}                                   &                                         & 125M   & 1.3B    & 2.7B  & 6.7B    & 7B               \\ \midrule[1pt]
\multicolumn{1}{c|}{{Dense}}                   & 0\%                                  & 27.65  & 14.62   & 12.47 & 10.86   & 5.12  & -           \\ \midrule
\multicolumn{1}{c|}{{Sparse GPT 2:4} \citep{frantar2023sparsegpt}}                   & 50\%                                  & 45.07 & 29.61   & 14.90 & 13.00   & 8.69  & $O(d_{\text{hidden}}^3)$          \\ \midrule
                                                      & 10\%                                     & 767.2  & 894.4   & 1229  & 3464    & 861.76           \\
                                                      
\multicolumn{1}{c|}{{Magnitude}}               & 20\%                                & 4685   & 1278    & 2788  & 16747   & 821.34    & $O(d_{\text{hidden}}^2)$       \\
                                                      & 30\%                                      & 17970  & 3098    & 9255  & 17312   & 9623          \\ \midrule
                                                      & 10\%                                       & 36.29  & 68.36   &   20.82    & 357.61  &        n/a  
                                                      \\
\multicolumn{1}{c|}{{SVD}}                     & 20\%                                & 55.48  & 1023.49 &  50.01     & 2387.39 &       n/a 
& $O(d_{\text{hidden}}^3)$       \\
\multicolumn{1}{c|}{\textbf{}}                        & 30\%                                       & 173.77 & 8851.45 &   707.17    & 9448.12 &      52719            \\ \midrule
                                                      & 10\%                                       & 33.3   & 20.76   & 17.69 & 27.2    & 14259            \\
\multicolumn{1}{c|}{{OBD} \citep{lecun1989optimal}}                     & 20\%                            & 94.14  & 1392    & 3236  & 7570    & 15630      &       $O(Td_{\text{hidden}}^3)$       \\
                                                      & 30\%                                       & 545.6  & 2147    & 7233  & 7628    & 21386            \\ \midrule
                                                      & 10\%                                       & 34.48  & 16.58   &   13.86    & 11.6    &   6.46               \\
                                                      & 20\%                                       & 42.87  & 19.15   &  15.86     & 12.62   &  7.68                \\
\multicolumn{1}{c|}{{SliceGPT}
\citep{ashkboos2024slicegpt}}               & 30\%                           & 59.87  & 23.87   &   19.91    & 14.19   & 10.47        &       $O(d_{\text{hidden}}^3)$     \\
                                                      & 40\%                                       & 102.41 & 36.2    &   30.77    & 17.99   & 15.19            \\
                                                      & 50\%                                       & 185.52 & 66.12   &  56.99     & 26.72   &    24.82              \\ \midrule
                                                      & 10\%                                       & 28.06  & 15.03   &  12.78     & 11.17   & 5.48             \\
                                                      & 20\%                                       & 29.62  & 15.98   &  13.56     & 11.79   & 6.16             \\
\multicolumn{1}{c|}{{\locogpt}}                 & 30\%                               & 33.27  & 17.91   &   14.71    & 12.67   & 7.51         &       $O(d_{\text{hidden}}^3)$     \\
\multicolumn{1}{c|}{(ours)} 
                                                      & 40\%                                       & 38.37  & 21.92   &   17.43    & 14.79   & 8.41             \\
                                                      & 50\%                                       & 51.81  & 32.67   &   24.75    & 20.39   & 11.88            \\ \midrule[1pt]
\end{tabular}
}
\end{table}

\paragraph{Generation Performance}

In Table \ref{tbl:opt}, we compare the perplexity of compressed OPT and \textsc{Llama}-2 7B models on WikiText-2 with other baselines that do not use gradient information. The rightmost column indicates their computational complexity per layer. We observe that \locogpt performs the best among all structured compression methods, and the 30-40\% compressed \locogpt models outperform the 2:4 SparseGPT. Notably, our method shows better performance in \textsc{Llama} than in OPT models. We suspect this is due to the higher nonlinearity, such as RoPE and Gated MLP, adopted in \textsc{Llama}, and that our method favors more nonlinear structures. This table also shows that our compression is effective on small language models.

\paragraph{Zero-Shot Task Performance}

\begin{table}[t!]
\centering
    \caption{Downstream zero-shot task performance of \textsc{Llama}-2 7B calibrated with 128 samples from WikiText2.\label{tbl:zero_shot_full}}
\setlength\extrarowheight{2pt}
\resizebox{1\linewidth}{!}{
\begin{tabular}{c|c|ccccc|c}
\midrule[1pt]
\textbf{Compression}                          & \textbf{Method} & \textbf{ARC-e}                & \textbf{ARC-c}                & \textbf{PIQA}                 & \textbf{WinoGrande}           & \textbf{HellaSwag}             & \textbf{Average}                 \\ \midrule[1pt]
0\%                        & \multicolumn{1}{l|}{Dense}              &  74.58\%                   &        46.25\%         &        79.11\%          &            69.06\%         &     75.99\%                &     69.00\%               \\ 
\midrule\multirow{5}{*}{20\%}    & \multicolumn{1}{l|}{ShortGPT \citep{men2024shortgpt}}             &          58.33\%        &     38.05\%                &  72.58\%              &      65.51\%            & 65.27\%               &       
59.95\%           \\
                                &\multicolumn{1}{l|}{SliceGPT \citep{ashkboos2024slicegpt}}               &    51.47\%              &        31.06\%            &      64.25\%              &         62.74\%            &              49.78\%          &        51.86\%                \\
                                & \multicolumn{1}{l|}{LLM surgeon \citep{van2023llm}}            & \underline{71.36}\%        &     41.89\%                &   \underline{77.09}\%              &      66.30\%            & \underline{71.30}\%               &       \underline{65.59}\%  \\ 
                                \rowcolor{blue!10!white} 
                                & \multicolumn{1}{l|}{\locogpt (ours)} & 69.07\%                 &      \underline{42.06}\%             &   74.05\%               &  \underline{68.03}\%                     &           69.05\%          &  64.46\%  \\
                                \rowcolor{blue!20!white} 
                                & \multicolumn{1}{l|}{\locogpt-Alpaca (ours)} & 71.71\%                 &      41.89\%             &   76.22\%               &  68.19\%                     &           69.59\%          &  65.52\%  \\
                                 \midrule
\multirow{5}{*}{30\%}       &\multicolumn{1}{l|}{ShortGPT \citep{men2024shortgpt}}            &   48.65\%                & 32.85\%                &   64.31\%                &                 64.33\%  &                     56.13\%   &       53.25\%            \\
                                & \multicolumn{1}{l|}{SliceGPT \citep{ashkboos2024slicegpt}}             &    44.44\%              &    29.27\%               &    57.56\%              &     58.48\%                &    41.08\%                   &        46.17\%            \\
                                & \multicolumn{1}{l|}{LLM surgeon \citep{van2023llm}}            &  63.09\%                & 36.69\%                &    \underline{73.56}\%                &                 61.09\%  &                     60.72\%   &       59.03\%  \\
                                \rowcolor{blue!10!white} 
                                 & \multicolumn{1}{l|}{\locogpt (ours)} & \underline{63.26}\%& \underline{38.73}\% & 70.40\%&\underline{67.32}\% & \underline{63.26}\% & \underline{60.78}\%  \\
                                 \rowcolor{blue!20!white} 
                                & \multicolumn{1}{l|}{\locogpt-Alpaca (ours)} & 65.49\%                 &      39.16\%             &   73.34\%               &  66.22\%                     &           65.90\%          &  62.02\%  \\
                                \midrule

\multirow{5}{*}{40\%}    &\multicolumn{1}{l|}{ShortGPT \citep{men2024shortgpt}}            &    41.16\% &29.94\% & 60.12\% &60.46\%& 43.67\% &  47.07\%\\
                                & \multicolumn{1}{l|}{SliceGPT \citep{ashkboos2024slicegpt}}             &     36.49\% &24.57\% & 54.90\% &53.43\%& 34.80\% &40.84\%          \\
                                & \multicolumn{1}{l|}{LLM surgeon \citep{van2023llm}}           & \underline{52.31}\% &\underline{30.29}\% & \underline{69.26}\% &54.38\%& 48.04\% &  50.86\%  \\
                                \rowcolor{blue!10!white} 
                                 & \multicolumn{1}{l|}{\locogpt (ours)} &  49.45\%                & 30.03\%                   &64.96\%                     &\underline{61.96}\%                      &      \underline{53.01}\%              &  \underline{51.88}\%    \\
                                 \rowcolor{blue!20!white} 
                                & \multicolumn{1}{l|}{\locogpt-Alpaca (ours)} & 59.76\%                 &      34.73\%             &   70.35\%               &  64.40\%                     &           58.63\%          &  57.58\%  \\
                                \midrule[1pt]
\end{tabular}
}
\end{table}
\begin{table}[t!]
\centering
    \caption{{Downstream zero-shot task performance of \textsc{Llama}-2 13B calibrated with 128 samples from WikiText2.}\label{tbl:13b_ppl}}
\setlength\extrarowheight{2pt}
\resizebox{1\linewidth}{!}{
\begin{tabular}{c|c|ccccc|c}
\midrule[1pt]
\textbf{Method}                          & \textbf{Compression} & \textbf{ARC-e}                & \textbf{ARC-c}                & \textbf{PIQA}                 & \textbf{WinoGrande}           & \textbf{HellaSwag}             & \textbf{Average}                 \\ \midrule[1pt]
Dense                           & 0\%              &                  77.48\%    &           49.23\%          &     80.47\%                &               72.22\%        &            79.39\%           &        71.76\%              \\ \midrule
\multirow{3}{*}{SliceGPT \citep{ashkboos2024slicegpt}}       & 20\%           &  55.81\%                   &   35.84\%                    &         65.83\%             &              67.17\%        &       53.58\%                  &      55.65\%              \\
                                & 30\%             &    45.96\%                  & 30.80\%                     &     59.63\%                 &          61.80\%            &       44.09\%                &      48.46\%                \\
                                & 40\%             & 38.59\% & 27.05\% &55.98\%  &56.51\% & 37.15\% & 43.06\% \\ \midrule
\multirow{3}{*}{\locogpt (ours)} & 20\%             &             74.07\%         &     46.16\%                 &     74.53\%                 &     70.32\%                    &                68.96\%       &      66.81\%                  \\
                                & 30\%             & 71.93\% & 43.60\% & 73.94\% &  71.90\%&68.21\% & 65.92\%\\
                                & 40\%             &  62.88\%                 &        38.40\%             &       69.10\%               &    67.72\%                  &  58.27\%                   &   59.27\%                   \\ \midrule[1pt]
\end{tabular}
}
\end{table}

\begin{table}[t!]
\centering
    \caption{Downstream zero-shot task performance of \textsc{Llama}-2 13B calibrated with 128 samples from Alpaca.\label{tbl:alpaca13b}}
\setlength\extrarowheight{2pt}
\resizebox{1\linewidth}{!}{
\begin{tabular}{c|c|ccccc|c}
\midrule[1pt]
\textbf{Method}                          & \textbf{Compression} & \textbf{ARC-e}                & \textbf{ARC-c}                & \textbf{PIQA}                 & \textbf{WinoGrande}           & \textbf{HellaSwag}             & \textbf{Average}                 \\ \midrule[1pt]
Dense                           & 0\%              &  77.48\%                   &        49.23\%         &        80.47\%          &            72.22\%         &     79.39\%                &     71.76\%               \\ \midrule
\multirow{3}{*}{SliceGPT \citep{ashkboos2024slicegpt}}       & 20\%            &       69.36\%            &    40.70\%                   &       74.97\%          &       65.67\%               &      61.01\%                   &  62.34\%                \\
                                & 30\%             &          60.27\%         &   36.18\%                 &          69.42\%         &         64.09\%             & 49.74\%                       &    55.94\%                 \\
                                & 40\%             & 48.99\%  & 32.51\% & 63.17\%  &56.75\% &  39.85\% & 48.25\% \\ \midrule
\multirow{3}{*}{\locogpt (ours)} & 20\%             &           74.24\%         &   45.90\%                 &       78.24\%            &            72.53\%            &          75.78\%           &    69.34\%                    \\
                                & 30\%             & 70.24\%& 41.47\% & 77.15\%&71.27\% & 71.84\% & 66.39\%\\
                                & 40\%             &  63.72\%                & 38.82\%                   &71.87\%                     &66.30\%                      &      62.10\%              &  60.56\%                   \\ \midrule[1pt]
\end{tabular}
}
\end{table}

In Table \ref{tbl:zero_shot_full}, we report the zero-shot performance of \textsc{Llama}-2 7B, calibrated with WikiText-2 and the Alpaca dataset, across various compression rates. We observe that \locogpt outperforms LLM Surgeon as the compression rate increases, and the benefits of using the Alpaca dataset also grow with higher compression rates. Notably, while ShortGPT performs poorly in generation tasks, it significantly outperforms SliceGPT in zero-shot tasks. Both LLM Surgeon and \locogpt maintain high performance in generation and zero-shot tasks, but our method requires only 3\% of the computational resources compared to LLM Surgeon.

We also test the performance on \textsc{Llama}-2 13B using the WikiText-2 calibration set, as shown in Table \ref{tbl:13b_ppl}. Similar to the 7B model, our method excels at higher compression rates (above 20\%). However, at a 40\% compression rate, we notice a performance drop in the HellaSwag task compared to the 30\% compression, likely due to inherent biases in our method. Nevertheless, with calibration from the Alpaca dataset, as shown in Table \ref{tbl:alpaca13b}, our method achieves high performance at 20\% and 30\% compression. Addressing these inherent biases and enhancing performance on the HellaSwag task is a promising area for future research.

\begin{table}[!htbp]
\centering
    \caption{Downstream zero-shot task performance of \textsc{Llama}-3 8B calibrated with 128 samples from Alpaca.}
    \label{table: llama3}
\setlength\extrarowheight{2pt}
\resizebox{1\linewidth}{!}{
\begin{tabular}{c|c|c|ccccc|c}
\midrule[1pt]
\textbf{Method}                          & \textbf{Compression}&\textbf{Perplexity } {\color{Red}$\downarrow$} & \textbf{ARC-e}                & \textbf{ARC-c}                & \textbf{PIQA}                 & \textbf{WinoGrande}           & \textbf{HellaSwag}             & \textbf{Average}                 \\ \midrule[1pt]
Dense                           & 0\%     &   2.98      &  77.69\%                   &        53.58\%         &        80.63\%          &            72.69\%         &     79.16\%                &     72.75\%               \\ \midrule
\multirow{2}{*}{ShortGPT \citep{men2024shortgpt}}       & 25\%  &   282.56       &       38.13\%            &    31.40\%                   &       60.94\%          &       54.22\%               &      31.52\%                   &  43.24\%                \\
                                & 30\%  &     659.33      &          36.83\%         &   30.72\%                 &          58.98\%         &         54.62\%             & 29.08\%                       &    42.04\%                 \\ \midrule
\multirow{2}{*}{SliceGPT \citep{ashkboos2024slicegpt}}       & 25\%  &  3.87        &       58.88\%            &    33.36\%                   &       68.55\%          &       58.01\%               &      49.86\%                   &  53.73\%                \\
                                & 30\%  &  4.52         &          52.02\%         &   29.18\%                 &          64.85\%         &         54.62\%             & 41.38\%                       &    48.41\%                 \\ \midrule
\multirow{2}{*}{\locogpt (ours)} & 25\%   &     3.52     &           67.05\%         &   41.13\%                 &       75.52\%            &            69.61\%            &          66.49\%           &    63.96\%                    \\
                                & 30\%    &    3.80    & 62.75\%& 38.65\% & 73.61\%&67.25\% & 62.10\% & 60.87\%\\
                  \midrule[1pt]
\end{tabular}
}
\end{table}

In Table \ref{table: llama3}, we test our method on \textsc{Llama}-3 8B using our modified algorithm tailored for grouped query attention. As this is a relatively new model, we could only compare results with SliceGPT and ShortGPT, which already support this model. We observe that compression has a more significant impact on performance degradation compared to \textsc{Llama}-2 for all tested methods. We believe this is due to denser information encoding in each parameter, making the model more sensitive to weight changes. However, \locogpt maintains approximately 90\% performance with 25\% compression. Another interesting observation is that the performance order of ShortGPT and SliceGPT is reversed for \textsc{Llama}-3 compared to \textsc{Llama}-2, with ShortGPT's layer pruning strategy performing much worse. This suggests fundamental differences between the two models and indicates that \textsc{Llama}-3 may require more sophisticated compression techniques.

\begin{table}[!htbp]
\centering
    \caption{Generation samples of \textsc{Llama-2} 7B with $30\%$ compression.\label{tbl:gen_sample}}
\setlength\extrarowheight{2pt}
\resizebox{1\linewidth}{!}{

\begin{tabular}{l|l}
\midrule[1pt]
         & \textbf{Prompt:} The universe is the entirety of space, time, matter, and energy that exists.                                                                                                                                                                                                                                                                                                                                                                                                                                                                                                                                                                                                                                                                                                                                                                                                                                                                                                                                                                            \\ \midrule[1pt]
\textbf{ShortGPT} \citep{men2024shortgpt} &   
 \begin{tabular}[c]{@{}l@{}}
sierpę 2018.
The universe is everything that ever did or ever will exist. It includes all the past, all the \\present, and all the future.
What are the 5 most important things you learned about life?
How do you \\
describe a person in words?
Can you be happy with nothing?
Is there a word for everything?
What is \\the meaning of universe?
Who is the best person to talk to when you need advice?
What is the \\meaning of the word universe?
What is the meaning of universe according to Godungsseite.\\
According to Wikipedia, the universe refers to everything that ever existed, continues existence, or \\could possibly exist. Thus it includes all the past, all the present, and all the future.                                                        
\end{tabular}
\\ \midrule
\textbf{SliceGPT} \citep{ashkboos2024slicegpt} &  \begin{tabular}[c]{@{}l@{}}                                                 
The universe contains many galaxies, stars, planets, other objects such as asteroids, planetary systems. \\It has been estimated to contain 100 billion trillion stars. The Sun is a star in the Milky Way galaxy , \\
which is one of about 400 billion galaxies in the Milky Way's vicinity . 
The Milky Way itself is part \\
of the Milk Way Group , which includes several smaller groups of galaxies .
\end{tabular}
\\ \midrule
\textbf{\locogpt} (ours)  & \begin{tabular}[c]{@{}l@{}}The universe is a vast, dynamic, and complex structure that contains many different types of objects, \\including stars, galaxies, planets, asteroids, and intergalactic gas clouds. It is also home to life-forms \\such as humans, animals, plants, microorganisms, and microscopic organisms. 
The universe has been
\\studied by astronomers for centuries, and its size and complexity have been estimated using various \\methods.
Astronomical observations have revealed that the universe is much larger than previously \\thought, with an estimated diameter of approximately 10 billion light years. This means that the \\universe is roughly 33 times larger than our solar system.
The universe is composed of two main \\components: the cosmos and the cosmic void. The cosmos consists of galaxies, which are large \\collections of stars and other objects bound together by gravitational forces. The cosmic void, on the \\other hand, is a region of space between galaxies that is largely empty of matter.
\end{tabular} \\ \midrule
\end{tabular}

}
\end{table}

\begin{table}[t!]
\centering
\caption{{Zero-shot task performance degradation of \textsc{Llama}-2 7B, calibrated with 128 samples from the Alpaca dataset, evaluated across a broader set of tasks.}\label{tbl:diverse_7b}}
\setlength\extrarowheight{5pt}
\resizebox{1\linewidth}{!}{
\begin{tabular}{lccccccccccc}
\midrule
\textbf{Method} & \textbf{BoolQ} & \textbf{PIQA} & \textbf{HellaS.} & \textbf{WinoG.} & \textbf{ARC-e} & \textbf{ARC-c} & \textbf{OBQA} & \textbf{COPA} & \textbf{Lamb.} & \textbf{MMLU-ml} & \textbf{Average} \\ \midrule
Dense& 77.68\% & 79.05\% & 76.00\% & 68.98\% & 74.58\%&46.33\% & 44.22\% & 87.00\% & 73.86\% & 39.29\% & 66.70\% \\ \midrule
SliceGPT \citep{ashkboos2024slicegpt} & 61.99\% & 68.55\% & 48.69\% & 59.75\% & 59.69\%&34.47\% & 31.40\% & 75.00\% & 21.02\% & 23.21\% & 48.08\% \\ \midrule
ShortGPT \citep{men2024shortgpt} & 62.17\% & 64.48\% & 56.15\% & 64.33\% & 48.70\% & 32.59\% & 32.80\% & 79.00\% & 29.03\% & 24.11\% & 49.34\% \\ \midrule
\rowcolor{blue!12!white}\locogpt (ours) & \bf 69.76\% & \bf 73.34\% & \bf 65.90\% & \bf 66.22\% & \bf 65.49\% & \bf 39.16\% & \bf 39.00\% & \bf 87.00\% & \bf 57.07\% & \bf 32.14\% & \bf 59.51\% \\
\midrule
\midrule
$\Delta$ SliceGPT  & -15.69\% & -10.50\% & -27.31\% & -9.23\% & -17.89\% & -11.86\% & -12.80\% & -12.00\% & -52.84\% & -16.08\% & -18.62\% \\ \midrule
 $\Delta$ ShortGPT& -15.51\% & -14.57\% & -19.85\% & -4.65\% & -25.88\% & -13.74\% & -11.40\% & -8.00\% & -44.83\% & -15.18\% & -17.36\% \\ 
\rowcolor{blue!12!white}$\Delta$ \locogpt (ours) & -7.92\% & -5.71\% & -10.10\% & -2.76\% & -9.09\% & -7.17\% & -5.20\% & 0\% & -16.79\% & -7.15\% & -7.19\% 
\\
\midrule
\end{tabular}
}
\end{table}

{In Table \ref{tbl:diverse_7b}, we evaluate \locogpt on a broader range of tasks to assess its generalizability. We compare \locogpt with two baseline methods: SliceGPT (a decomposition approach) and ShortGPT (a layer-pruning method), all using 30\% compression of \textsc{Llama}-2 7B, with calibration performed on the Alpaca dataset. The top rows of the table show the raw accuracies, while the bottom rows display the relative degradation compared to the original dense model.

\locogpt demonstrates the least degradation across all tasks, with an average drop of only 7.19\%, while the other methods experience drops exceeding 17\%. Notably, the degradation is generally consistent across tasks, except for Lambada and MMLU, which show more significant drops. These tasks also exhibit the largest degradations in the baseline methods, suggesting they are more sensitive to compression.

Lambada, in particular, exhibits an extreme degradation in both SliceGPT and ShortGPT (over 40\% for both), making it the most challenging task to maintain accuracy after compression. In contrast, \locogpt shows a relatively small degradation of just 16.8\%, almost 25\% lower than the other methods. This hints that \locogpt is better at preserving important information, which is crucial for excelling on more difficult tasks like Lambada.
}

Finally, we compare the generation quality using samples from the three methods’ generations for 30\% compressed \textsc{Llama}-2 7B, as shown in Table \ref{tbl:gen_sample}. ShortGPT produces the lowest quality generation, while both SliceGPT and \locogpt generate high-quality responses, with \locogpt providing more detailed responses than SliceGPT.

\subsection{Additional Baseline Comparisons: Feature-Mimic and SVD Approaches}
\label{sec:fm_svd}
In Table \ref{tbl:vsfeature} and \ref{tbl:vsSVD}, we compare our method against feature-mimic and SVD-based approaches, respectively. In the former case, we observe that alternative methods generally underperform compared to state-of-the-art gradient-based techniques like LLM Surgeon, while our approach achieves comparable or even superior results. In the latter comparison, our advantage is even more pronounced, which we attribute to our more refined decomposition algorithms, tailored specifically to different components of the transformer architecture based on their levels of nonlinearity, rather than relying solely on SVD-based decompositions. 
\begin{table}[htbp]
\centering
\caption{Comparisons of feature-mimic based methods for 30\% compression of \textsc{Llama}-2 7B and 13B models.\label{tbl:vsfeature}}
\setlength\extrarowheight{2pt}
\resizebox{1\linewidth}{!}{
\begin{tabular}{c|l|ccccccc|c}
\midrule[1pt]
\textbf{Model} &  \multicolumn{1}{c|}{ \textbf{Method}} & \textbf{ARC-e} & \textbf{ARC-c} & \textbf{PIQA} & \textbf{WinoG.} & \textbf{HellaS.} & \textbf{BoolQ} & \textbf{OBQA} & \textbf{Average.} \\ 
\midrule[1pt]
\multirow{5}{*}{\textbf{\textsc{llama}-2 7B}} 
&Dense & 74.58 & 46.25 & 79.11 & 69.06 & 75.99 & 77.74 & 44.20 & 66.70 \\ 
&LLM Pruner \citep{ma2023llm} & 61.41 & 33.96 & 71.93 & 58.72 & 59.49 & 61.41 & 36.60 & 53.52 \\ 
&FLAP \citep{an2024fluctuation} & 60.65 & \underline{34.47} & \underline{72.74} & 64.01 & \underline{63.80} & 66.88 & 36.40 & 56.99 \\ 
&Bolaco (5 $\times$ 4) \citep{ji2024feature} & \bf 65.87 & 34.30 & 71.27 & \underline{64.48} & 57.85 & \bf 73.85 & \underline{37.80} & \underline{57.92} \\ 
\rowcolor{blue!12!white}\cellcolor{white}
&\locogpt (ours) & \underline{65.49} & \bf 39.16 & \bf 73.34 & \bf 66.22 & \bf 65.90 & \underline{69.76} & \bf 39.00 & \bf 59.83 \\ 
\midrule \midrule
\multirow{5}{*}{\textbf{\textsc{llama}-2 13B}} 
&Dense & 77.48 & 49.23 & 80.47 & 72.22 & 79.39 & 80.52 & 45.20 & 69.22 \\ 
&LLM Pruner \citep{ma2023llm} & 65.45 & \underline{40.36} & \underline{75.90} & 60.22 & 67.90 & 62.43 & \bf 44.60 & 59.55 \\ 
&FLAP \citep{an2024fluctuation} & 67.38 & 38.23 & 74.81 & 67.48 & \underline{70.29} & 65.54 & 40.00 & 60.53 \\ 
&Bolaco (5 $\times$ 4) \citep{ji2024feature} & \bf 71.76 & 40.10 & 74.16 & \underline{69.06} & 66.66 & \bf 75.63 & \underline{41.60} & \underline{62.71} \\ 
\rowcolor{blue!12!white}\cellcolor{white}&\locogpt (ours) & \underline{70.24} & \bf 41.47 & \bf 77.15 & \bf 71.27 & \bf 71.84 & \underline{73.7} & 41.00 & \bf 63.81 \\ 
\midrule[1pt]
\end{tabular}
}
\end{table}

\begin{table}[htbp]
\centering
    \caption{{Comparisons with SVD-based methods in \textsc{Llama}-1 7B.}\label{tbl:vsSVD}}
\setlength\extrarowheight{1.5pt}
\resizebox{1\linewidth}{!}{
\begin{tabular}{c|c|cc|ccccccc|c}
\midrule[1pt]
    \bf Compress. Rate        &  \bf Method                &  \bf  WikiText-2 {\color{Red} $\downarrow$} &  \bf  PTB {\color{Red} $\downarrow$}  & \multicolumn{1}{l}{ \textbf{ARC-e}} & \multicolumn{1}{l}{ \textbf{ARC-c}} & \multicolumn{1}{l}{ \textbf{PIQA}} & \multicolumn{1}{l}{ \textbf{WinoG.}} & \multicolumn{1}{l}{ \textbf{HellaS.}} & { \textbf{MathQA}} & { \textbf{OBQA}} &\multicolumn{1}{l}{ \textbf{Avg.}} \\ \midrule[1pt]
\large 0\%& \multicolumn{1}{l|}{\large Dense}  &  \large 5.68                            &   \large    8.35                        &     \large 73                       &  \large 42                                  &  \large 79                                 & \large 70& \large 50& \large 27        & \large 34   &         \large 54            \\ \midrule
\multirow{4}{*}{\large 20\%}&
\multicolumn{1}{l|}{\large FWSVD \citep{hsu2022language}}                      & \large 1727                            &       \large 2152                       &               \large 31              &   \large 23                                &    \large 56                          & \large 50 & \large 26 & \large 21               &  \large 15  &   \large  32          \\ 
 &\multicolumn{1}{l|}{\large ASVD \citep{yuan2023asvd}}              &   \large  11.14                &   \large  \underline{16.55}                           & \large  53                            &     \large 27                         &    \large  68                               &        \large    \underline{64}              & \large  41& \large   \underline{24}      &   \large  \underline{25}        & \large {43}                   \\ 
&\multicolumn{1}{l|}{\large SVD-LLM \citep{wang2024svd}}                &   \large \underline{7.94}                            &         \large \bf 16.22                     &    \large \underline{58}                          &   \large \underline{29}                                 &  \large \underline{69}                          & \large 58 &  \large   43    &   \large \underline{24}      &  \large 22  &\large        \underline{44}          \\ 
\rowcolor{blue!12!white}\cellcolor{white}&\multicolumn{1}{l|}{\large \locogpt (ours)}      & \large \bf6.53    &    \large  39.17                                                       &   \large \bf70                          &  \large   \bf36                              &     \large  \bf74      &\large  \bf69 &      \large \bf50              &   \large  \bf26               & \large  \bf31  &  \large   \bf51      \\ \midrule
\multirow{4}{*}{\large 40\%}
&\multicolumn{1}{l|}{\large FWSVD \citep{hsu2022language}}                           & \large 18156                              & \large  20990                           &            \large   26              &   \large  22                              &   \large 53                        &\large  51 & \large  26      &  \large \underline{21}           & \large  16  & \large  30             \\ 
&\multicolumn{1}{l|}{\large ASVD \citep{yuan2023asvd}}                                & \large  1407                             &    \large 3292                           &   \large  28                          &     \large   22                            &       \large 55                   & \large 48 & \large 26        &   \large 19            &\large  12   & \large 30              \\ 
&\multicolumn{1}{l|}{\large SVD-LLM \citep{wang2024svd}}                &    \large   \underline{13.11}                         &               \large    \underline{63.75}            &  \large \underline{42}                           &   \underline{25}                                &  \large  \underline{60}                        &\large  \underline{58} &\large  \underline{33}          &  \large  \underline{21}              &  \large \underline{19}  &  \large \underline{37}    \\ 
\rowcolor{blue!12!white}\cellcolor{white}&\multicolumn{1}{l|}{\large \locogpt (ours)}       &   \large  \bf9.39                          &  \large \bf60.55                            & \large  \bf58                            &  \large  \bf30                               &   \large \bf65         &\large  \bf64 &       \large  \bf40                &   \large \bf23     &   \large  \bf22    &    \large  \bf43      \\ \midrule[1pt]
\end{tabular}
}
\end{table}

\subsection{Additional Baseline Comparisons: Unstructured/Sem-Structured Compression}
\label{sec:unstruct-vs}

In Table \ref{tbl:vsSemi}, We compare \locogpt to the state-of-the-art non-structured methods, Wanda, SparseGPT, and ZeroPruner. These methods generally outperform \locogpt at 50\% compression rate. However, \locogpt with 40\% compression achieves a significantly better perplexity (8.41 versus 10.17). 

\begin{wrapfigure}[6]{r}{0.55\textwidth}
    \centering
    \vspace{-13pt}
    \ttabbox{
    \Large 
        \setlength\extrarowheight{2pt}
        \resizebox{.95\linewidth}{!}{
        \renewcommand{\arraystretch}{1} 
            \begin{tabular}{|l|c|cc|}
            \hline
            \multicolumn{1}{|c|}{\textbf{Method}} & \bf Structure       & 40\% & 50\%           \\ \hline
            SparseGPT (2:4)                                 & Semi-structured & -    & \textbf{10.17} \\
            Wanda (2:4)                                     & Semi-structured & -    & 11.02          \\
            ZeroPruner (2:4)                                & Semi-structured & -    & 10.52          \\
            MoDeGPT (ours)                                  & Structured      & 8.41 & 11.88          \\ \hline
            \end{tabular}
        }
    }{\caption{Comparisons with semi-structured pruning.}
        \label{tbl:vsSemi}
    }
\end{wrapfigure}
The observation suggests that with a small concession on compression rate, our structured compression can be on par with the semi-structured method that requires special GPU support for efficient inference. 
\subsection{Recovery Fine-tuning}
\begin{table}[htbp]
\centering
    \caption{{Compression and recovery fine-tuning for \textsc{llama}-2 7B using Alpaca dataset}\label{tbl:recovery-fintune}}
\setlength\extrarowheight{2pt}
\resizebox{1\linewidth}{!}{
\begin{tabular}{c|c|ccccc|c}
\midrule[1pt]
\textbf{Method}                          & \textbf{Compress.} & \textbf{ARC-e}                & \textbf{ARC-c}                & \textbf{PIQA}                 & \textbf{WinoGrande}           & \textbf{HellaSwag}             & \textbf{Average}                 \\ \midrule[1pt]
Dense                         & 0\%        &     74.58\%                   &        46.25\%         &        79.11\%          &            69.06\%         &     75.99\%                &     69.00\%        \\ \midrule
   \multirow{3}{*}{\begin{tabular}[c]{@{}c@{}}\locogpt\\    RCT-MLP\end{tabular}}  
   &20\%&
   69.78 \% ({\color{Red} $\downarrow$ 1.93\%})        &     44.20\% ({\color{Green} $\uparrow$ 2.31\%})                &     76.99\% ({\color{Green} $\uparrow$ 0.77\%})                &     66.61\% ({\color{Red} $\downarrow$ 1.58\%})                    &                69.23\% ({\color{Red} $\downarrow$ 0.36\%})      &      65.36\% ({\color{Red} $\downarrow$ 0.16\%})                        \\
                                & 30\%             &   64.94\% ({\color{Red} $\downarrow$ 0.55\%})                 & 42.15\% ({\color{Green} $\uparrow$ 2.99\%})                    &     73.83\% ({\color{Green} $\uparrow$ 0.49\%})                &          66.54\% ({\color{Green} $\uparrow$ 0.32\%})            &       67.08\% ({\color{Green} $\uparrow$ 1.18\%})                &      62.91\% ({\color{Green} $\uparrow$ 0.89\%})                \\
                                & 40\%             & 59.26\% ({\color{Red} $\downarrow$ 0.50\%}) & 37.12\% ({\color{Green} $\uparrow$ 2.39\%}) &72.09\% ({\color{Green} $\uparrow$ 1.74\%}) &64.33\% ({\color{Red} $\downarrow$ 0.07\%}) & 60.82\% ({\color{Green} $\uparrow$ 2.19\%})& 58.72\% ({\color{Green} $\uparrow$ 1.14\%})\\ \midrule

\multirow{3}{*}{\begin{tabular}[c]{@{}c@{}}\locogpt\\    RCT-ALL\end{tabular}} &    20\% & 70.45\%  ({\color{Red} $\downarrow$ 1.26\%})                 &   42.92\% ({\color{Green} $\uparrow$ 1.03\%})                  &         77.20\% ({\color{Green} $\uparrow$ 0.98\%})            &              66.30\%   ({\color{Red} $\downarrow$ 1.89\%})      &       68.07\% ({\color{Red} $\downarrow$ 1.52\%})                  &      64.99\% ({\color{Red} $\downarrow$ 0.53\%})                \\
                                & 30\%             & 63.38\% ({\color{Red} $\downarrow$ 2.11\%})& 41.47\% ({\color{Green} $\uparrow$ 2.31\%}) & 74.81\% ({\color{Green} $\uparrow$ 1.47\%}) &  66.06\% ({\color{Red} $\downarrow$ 0.16\%})&65.64\% ({\color{Red} $\downarrow$ 0.58\%})& 62.27\% ({\color{Green} $\uparrow$ 0.25\%})\\
                                & 40\%             &  58.42\% ({\color{Red} $\downarrow$ 1.34\%})                & 38.23\%  ({\color{Green} $\uparrow$ 3.50\%})           &       72.03\%         ({\color{Green} $\uparrow$ 1.68\%})      &    63.61\%  ({\color{Red} $\downarrow$ 0.79\%})                &  59.55\% ({\color{Green} $\uparrow$ 0.92\%})                  &   58.34\% ({\color{Green} $\uparrow$ 0.76\%})                  \\ \midrule[1pt]
\end{tabular}
}
\end{table}

While \locogpt does not require recovery fine-tuning, in this section, we explore how RFT can further enhance performance.
In Table \ref{tbl:recovery-fintune}, we present recovery fine-tuning results for our method on \textsc{Llama}-2 7B, following the same tuning setting as SliceGPT \citep{ashkboos2024slicegpt}. We use a calibration set of 128 random samples, each 2048 in length, from the Alpaca dataset, and a recovery fine-tuning set of 8000 samples, each 1024 in length, employing LoRA \citep{hu2021lora}. We use SliceGPT’s hyperparameters for LoRA, except for the learning rate, which is set to $5 \times 10^{-5}$.  The other primary hyperparameters used are $lora\_alpha=10$, $lora\_r=32$, $lora\_dropout=0.05$, and $batch\_size=3$. We evaluate two scenarios: 1) fine-tuning all linear matrices, and 2) tuning only the MLP.

The green and red indicators in the table denote performance increases or decreases relative to the compressed model before fine-tuning. Notably, tuning exclusively within the MLP consistently yields better performance than tuning all parameters. Since we followed the same tuning setting as SliceGPT \citep{ashkboos2024slicegpt} for a fair comparison, it is likely that better configurations exist for our method, potentially enhancing performance further. Another key observation is that despite fine-tuning using 40 times more data than calibration and employing backpropagation, \locogpt without RFT achieves very similar performance. The percentage difference is minimal, suggesting that using local reconstruction error as the objective is an effective and efficient method with our compression technique.

The table demonstrates that fine-tuning can slightly improve performance for higher compression rates, with the most significant increase observed in the ARC-c task. Evaluating the full benefits of fine-tuning remains a subject for future research.

\subsection{Experiments with Equal Computational Budgets}
\begin{table}[htb!]
\centering
\caption{{Compression comparisons with approximately equal computational budgets.}\label{tbl:fair_vs}}
\setlength\extrarowheight{6pt}
\resizebox{1\linewidth}{!}{
\begin{tabular}{l|c|c|ccccc|c}
\midrule[1pt]
\multicolumn{1}{c|}{\textbf{Method}} & \textbf{Time (Compress / Fine-tune)} & \textbf{PPL} & \textbf{ARC-e} & \textbf{ARC-c} & \textbf{PIQA} & \textbf{WinoG.} & \textbf{HellaS.} & \textbf{Average.} \\ 
\midrule[1pt]
SliceGPT& 26m / 4h05m & {\bf 2.59} (3.52) & 56.82 (56.69) & 38.48 (34.47) & 71.82 (68.55) & 59.83 (59.75) & 59.30 (48.69) & 57.26 (53.63) \\ 
SLEB& 9m / 4h50m & 2.67 (4.36) & 52.36 (52.36) & 34.04 (31.91) & 71.00 (69.58) & 59.98 (58.17) & 60.16 (58.28) & 55.51 (54.06) \\  
\locogpt & 4h09m / 31m & 2.70 ({\bf 3.08}) & \bf 67.42 (65.49) & \bf 40.96 (39.16) & \bf 74.10 (73.34) & \bf 65.98 (65.49) & \bf 66.57 (65.90) & \bf 63.01 (62.02) \\ 
\midrule[1pt]
\end{tabular}
}
\end{table}

{We study the combined effect of compression and recovery fine-tuning for different approaches with equal computational cost, as shown in Table \ref{tbl:fair_vs}
. In this experiment, we compress \textsc{LLAMA}-2 7B with a 30\% compression rate on a single A100 GPU. The model is first compressed using 128 samples from the Alpaca dataset for calibration, followed by fine-tuning with LoRA on 5k Alpaca samples. For fair comparisons, we fix the hyperparameters as $lora\_alpha=10$, $lora\_r=32$, and $lora\_dropout=0.05$. We compare \locogpt against SliceGPT and SLEB, which serve as baselines for decomposition-based and layer-pruning-based approaches, respectively.

Since the methods vary in compression time, we adjust the fine-tuning epochs to equalize the total time spent across methods. The table reports the time spent in each phase for different methods. Notably, \locogpt has the longest compression time and is therefore fine-tuned for only one epoch. 
The table presents zero-shot accuracies both before and after fine-tuning (after/before).

\locogpt achieves the highest zero-shot performance across all tasks, excluding perplexity, both before and after fine-tuning, with its performance advantage primarily arising from the compression phase. The superior perplexity but lower zero-shot performance of SliceGPT compared to \locogpt underscores the pivotal role of the compression stage, suggesting that an excessive computational focus on fine-tuning may lead to overfitting.

Lastly, SLEB, despite having the longest fine-tuning time, exhibits smaller improvements than SliceGPT in zero-shot performances, further emphasizing the pivotal role of the compression phase in determining the final model's performance. Moreover, \locogpt outperforms the baselines even without fine-tuning, demonstrating its effectiveness during the compression stage.
}
\subsection{Combination of \locogpt and SliceGPT}
\begin{table}[t!]
\centering
\caption{Perplexity performance of SliceGPT + \locogpt on \textsc{llama}-2 7B\label{tbl:slice+loco}}
\setlength\extrarowheight{2pt}
\resizebox{.9\linewidth}{!}{
\tiny
\begin{tabular}{c|c|c|c|c}
\midrule[1pt]
\begin{tabular}[c]{@{}c@{}}Slice-MLP-MHA\\ (\%-\%-\%)\end{tabular} & \begin{tabular}[c]{@{}c@{}}Compression\\ Rate\end{tabular} & \begin{tabular}[c]{@{}c@{}}WikiText2 \\ Perplexity {\color{Red}$\downarrow$}\end{tabular} & \begin{tabular}[c]{@{}c@{}}MLP Sparsity \\ Allocation\end{tabular} & \begin{tabular}[c]{@{}c@{}}MHA Sparsity\\ Allocation\end{tabular} \\ \midrule[1pt]
Dense                                                              & 0\%                                                        & 5.12                                                            & -                                                                  & -                                                                 \\ \midrule
20-20-0                                                            & 19.65\%                                                    & 7.38                                                            & \cmark                                                             & \xmark                                                            \\ \midrule
20-20-0                                                            & 19.65\%                                                    & 7.33                                                            & \xmark                                                             & \xmark                                                            \\ \midrule
25-25-0                                                            & 27.38\%                                                    & 8.42                                                            & \cmark                                                             & \xmark                                                            \\ \midrule
30-30-0                                                            & 34.93\%                                                    & 9.99                                                            & \cmark                                                             & \xmark                                                            \\ \midrule
20-25-0                                                            & 22.25\%                                                    & 7.70                                                            & \xmark                                                             & \xmark                                                            \\ \midrule
15-30-0                                                            & 11.77\%                                                    & 7.27                                                            & \xmark                                                             & \xmark                                                            \\ \midrule
10-30-0                                                            & 9.03\%                                                     & 6.83                                                            & \xmark                                                             & \xmark                                                            \\ \midrule
10-25-25                                                           & 28.00\%                                                    & 7.31                                                            & \xmark                                                             & \xmark                                                            \\ \midrule
10-30-25                                                           & 30.91\%                                                    & 7.78                                                            & \xmark                                                             & \xmark                                                            \\ \midrule
20-20-20                                                           & 29.18\%                                                    & 8.00                                                            & \xmark                                                             & \xmark                                                            \\ \midrule
\end{tabular}
}
\end{table}

\locogpt is orthogonal to SliceGPT as it reduces dimensions from different sides of a weight matrix. Figures \ref{fig:overview} (c) and (d) provide an illustrative comparison. Combining SliceGPT with \locogpt seems to be a natural extension. To demonstrate their compatibility, we experimented with various configurations as shown in Table \ref{tbl:slice+loco}. The numbers $x$-$y$-$z$ in the leftmost column indicate $x$\% slicing rate of SliceGPT, and $y$\% and $z$\% compression rates of \locogpt in MLP and MHA modules, respectively. The two rightmost columns test the use of sparsity allocation in the MLP and/or MHA modules.

Notably, our tests show that applying sparsity allocation with SliceGPT barely improves performance, consistent with the findings in the SliceGPT paper \citep{ashkboos2024slicegpt}. Therefore, we do not use sparsity allocation for slicing. Compared to the results in Table \ref{tbl:gradient}, the combination of SliceGPT and \locogpt does not improve perplexity over pure \locogpt. We attribute this to two points: 1. the significant overhead induced by slicing: to achieve a target compression rate, the model must slice at a higher rate.
2. the slicing and compression ratio might not be optimal, and it might changes from layer to layer.

Although we did not make exhaustive search, we believe there is an efficient sparsity allocation for slicing, and better tuning of the slicing and compression ratios could enhance the performance of the combined method. We leave this as a topic for future research.

\subsection{Compression Time and Memory Consumption}\label{supp:memory}
\begin{table}[!htbp]
\centering
    \caption{{Compression computations for calibration set of size 128 in WikiText2.}\label{tbl:compute_full}}
\setlength\extrarowheight{2pt}
\resizebox{1\linewidth}{!}{
\begin{tabular}{c|ll|ll|ll}
\midrule[1pt]
\multirow{2}{*}{\textbf{Model}} & \multicolumn{2}{c|}{\textbf{\locogpt}}                 & \multicolumn{2}{c|}{\textbf{SliceGPT }\cite{ashkboos2024slicegpt}}                & \multicolumn{2}{c}{\textbf{LLM surgeon }\cite{van2023llm}}                \\
                           & \multicolumn{1}{c|}{Time} & \multicolumn{1}{c|}{GPUs} & \multicolumn{1}{c|}{Time} & \multicolumn{1}{c|}{GPUs} & \multicolumn{1}{c|}{Time}    & \multicolumn{1}{c}{GPUs} \\ \midrule[1pt]
\textsc{Llama-2} 7B                  & \multicolumn{1}{l|}{4h09m}     & 1xA100 80GB               & \multicolumn{1}{l|}{0h26m}     & 1xA100 80GB               & \multicolumn{1}{l|}{17h08m}  & 4xH100 80GB              \\ \midrule
\textsc{Llama-2} 13B                 & \multicolumn{1}{l|}{8h26m}     & 1xA100 80GB               & \multicolumn{1}{l|}{0h45m}     & 1xA100 80GB               & \multicolumn{1}{l|}{1d9h26m} & 8xH100 80GB              \\ \midrule[1pt]
\end{tabular}
}
\end{table}
\begin{table}[htb!]
\centering
    \caption{{Memory consumption and compute time of 30\% compression for blocks in transformer layers tested on a single A100 80GB GPU.}\label{tbl:Cov_time}}
\setlength\extrarowheight{3pt}
\resizebox{.9\linewidth}{!}{
\begin{tabular}{c|cc|cc}
\midrule[1pt]
\multirow{2}{*}{\bf Block} & \multicolumn{2}{c|}{\textbf{\textsc{LLAMA}-7B} (13.81 GiB)}                & \multicolumn{2}{c}{\textbf{\textsc{LLAMA}-13B} (25.92 GiB)}                \\ \cline{2-5} 
                       & \multicolumn{1}{c|}{Peak Memory (GiB)} & GPU hours & \multicolumn{1}{c|}{Peak Memory (GiB)} & GPU hours \\ \midrule
MHA                    & 15.54 ({+11.5\%})                                 & 2h52m    & 28.60 ({+9.4\%})                                &5h04m     \\ \cline{1-1}
MLP                    & 23.33 ({+68.9\%})                                  &  1h13m    & 41.40 ({+54.1\%})                                 & 3h22m     \\ \midrule[1pt]
\end{tabular}
}
\end{table}

In Table \ref{tbl:compute_full}, we compare the compression times of \locogpt, SliceGPT, and LLM Surgeon. Since \locogpt and SliceGPT do not leverage gradients, they can compress a model of size 13B using a single GPU. From previous tables, we observe that while our compute time is longer than SliceGPT, \locogpt achieves significantly better performance. Conversely, our computation time is considerably shorter than LLM Surgeon, yet we achieve comparable performance. Even when equating 1 H100 to 1 A100, our method can save up to 97\% of computations.
In Table \ref{tbl:Cov_time}, we report the peak GPU memory usage when compressing \textsc{LLAMA}-2 7B and 13B models on a single A100 GPU. The primary source of additional memory overhead, beyond the model itself, is the storage of intermediate activations required for correlation estimation in the MLP. The table shows that this overhead ranges from approximately 50\% to 70\%. However, for the 13B model, the peak memory usage remains under 50\% of the total GPU memory capacity.

\subsection{Global Sparsity Allocation}
\label{supp:global_sa}
In Table \ref{tbl:global_rank}, we report the perplexity and zero-shot performance as we vary the temperature parameter in the global sparsity allocation. Initially, the uniform strategy, corresponding to an infinite temperature, performs significantly worse than our sparsity allocation strategy.

\begin{table}[t!]
\centering
    \caption{Downstream zero-shot task performance of $30\%$ \locogpt on \textsc{Llama}-2 7B for varying global rank temperature.\label{tbl:global_rank}}
\setlength\extrarowheight{2pt}
\resizebox{1\linewidth}{!}{
\begin{tabular}{c|c|c|ccccc|c}
\midrule[1pt]
\textbf{Method}                          & $\varepsilon$ & \textbf{Perplexity {\color{Red}$\downarrow$}} & \textbf{ARC-e}                & \textbf{ARC-c}                & \textbf{PIQA}                 & \textbf{WinoGrande}           & \textbf{HellaSwag}             & \textbf{Average}                 \\ \midrule[1pt]
Dense                           & -          & 5.12 &  74.58\%                   &        46.25\%         &        79.11\%          &            69.06\%         &     75.99\%                &     69.00\%               \\ \midrule
\multirow{6}{*}{\locogpt (ours)} 
& 0.075            &  7.44  &      59.72\%         &   37.29\%                 &      68.50\%            &            65.90\%            &          61.55\%           &    58.59\%                    
\\
& 0.1            &    7.46 &      63.43\%         &   39.42\%                 &       70.78\%            &            65.59\%            &          63.24\%           &    60.49\%                    
\\
& 0.5             &7.03 & 56.14\%& 32.34\% & 67.68\%&64.88\% & 58.01\% & 55.81\%\\
                                & 1             &  7.25& 53.20\%& 31.06\% & 66.16\%&64.17\% & 56.66\% & 54.25\%\\
                                & 2             &  7.35 &53.62\%                & 31.06\%                   &65.83\%                     &63.14\%                      &      55.98\%              &  53.93\%                   \\ 
                                & $\infty$            &   9.06  &      52.36\%         &   30.80\%                 &       65.18\%            &            63.69\%            &          55.31\%           &    53.47\%                  \\
                                \midrule[1pt]
\end{tabular}
}
\end{table}
\begin{figure}[!ht]
\includegraphics[clip, trim=0cm 0cm 18.5cm 0cm,width=1\linewidth]{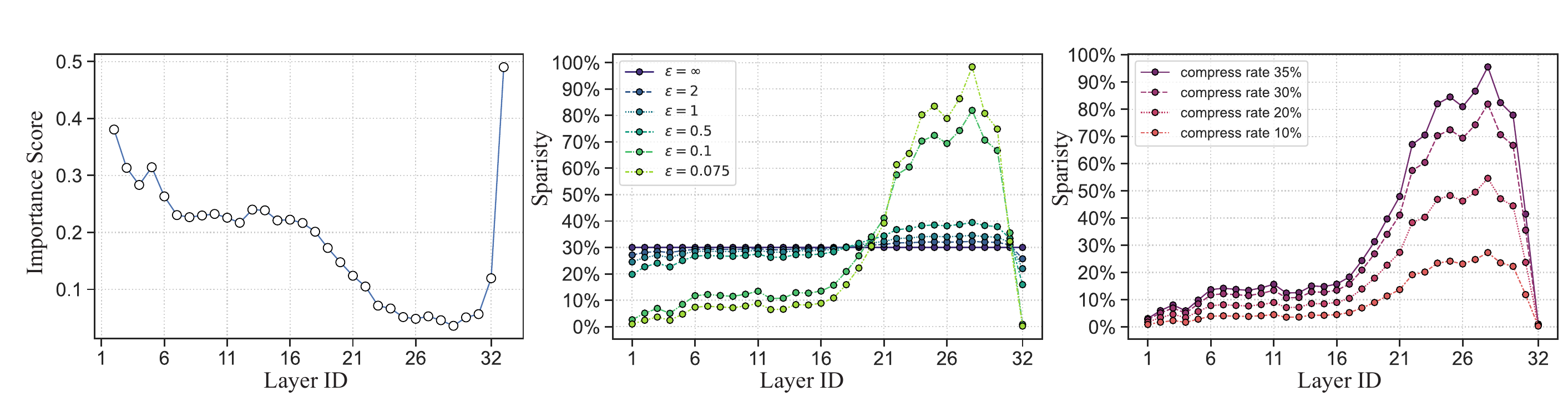}
\caption{Dynamic sparsity allocation across layers for \textsc{Llama-2} 7B.}
\label{fig:global_rank}
\end{figure}

For $\varepsilon=0.075$, this results in extreme sparsity, as shown in Figure \ref{fig:global_rank}, and performance begins to drop. For $\varepsilon \geq 1$, the allocation becomes too similar to the uniform strategy. In practice, we find that the $\varepsilon$ value that yields a minimum layer allocation around 20\% performs exceptionally well. In Figure \ref{fig:global_rank}, left and middle panels, we also observe how the allocation shape for different $\varepsilon$ values corresponds to the importance score in the left figure. For LLMs, both OPT and \textsc{Llama} show that the first and last few layers have significantly higher importance and, therefore, should allocate less sparsity, as depicted in the figure.
On the right of Figure \ref{fig:global_rank}, we also show the allocation for different compression levels. The shape remains similar across different levels using the same $\varepsilon$. For higher compression rates, the maximum layer sparsity also increases, suggesting that we should increase $\varepsilon$ to avoid extreme sparsity. {We report the ranks of the QKV projection matrices across various layers of compressed \textsc{LLaMA}-2 7B and 70B models, as determined by the global sparsity allocation used in this study (equation \ref{eq: softmax}), with their distributions visualized in Figure \ref{fig:ranks}.

The ranks were computed using \eqref{eq: softmax} with 128 samples from WikiText-2 and $\varepsilon$ values of 0.1 and 0.02 for the 7B and 70B models, respectively. These $\varepsilon$ values were selected to ensure the maximum layer sparsity remains around 70–80\%, as a 90\% sparsity level is often too extreme, based on our experience from experiments.

Interestingly, we found that the rank distributions exhibit similar shapes across the models, suggesting a deep connection between the allocation strategy and the \textsc{LLaMA}-2 family architectures.


\begin{table}[ht!]
\centering
\begin{tabular}{c|c}
\midrule
\textbf{Model}           & \textbf{Layer Rank}                            \\ \midrule
\textsc{Llama}-2 7B      & 3989, 3886, 3813, 3889, 3750, 3616, 3598, 3612         \\ 
                         & 3625, 3593, 3546, 3660, 3654, 3568, 3575, 3544         \\ 
                         & 3453, 3241, 2997, 2703, 2413, 1741, 1620, 1217         \\ 
                         & 1129, 1254, 1054, 741, 1203, 1363, 2640, 4060          \\ \midrule
\textsc{Llama}-2 70B     & 8192, 8183, 8186, 8169, 8143, 8103, 8130, 8088         \\ 
                         & 8134, 7983, 7908, 7873, 7957, 8018, 7932, 7968         \\ 
                         & 7772, 8000, 7858, 7784, 7486, 7419, 7079, 7016         \\ 
                         & 7090, 7596, 7214, 6784, 6620, 6556, 6204, 6384         \\ 
                         & 6366, 6762, 6719, 6411, 6472, 6356, 6651, 6918         \\ 
                         & 7138, 6839, 6872, 6112, 6620, 5467, 5042, 5328         \\ 
                         & 4402, 3940, 3563, 3745, 3632, 3076, 2814, 3051         \\ 
                         & 2814, 2622, 3025, 2395, 2189, 2128, 2158, 2128         \\ 
                         & 2248, 2037, 2760, 2947, 2453, 3051, 3152, 3609         \\ 
                         & 3446, 3540, 4148, 4694, 5548, 5994, 7355, 8187         \\ \hline
\end{tabular}
\caption{{Layer ranks for various models.}}
\label{tbl:sparsity_list}
\end{table}

\begin{figure}[!ht]
   \includegraphics[clip, trim=0cm 0cm 0cm 0cm,width=1\linewidth]{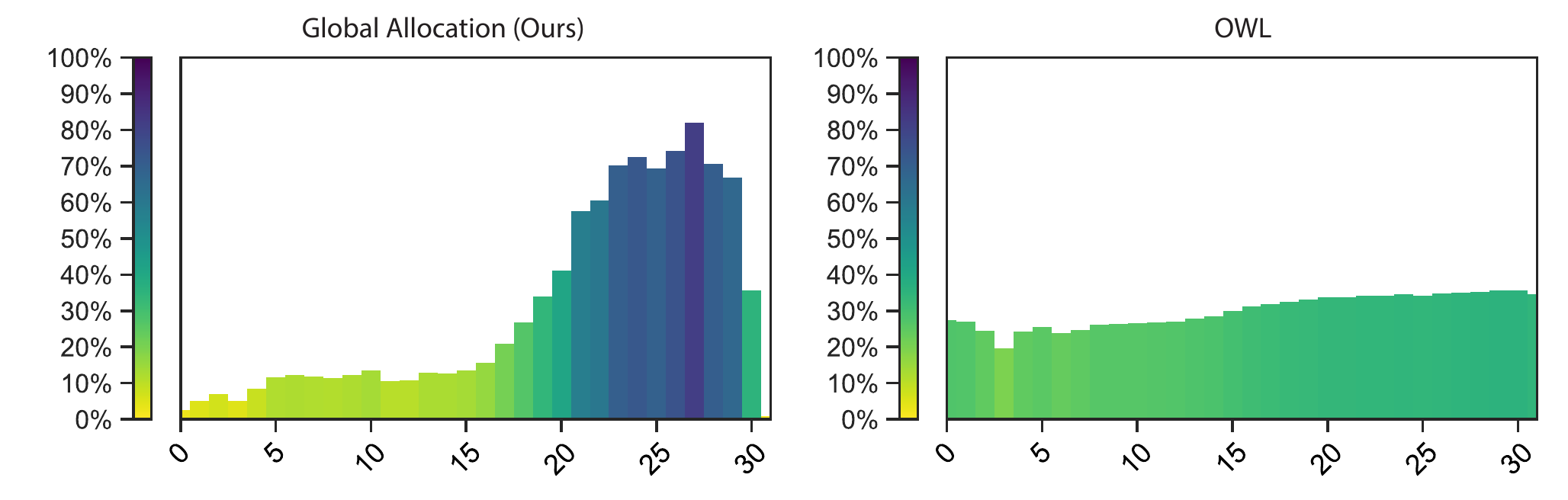}
    \caption{{{Layer sparsity distribution comparisons.
    }}
    \label{fig:owl_vs}}
\end{figure}
\begin{table}[htb!]
\centering
\renewcommand{\arraystretch}{2} 
\resizebox{1.\linewidth}{!}{ 
    \Large
    \begin{tabular}{|l|c|c|c|c|c|c|c|c|c|}
    \Xhline{1\arrayrulewidth}
    \textbf{Method} & \textbf{Sparsity Mean} & \textbf{Sparsity Std} & \textbf{Perplexity} {\color{Red}$\downarrow$} & \textbf{PIQA} & \textbf{HellaS.} & \textbf{WinoG.} & \textbf{ARC-E} & \textbf{ARC-C} & \textbf{Average} \\ \Xhline{1\arrayrulewidth}
    Uniform Allocation & 30\% & 0\% & 9.06&65.18 & 55.31 & 63.69 & 52.36 & 30.80 & 53.47  \\ \hline
    Global Sparsity Allocation (Ours) & 30\% & 26.72\% &7.51& \bf71.40&	\bf63.26&	\bf67.32	&\bf63.26&	\bf38.73&	\bf60.79 \\ \hline
    OWL \cite{yin2023outlier}& 30\% & 4.46\% & \bf 6.9	&68.17	&59.12&	65.67	&56.9&	33.36	&56.64 \\ \hline
    \end{tabular}
}
\caption{{Global sparsity allocation comparisons.}}
\label{tbl:vs_owl}
\end{table}

We also compare our global allocation strategy in \eqref{eq: softmax} with a state-of-the-art alternative, OWL \citep{yin2023outlier}, as shown in Table \ref{tbl:vs_owl}. In this experiment, we compress \textsc{LLAMA}-2 7B using \locogpt with different global sparsity strategies. Despite our method having a higher perplexity, it consistently achieves better zero-shot performance across all reported tasks. Figure \ref{fig:owl_vs}
visualizes the layer sparsity distributions of the two approaches. Unlike OWL, our distribution exhibits much greater heterogeneity across layers, showing less sparsity in the first and last layers.
This figure suggests that heterogeneity may play a crucial role in structured compression.
}
\begin{figure}[!ht]
   \includegraphics[clip, trim=0cm 0cm 0cm 0cm,width=1\linewidth]{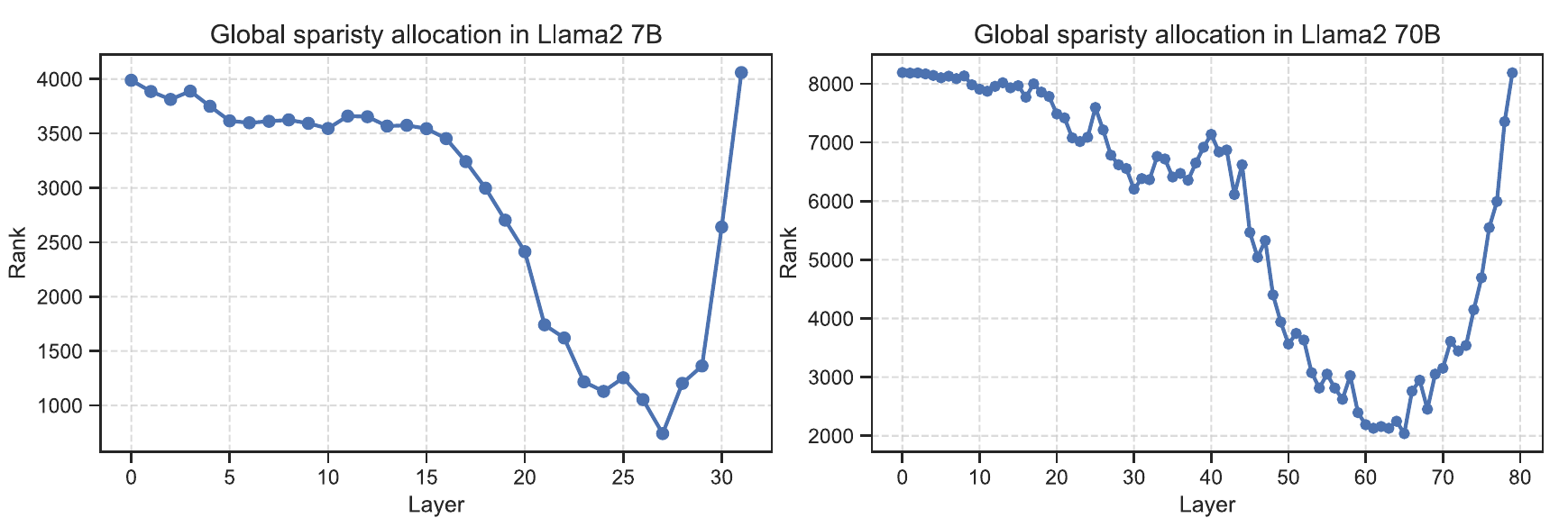}
    \caption{{Layer Ranks of \textsc{llama}-2 7B and 70B.
    }
    \label{fig:ranks}}
\end{figure}


\subsection{Refined Sparsity Allocation: Differentiating MLP and MHA Blocks with Finer-Grained Scores }
In this subsection, we refined our global sparsity allocation strategy by introducing different sparsity levels for the MLP and MHA blocks within each transformer layer. A similar approach to layer pruning, which employs distinct sparsity levels for MLP and MHA, has been explored by Finercut \citep{zhang2024finercut}. This strategic refinement has significantly improved both compression accuracy and inference throughput, particularly enhancing inference speed. Notably, in our 30\% compression experiments on \textsc{Llama}-2 7B, as illustrated in Table \ref{tbl:finergrained}, we achieved the highest throughput among all baselines, surpassing even dedicated layer pruning strategies, while maintaining exceptional accuracy.

Instead of computing a single score per layer, we now calculate two distinct scores—one for MLP and another for MHA—applying the correlation methodology outlined in Section \ref{sec:global_sparsity}. This dual-score system enables a more nuanced sparsity allocation that aligns better with the unique computational and structural characteristics of each block type, thereby optimizing performance without incurring significant computational overhead.The updated global sparsity allocation in Equation \eqref{global rank eq} is as follows:
\begin{align}
    &\max_{\phi_{1:L}}\sum_{i=1}^L\sum_{j\in \text{mlp,mha}} w_j\left(s^j_i (1-\phi^j_i) +\varepsilon H(\mathbf{\phi}_i)\right),\quad\nonumber\\ \text{such that } ~&\frac{1}{L(w_{\text{mlp}} + w_{\text{mha}})}\sum_{i=1}^L \sum_{j\in \text{mlp,mha}}w_j\phi^j_i = \phi_{\text{avg}}, \quad 0\leq \phi^j_i\leq 1,
\end{align}
where $\phi_i^j$ and $s_i^j$ represent the sparsity and score for the 
$j$-th block in layer $i$
, respectively, and the weights 
 $w_{\text{mlp}}=2,w_{\text{mha}}=1$ are applied to preserve the average sparsity, consistent with the parameter size ratio in transformer blocks. The solution has a similar closed-form expression:
\begin{align}\label{eq: softmax2}
    \bm{\phi} = L(w_{\text{mlp}} + w_{\text{mha}})\phi_{\text{avg}}\times\text{Softmax}(-\rvw\odot \rvs/\varepsilon).
\end{align}
Importantly, these updates come with minimal computational overhead. Although we now calculate two scores per layer (instead of one), the computational cost is negligible as score calculation remains lightweight and does not increase compression time. 
\begin{table}[htbp]
\centering
\renewcommand{\arraystretch}{2} 
\resizebox{1\linewidth}{!}{ 
    \Large
    \begin{tabular}{|l|c|c|c|c|c|c|c|c|c|}
    \Xhline{1\arrayrulewidth}
    \textbf{Method} & \textbf{MLP Sparsity Mean} & \textbf{MHA Sparsity Mean} & \textbf{{\color{Green} $\uparrow$} Throughput (token/s)}  & \textbf{PIQA} & \textbf{HellaS.} & \textbf{WinoG.} & \textbf{ARC-e} & \textbf{ARC-c} & {\color{Green} $\uparrow$}\textbf{Average} \\ \Xhline{1\arrayrulewidth}
    SLEB \cite{song2024sleb}& 30.0\% & 30.0\% & 2539.39 (1.49$\times$)&69.58 & 58.28 & 58.17 & 52.36 & 31.91 & 54.06  \\ \hline
    SliceGPT \cite{ashkboos2024slicegpt} & 30.0\% & 30.0\% &1815.67 (1.07$\times$)& 68.55&	48.69&	59.75	&56.69&	34.47&	53.63 \\ \hline
    \locogpt (uniform module sparsity)& 30.0\% & 30.0\% & \ 2490.15 (1.46$\times$)	&73.34	&\bf65.90	&66.22&	65.49	&39.16	&62.02 \\ \hline
    \locogpt (nonuniform module sparsity)& 26.8\% & 36.4\% & \bf 2722.98 (1.60$\times$)		&\bf73.78	&65.14	&\bf68.03	&\bf66.79	&\bf38.40	&\bf62.43 \\ \hline
    \end{tabular}
}
\caption{{Enhanced Compression through Nonuniform Sparsity in MLP and MHA Blocks.}}
\label{tbl:finergrained}
\end{table}

\subsection{Ablation Study on Compression in Each Module}
\paragraph{Impact of Module-Wise Compression on Perplexity.}
\begin{table}[htb!]
\centering
    \caption{{Perplexity of compressed \textsc{LLAMA}-2 7B in each module.}\label{tbl:ppl_module}}
\setlength\extrarowheight{3pt}
\resizebox{.9\linewidth}{!}{
\begin{tabular}{c|cccccc|c}
\midrule[1pt]
               \diagbox{\bf Module}{\bf Compression Rate}          & \bf 0\%  & \bf 10\%          & \bf 20\%          & \bf 30\%          & \bf 40\%          & \bf 50\%          & \bf Normalized Slope \\ \midrule[1pt]
Type I: MLP             & 5.12 & 5.34          & 5.68          & 6.71          & 7.12          & 8.24          & \textbf{0.094}   \\ \cline{1-1}
Type II: Query, Key     & 5.12 & \textbf{5.14} & \textbf{5.23} & 5.43          & \textbf{5.58} & 6.33          & 0.121            \\ \cline{1-1}
Type III: Value, Output & 5.12 & 5.16          & 5.24          & \textbf{5.37} & 5.62          & \textbf{5.92} & 0.095            \\ \midrule[1pt]
\end{tabular}
}
\end{table}

Table \ref{tbl:ppl_module} presents the perplexity changes in the \textsc{LLAMA}-2 7B model when compressing each module individually. The rightmost column shows the normalized slope of perplexity change relative to the parameter size in each module. The results reveal that the MLP module has the most significant impact on overall performance, likely due to its containing 66\% of the model's parameters. On the other hand, the slope indicates that the compression algorithms for Type I and Type III modules perform comparably, while Type II performs the worst. This finding aligns with our theoretical results, which suggest that the reconstruction bounds are weakest in Type III. From a decomposition perspective, the CR approximation is the most coarse-grained, leading to the least effective compression outcomes.
\paragraph{Impact of Module-Wise Compression on Throughput.}
\begin{wrapfigure}[7]{r}{0.55\textwidth}
    \centering
    \vspace{-5pt}
    \ttabbox{
    \Large 
        \setlength\extrarowheight{2pt}
        \resizebox{.9\linewidth}{!}{
        \renewcommand{\arraystretch}{1} 
            \begin{tabular}{|c|c|}
                \Xhline{2\arrayrulewidth}
                                 \textbf{Module}       & \textbf{Throughputs (tokens/s)} \\ \Xhline{2\arrayrulewidth}
                Type I: MLP             & 1585                            \\ \cline{1-1}
                Type II: Query, Key     & \textbf{2136}                   \\ \cline{1-1}
                Type III: Value, Output & 2121                            \\ 
                \Xhline{2\arrayrulewidth}
                \end{tabular}
        }
    }{ \caption{{Module-Wise Throughputs of 30\% Compressed \textsc{LLAMA}-2 7B}}
       \label{tbl:speed_module}
    }
\end{wrapfigure}
Table \ref{tbl:speed_module} presents the throughputs for the 30\% compressed \textsc{LLAMA}-2 7B across different modules. The results indicate that the compression yields similar speedups for both Type-II and Type-III modules. This sharp difference in speedups highlights the potential for uniform compression across modules, which we leave as a direction for future research.

\begin{table}[t!]
\centering
    \caption{{Heterogeneous sparsity allocation in modules.} \label{tbl:heter_module}}
\setlength\extrarowheight{2pt}
\resizebox{1\linewidth}{!}{
\begin{tabular}{c|c|ccccc|c}
\midrule[1pt]
\textbf{Sparsity (MLP, MHA)}                          & \textbf{Perplexity} {\color{Red}$\downarrow$}& \textbf{ARC-e}                & \textbf{ARC-c}                & \textbf{PIQA}                 & \textbf{WinoGrande}           & \textbf{HellaSwag}             & \textbf{Average}                 \\ \midrule[1pt]
30\%, 30\%                          & 7.51              &  \bf 65.49\%                   &        \bf 39.16\%         &       \bf 73.34\%          &            \bf 66.22\%         &     \bf 65.90\%                &     \bf 62.02\%               \\ \midrule
35\%, 20\%      & 7.79            &       \underline{60.52}\%            &    \underline{38.48}\%                   &       68.82\%          &       \underline{65.98}\%               &      61.34\%                   &  \underline{59.03}\%               \\ \midrule
25\%, 40\% &7.14             &           57.03\%         &   35.15\%                 &       \underline{70.89}\%            &            65.27\%            &          \underline{61.63}\%           &    57.99\%                   \\ \midrule[1pt]
\end{tabular}
}
\end{table}
{
\paragraph{Heterogeneous Sparsity Across Modules}
In our work, we apply nonuniform sparsity across layers while maintaining uniform sparsity across modules within the same layer. To investigate whether heterogeneity in module sparsity can enhance performance, we conducted an experiment compressing \textsc{LLAMA}-2 7B by 30\%, with varying sparsity levels in the MLP and MHA blocks. The sparsity levels were adjusted to ensure the average compression rate remained at 30\%.

We tested three configurations: equal sparsity for MLP and MHA, higher sparsity in MLP, and higher sparsity in MHA. The results are presented in Table \ref{tbl:heter_module}
. From the table, we observe that while lower sparsity in MLP yields the best perplexity, it results in the worst zero-shot performance among the three configurations. Conversely, uniform sparsity across modules outperforms the other configurations in all tasks, while high and low MLP sparsity each demonstrate strengths in specific tasks compared to one another.
These findings underscore the sensitivity of compression performance to variations in module sparsity, suggesting that a more sophisticated allocation method may be necessary to surpass the performance of uniform allocation.
}
\subsection{Scalability to Larger Models}
While our work is tested on a single GPU, it can be extended to multi-GPU setups to compress larger models, such as those with 70B parameters. To apply our method to larger models, the model must fit within the GPU memory to perform the forward pass. As shown in Table \ref{tbl:Cov_time}
, memory utilization is less than twice the model size, so approximately double the model's size in GPU memory is expected for running our method. In our compression process, the most computationally intensive part is the compression of the value-output module, as highlighted in Table \ref{tbl:component}. Since the computational complexity of this module scales cubically with the hidden dimension (due to the SVD in the value-output compression) and is proportional to the number of layers being compressed, the time required to compress a 70B model using multi-GPUs can be estimated using the following formula:
\begin{align*}
&\text{Compute Time (70B)}\\
=~&\text{Compute Time (7B)}\times \left( \frac{\text{hidden dim(70B)}}  {\text{hidden dim(7B)}}\right)^3\times \frac{\text{layer num(70B)}}{\text{layer num(7B)}}\\
=~& \text{4 hours} \times (8192 / 4096)^3 \times (80 / 32)~= \text{80 hours}
\end{align*}
For a sanity check, we applied the same formula to estimate the compression time for a 13B model and obtained an estimate of 9 hours, which aligns closely with our empirical result of 8 hours and 26 minutes, as shown in Table \ref{tbl:component}.

\subsection{High Compression Rate Experiments}
\begin{table}[htb!]
\centering
    \caption{{Perplexity of \textsc{LLAMA}-2 7B Across 10\% to 80\% Compressions}\label{tbl:large_ppl}}
\setlength\extrarowheight{3pt}
\resizebox{.9\linewidth}{!}{
\begin{tabular}{|c|ccccccccc|}
\Xhline{2\arrayrulewidth}
\bf Compression Rate & 0\%  & 10\% & 20\% & 30\% & 40\% & 50\%  & 60\%  & 70\%  & 80\%   \\ \Xhline{2\arrayrulewidth}
\bf Perplexity       & 5.12 & 5.48 & 6.16 & 7.51 & 8.41 & 11.88 & 26.59 & 84.22 & 245.84 \\ \Xhline{2\arrayrulewidth}
\end{tabular}
}
\end{table}

We analyzed the perplexity of \textsc{LLAMA}-2 7B at high compression rates, using 128 samples from WikiText2 for calibration. We observed a significant breakdown point at 50\% compression, where the perplexity increased sharply from 41\% to 123\%. This indicates the compression limit of our method.

\subsection{Sensitivity Analysis of Dfferent Calibration Sets}
\begin{table}[t!]
\centering
    \caption{{Perplexity results under different calibration datasets.}\label{tbl:sen_cal}}
\setlength\extrarowheight{2pt}
\resizebox{.8\linewidth}{!}{
\begin{tabular}{c|ccc}
\midrule[1pt]
     \diagbox{\bf Calibration Set}{\bf Test Set}     & \multicolumn{1}{c|}{ WikiText2 {\color{Red}$\downarrow$}} & \multicolumn{1}{c|}{ PTB {\color{Red}$\downarrow$}} &  Alpaca {\color{Red}$\downarrow$}\\ \midrule
 WikiText2 & \bf 6.16                           & 27.69 (+22\%)                   & 3.12 (+11\%)   \\ \cline{1-1}
 PTB       & 6.99 (+13\%)                          & \bf 22.75                    & 3.14 (+12\%)  \\ \cline{1-1}
 Alpaca    & 7.64 (+24\%)                          & 40.71 (+79\%)                   & \bf 2.80   \\ \Xhline{2\arrayrulewidth}
\end{tabular}
}
\end{table}
In Table \ref{tbl:sen_cal}, we evaluate in-domain and out-of-domain perplexity using different calibration sets: WikiText2, PTB \citep{marcus1993building}, and Alpaca. Our results indicate that perplexity is minimized when the model is calibrated with the same dataset as the test set. Notably, when calibrated with different datasets, the results on Alpaca demonstrate the most consistent performance with the least variance, while PTB shows the highest variance. Nevertheless, calibration with PTB provides the most robust results across all three datasets.

\subsection{Additional Speedup Experiments}\label{supp:speedup}
\begin{table}[!htbp]
\centering
    \caption{Inference speed and computational complexity of the pruned \textsc{Llama}-2 7B model.}
    \label{table:inf time}
\setlength\extrarowheight{2pt}
\resizebox{1\linewidth}{!}{
\begin{tabular}{c|c|c|c|c|c|c}
\midrule[1pt]
           \multirow{1}{*}{\bf Method}   & \multicolumn{1}{c|}{\begin{tabular}[c]{@{}c@{}}\bf \# Parameter\\ (B)\end{tabular}} & \multicolumn{1}{c|}{\begin{tabular}[c]{@{}c@{}}\bf Memory\\ (GiB)\end{tabular}} & \multicolumn{1}{c|}{\begin{tabular}[c]{@{}c@{}}\bf Compute Complexity\\ (GMACs) {\color{Red} $\downarrow$}\end{tabular}} & \multicolumn{1}{c|}{\begin{tabular}[c]{@{}c@{}}\bf Latency CPU\\ (s/token) {\color{Red} $\downarrow$}\end{tabular}} & \multicolumn{1}{c|}{\begin{tabular}[c]{@{}c@{}}\bf Latency GPU\\ (s/token) 
 {\color{Red} $\downarrow$}\end{tabular}} & \begin{tabular}[c]{@{}c@{}}\bf 256-Batch Throughputs \\ (tokens/s) {\color{Green} $\uparrow$}\end{tabular} \\ \midrule
Dense         & 6.74                                                                            & 12.92                                                                       & 425.12 (1.00$\times$)                                                                                   & 32.41 (1.00$\times$)                                                                        & 0.035 (1.00$\times$)                                                                        & 1700 (1.00$\times$)                                                                  \\ \midrule
20\% SliceGPT & 5.45                                                                            & 10.45                                                                       & \bf 339.04 (0.80$\times$)                                                                                   & 26.46 (0.82$\times$)                                                                        & 0.037 (1.06$\times$)                                                                        & 1802 (1.06$\times$)                                                                  \\ 
\rowcolor{blue!5!white}20\% \locogpt  & 5.44                                                                            & 10.43                                                                       & 339.34 (0.80$\times$)                                                                                   & \bf 22.66 (0.70$\times$)                                                                        & \bf 0.034 (0.97$\times$)                                                                        & \bf 2168 (1.28$\times$)                                                                  \\ \midrule
30\% SliceGPT & 4.73                                                                            & 9.07                                                                        & 298.36 (0.70$\times$)                                                                                   & 25.28 (0.78$\times$)                                                                        & 0.037 (1.06$\times$)                                                                        & 1830 (1.08$\times$)                                                                  \\ 
\rowcolor{blue!5!white}30\% \locogpt  & 4.79                                                                            & 9.07                                                                        & \bf 297.91 (0.70$\times$)                                                                                   & \bf 19.20 (0.59$\times$)                                                                        & \bf 0.034 (0.97$\times$)                                                                        & \bf 2521 (1.48$\times$)                                                                  \\ \midrule
40\% SliceGPT & 4.11                                                                            & 7.88                                                                        & 262.12 (0.62$\times$)                                                                                   & 22.68 (0.70$\times$)                                                                        & 0.037 (1.06$\times$)                                                                        & 1839 (1.08$\times$)                                                                  \\ 
\rowcolor{blue!5!white}40\% \locogpt  & 4.14                                                                            & 7.94                                                                        & \bf 256.34 (0.60$\times$)                                                                                   & \bf 18.57 (0.57$\times$)                                                                        & \bf 0.036 (1.03$\times$)                                                                        & \bf 2568 (1.51$\times$)                                                                  \\ \midrule[1pt]
\end{tabular}
}
\end{table}

\begin{figure}[!ht]
    \includegraphics[clip, trim=0cm 0cm 15cm 0cm,width=0.8\linewidth]{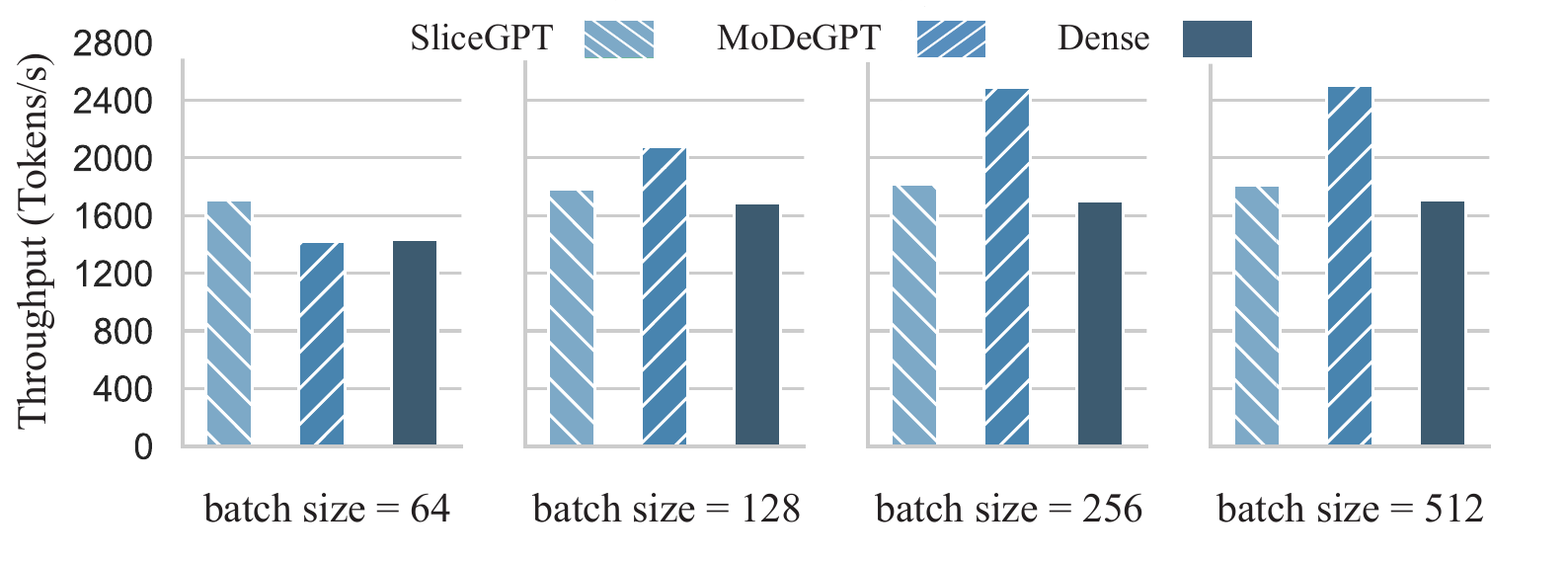}
    \caption{{Throughput benchmarks of compressed \textsc{Llama-2} 7B on a single A100 80GB GPU.
    }
    \label{fig:speedup}}
\end{figure}

\begin{figure}[!ht]
   \includegraphics[clip, trim=16.6cm 0cm 0cm 0cm,width=.6\linewidth]{figures/speedup5.pdf}
    \caption{{Speedup vs. compression.
    }
    \label{fig:speed_compress}}
\end{figure}

Table \ref{table:inf time}
reports the throughputs and latency on fast and slow parallel computing environments using NVIDIA A100 and Intel(R) Xeon(R) CPU E5-2670 v2 @ 2.50 GHz with 20 cores. The results indicate that the reduction in computational complexity is proportional to the compression percentage. While the speedup for single-batch inference is comparable to that of the original model, we observe significant improvements in speedup for multi-batch inference on the GPU and on the CPU. Therefore, our method performs optimally when the parallel computing capabilities of the environment are fully utilized.

In Figure \ref{fig:speedup}, we explored various batch sizes, comparing the throughput of 30\% compressed \locogpt with 30\% sliced SliceGPT \citep{ashkboos2024slicegpt} and the dense model. We found that throughput surpassed that of the dense model for batch sizes over 64. Particularly, at batch sizes exceeding 256, \locogpt's throughput was 1.46 times greater than the dense model, while SliceGPT only achieved 1.07 times the throughput. \locogpt's increased throughput stems from its reduced matrix size and avoidance of extra adapters in residual paths.

Finally, we benchmark throughput. We set the sequence length to 256 and recorded the average generation time of \textsc{Llama}-2 7B on a single A100 GPU with batch size 256. 
In Figure \ref{fig:speed_compress}, SVD exhibits lower throughput than the uncompressed model due to the doubled amount of matrices in its decomposed form which makes computation less parallelizable. SliceGPT, while achieving greater throughput, sees less than a 10\% speedup, hindered by additional computations in residual paths. In contrast, \locogpt achieves non-trivial speedups that increase with compression rates; at 50\% compression, it achieves a 58\% increase in throughput, significantly surpassing both SVD and SliceGPT. However, at compression rates below 10\%, throughput drops below that of the uncompressed model. This decrease is attributed to the implementation of the compressed Type-II module, which needs an optimized kernel to better parallelize the computation of the pruned attention heads. We leave the implementation of such an optimized computation kernel as future work to address the corner case.


\section{Limitations and Broader Impacts}
\paragraph{Intrinsic Bias}
Our experiments on zero-shot tasks show that \locogpt excels in certain zero-shot tasks while underperforming in others, indicating an intrinsic bias toward specific tasks. Our current method does not offer a definitive solution to eliminate this bias. Addressing bias removal will be a critical area for future research.
\paragraph{Overfitting the Reconstruction Loss}
While \locogpt excels in zero-shot tasks by minimizing local reconstruction error, we noted instances where compressed models, despite achieving lower perplexity, underperformed in zero-shot tasks. This discrepancy may stem from the models overfitting local reconstructions to calibration data. Addressing this overfitting remains a challenge for our method.
\paragraph{Broader Impacts}
The introduction of \textbf{Mo}dular \textbf{De}composition (\locogpt) significantly impacts the ethical deployment and broader adoption of Large Language Models (LLMs). By minimizing computational demands, \locogpt enables effective deployment on resource-constrained devices, democratizing access to cutting-edge AI and potentially reducing the technological divide between large and small entities.

Additionally, \locogpt’s efficiency in using computational resources can decrease energy consumption during AI training and inference, promoting sustainable AI practices and reducing the environmental impact of large-scale computations. However, the potential for increased misuse of AI technologies, such as surveillance and disinformation, highlights the need for robust governance and ethical frameworks.

Ultimately, by maintaining high accuracy while reducing model size, \locogpt ensures the reliability of AI applications in critical domains such as healthcare. The development of \locogpt thus promises greater AI accessibility and sustainability, but it also introduces new challenges in governance and ethical technology use.

\end{document}